\documentclass[10pt,twocolumn,letterpaper]{article}

\usepackage[pagenumbers]{cvpr} 

\definecolor{cvprblue}{rgb}{0.21,0.49,0.74}
\usepackage[pagebackref,breaklinks,colorlinks,allcolors=cvprblue]{hyperref}

\usepackage{tikz}
\usetikzlibrary{positioning}
\usepackage[svgnames]{xcolor}
\newcommand{\highlight}[2]{\tikz[baseline]{\node[fill=#1, anchor=base, minimum width=2em, inner sep=1pt, outer sep=0.5pt, rounded corners] {#2};}}

\usepackage[most]{tcolorbox} 

\def\paperID{*****} 
\def\confName{CVPR}
\def\confYear{2026}

\title{\raisebox{-0.2\height}{\includegraphics[height=1.5em]{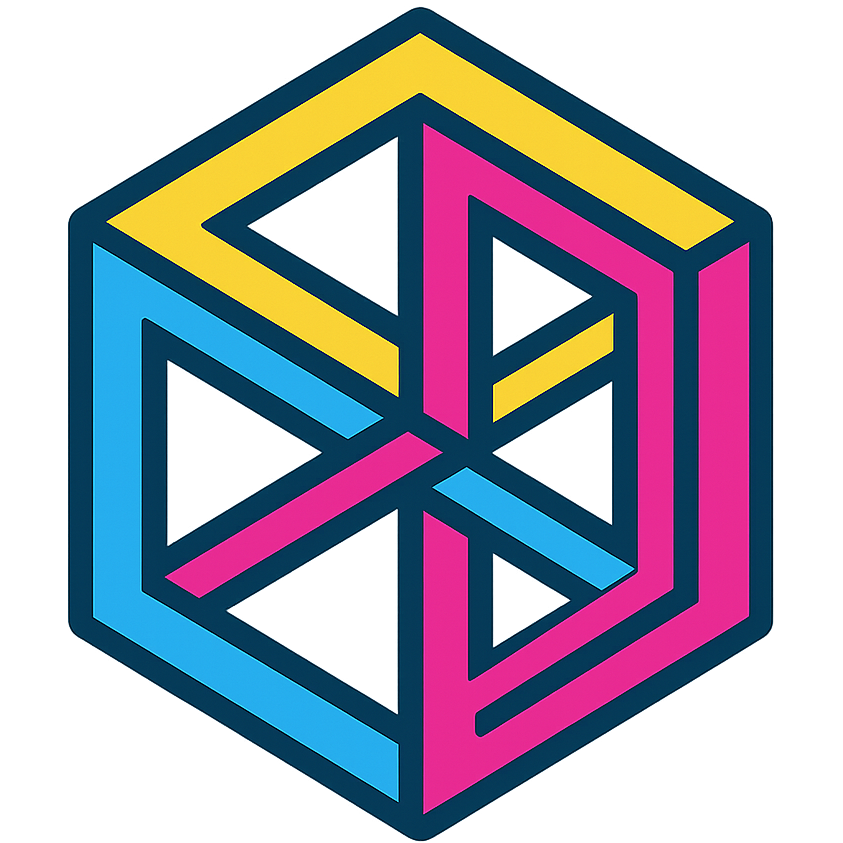}} DynaSolidGeo: A Dynamic Benchmark for Genuine Spatial Mathematical Reasoning of VLMs in Solid Geometry}

\author{
    Changti Wu\textsuperscript{\rm 1,\rm 2}\thanks{These authors contributed equally},\space 
    Shijie Lian\textsuperscript{\rm 3,\rm 2}\footnotemark[1],\space
    Zihao Liu\textsuperscript{\rm 4,\rm 2},\space
    Lei Zhang\textsuperscript{\rm 1}\thanks{Corresponding author},\space
    Laurence Tianruo Yang\textsuperscript{\rm 5,\rm 3},\space
    Kai Chen\textsuperscript{\rm 6,\rm 2}\footnotemark[2] \\
    \textsuperscript{1}East China Normal University \space
    \textsuperscript{2}Zhongguancun Academy\\
    \textsuperscript{3}Huazhong University of Science and Technology \space
    \textsuperscript{4}Peking University\\
    \textsuperscript{5}Zhengzhou University \space 
    \textsuperscript{6}Zhongguancun Institute of Artificial Intelligence
}

\begin{document}
\maketitle
\begin{abstract}
Solid geometry problem solving demands spatial mathematical reasoning that integrates spatial intelligence and symbolic reasoning.
However, most existing multimodal mathematical reasoning benchmarks focus primarily on 2D plane geometry, rely on static datasets prone to data contamination and memorization, and evaluate models solely by final answers, overlooking the reasoning process.
To address these limitations, we introduce DynaSolidGeo, the first dynamic benchmark for evaluating genuine spatial reasoning in Vision-Language Models (VLMs).
Constructed through a semi-automatic annotation pipeline, DynaSolidGeo contains 503 expert-curated seed questions that can, in principle, dynamically generate an unbounded number of diverse multimodal text-visual instances. Beyond answer accuracy, we incorporate process evaluation based on expert-annotated reasoning chains to measure logical validity and causal coherence.
Experiments across representative open-source and closed-source VLMs reveal large performance gaps, severe degradation in dynamic settings, and poor performance on tasks requiring high-level spatial intelligence, such as mental rotation and visualization.
The code and dataset are available at \href{https://zgca-ai4edu.github.io/DynaSolidGeo/}{DynaSolidGeo}.
\end{abstract}    
\begin{table*}[!thp]
\centering
\renewcommand{\arraystretch}{1.2}
\setlength\tabcolsep{6pt} 
\resizebox{\linewidth}{!}
{
\begin{tabular}{l|ccccccc}
\toprule
\textbf{Benchmarks}   & \textbf{Language} & \textbf{S.G. Size (Proportion)} & \textbf{S.G. Category} & \textbf{Level} & \textbf{Static/Dynamic} &  \textbf{Modality} &  \textbf{Metric} \\
    \midrule
    GeoQA \citep{chen2021geoqa} & EN\&CN & 0 (0.0\%) & - & K. & Static & Text\&Image & A.E. \\
    PGPS9K \citep{zhang2023multi} & EN & 0 (0.0\%) & - & K. & Static & Text\&Image & A.E. \\
    MMMU-MATH \citep{yue2024mmmu} & EN & 0 (0.0\%) & - & U. & Static & Text\&Image & A.E. \\
    GeoEval \citep{zhang2024geoeval} & EN & 100 (2.0\%) & - & K. & Static & Text\&Image & A.E. \\
    MATH-Vision \citep{wang2024measuring} & EN & 263 (8.7\%) & - & K.\&U. & Static & Text\&Image & A.E. \\
    OlympiadBench \citep{he2024olympiadbench} & EN\&CN & 784 (9.2\%) & - & C. & Static & Text\&Image & A.E. \\
    MathVerse \citep{zhang2024mathverse} & EN & 119 (15.1\%) & - & K. & Static & Text\&Image & A.E. \\
    GeomRel \citep{wang2025large} & EN & 0 (0.0\%) & - & K. & Static & Text\&Image & A.E.\&P.E. \\
    GeoSense \citep{xu2025geosense} & EN\&CN & 350 (20.0\%) & 2 & K. & Static & Text\&Image & A.E.\&P.E. \\
    SolidGeo \citep{wang2025solidgeo} & EN\&CN & 3113 (100.0\%) & 8 & K.\&U.\&C. & Static & Text\&Image & A.E. \\
    GeoLaux \citep{fu2025geolaux} & CN & 0 (0.0\%) & - & K. & Static & Text\&Image & A.E.\&P.E. \\
    DynaMath \citep{zou2024dynamath} & EN\&CN & 15 seed questions (3.0\%) & - & K.\&U. & Dynamic & Text\&Image & A.E. \\
    \midrule
    DynaSolidGeo (Ours) & EN\&CN & 503 seed questions (100.0\%) & 8 & K.\&C. & Dynamic & Text\&Image\&Video & A.E.\&P.E. \\
     \bottomrule
\end{tabular}
}
\caption{Comparison with existing geometry-related mathematical reasoning benchmarks.
S.G.=Solid Geometry;
\textbf{Level}: K.=K-12, U.=University, C.=Competitions;
\textbf{Metric}: P.E.=Process Evaluation, A.E.=Answer Evaluation.
}
\label{tab:comp_relatedwork}
\end{table*}

\section{Introduction}
\label{sec:intro}

Geometry problem solving has long played a central role in mathematical reasoning, requiring integrating visual understanding and symbolic reasoning across complex graphic and textual contexts \cite{xia2024geox,xu2025geosense,wang2025large,zhao2025towards,sharma2025geocoder,ning2025gns,weng2025geosketch,ma2025survey,wang2025geometryzero,cheng2025geouni,huang2025autogeo}.
According to structural properties, geometry can be categorized into plane geometry and solid geometry.
Compared to plane geometry, solid geometry imposes substantially higher spatial mathematical reasoning ability, as reasoning in three dimensions entails spatial intelligence, including spatial perception, spatial relation, spatial orientation, spatial rotation, and spatial visualization that goes beyond two-dimensional recognition \citep{xu2025defining,yang2025thinking,gardner2011frames, shepard1971mental, hegarty2005individual, lian2025euclid}.
Such tasks remain difficult even for human learners \cite{battista2007development}, and represent a formidable open challenge for current AI systems \citep{zhang2025mllms}.

In parallel, recent years have witnessed remarkable progress in multimodal large language models (MLLMs). Building on the successes of foundation models, vision-language models (VLMs) \citep{liu2023visual,bai2025qwen2,comanici2025gemini,openai2025gpt5systemcard,anthropic2025claudesonnet4.5,an2025llava,hong2025glm,fu2025trustgeogen,guo2025geovlmath,zhang2025geofm} have rapidly advanced the state of the art in a wide spectrum of multimodal understanding tasks. Among these tasks, multimodal mathematical reasoning has emerged as a challenging yet vibrant frontier, with benchmarks such as GeoQA \citep{chen2021geoqa}, MathVista \citep{lu2023mathvista}, GeomRel \citep{wang2025large}, and GeoSense \citep{xu2025geosense} exposing both the promise and the limitations of current VLMs.
These carefully designed benchmarks have played a pivotal role in advancing the field, providing standardized evaluation and catalyzing iterative improvements in model design and training paradigm.

Despite this progress, current multimodal mathematical reasoning benchmarks exhibit critical limitations. First, the vast majority of existing geometry-related benchmarks focus on plane geometry or diagram-based word problems, leaving solid geometry, which places heightened demands for spatial intelligence, largely underexplored. 
For example, PGPS9K \citep{zhang2023multi} contains more than 9,000 plane geometry questions but no solid geometry items, and in GeoEval \citep{zhang2024geoeval}, tasks involving solid geometry constitute merely 2\% of the benchmark.
Second, nearly all existing multimodal mathematical reasoning benchmarks are static, relying on fixed and finite test sets that are susceptible to data contamination and memorization. Recent analyses demonstrate that large models can memorize and regurgitate benchmark data \citep{magar2022data, deng2023investigating, li2023open, oren2023proving, cheng2025survey, tao2025detecting}, and some studies show that decontaminated re-releases often lead to substantial drops in performance \citep{zhao2024mmlu}, confirming that static evaluation may significantly overestimate true reasoning and generalization \citep{wang2025fragility}.
Similar concerns have motivated dynamic evaluations in coding \citep{jain2024livecodebench, zheng2025livecodebench} and general-purpose QA \citep{white2024livebench, wu2024antileakbench}, indicating a community-wide shift towards dynamic, contamination-resistant evaluation protocols \citep{chen2025recent,zhang2025dsi}.
Moreover, most existing multimodal mathematical reasoning benchmarks \citep{lu2023mathvista, wang2025solidgeo, wang2024measuring, zhang2024geoeval, yue2024mmmu, zhang2023multi, zou2024dynamath} evaluate models solely based on answer accuracy, which allows models suffering from data contamination or over-reliance on memorization to appear strong, while failing to reveal their genuine reasoning ability.

To address these limitations, we present DynaSolidGeo, a new benchmark for the dynamic evaluation of VLMs' genuine spatial mathematical reasoning in solid geometry.
Unlike existing static resources, DynaSolidGeo consists of 503 seed questions of solid geometry problem solving, each represented by a Python program paired with a corresponding MATLAB program.
With the correctness of the question guaranteed, each seed question is parameterized: textual variables in the question statement (e.g., endpoint labels, side lengths, areas, volumes, ratios) as well as rendering parameters of the solid geometry (e.g., camera viewpoints) can all be randomized. By supplying different random seeds, DynaSolidGeo can, in principle, generate an unbounded number of question-answer instances, where each instance can optionally include two visual versions: a randomized-view image and a 360-degree rotation video.
The seed questions of DynaSolidGeo are drawn from diverse and authoritative sources, including China’s Gaokao examinations, international mathematics competitions, and widely used training materials for competition preparation.
Together, they cover nearly all major categories of high-school and competition-level solid geometry problems (eight in total), including positional relations, angle, length, area, and volume calculations, as well as counting, dynamic, and folding tasks.
Moreover, we move beyond answer-only evaluation by incorporating process-level assessment grounded in expert-annotated reasoning chains. Through Answer Accuracy (AA), Process Score (PS), and Process-Qualified Accuracy (PA), we jointly measure answer correctness, reasoning quality, and reasoning-qualified accuracy, offering a more faithful reflection of VLMs' genuine spatial mathematical reasoning ability.
To ensure reliability, all solutions are expert-annotated by undergraduates and graduate students from the School of Mathematical Sciences, Peking University, including Chinese Mathematical Olympiad (CMO) gold medalists.
A comparative summary with related benchmarks is provided in Table \ref{tab:comp_relatedwork}.

We evaluate a range of mainstream, latest closed- and open-source VLMs on DynaSolidGeo. Experiments reveal a clear gap between most open-source and closed-source VLMs. Notably, nearly all models struggle with Counting problems, highlighting the lack of higher-order spatial intelligence, such as mental rotation and spatial visualization. Compared to the static source-question dataset, models exhibit a significant performance drop on DynaSolidGeo (up to 20.4\% for Claude-Sonnet-4.5), exposing potential data contamination and memorization effects. Furthermore, the additional metric degradation after introducing process evaluation indicates that previous static, answer-only benchmarks likely overestimated model capabilities, whereas DynaSolidGeo provides a more faithful and comprehensive evaluation of genuine spatial mathematical reasoning ability.
In summary, our contributions are as follows:
\begin{itemize}
    \item We design a semi-automatic data annotation pipeline for the seed question annotation of solid geometry problems, which minimizes human involvement without compromising annotation correctness or usability. 
    \item We propose DynaSolidGeo, the first dynamic benchmark for solid geometry problem solving, consisting of 503 carefully curated seed questions that can, in principle, automatically generate an unbounded number of diverse question instances across multiple geometry categories.
    \item We introduce a process evaluation using expert-annotated reasoning chains that, together with answer evaluation, provides a holistic measure of VLMs’ genuine spatial mathematical reasoning capability.
    \item We evaluate a series of popular and SOTA VLMs on DynaSolidGeo to gain deeper insights into their spatial mathematical reasoning abilities and conduct extensive analyses, including revealing potential data contamination and memorization phenomenon on static datasets.
\end{itemize}
\section{Related Work}\label{sec:relatedwork}

\subsection{Multimodal Mathematical Reasoning Benchmarks.}
Recent years have witnessed the emergence of multimodal benchmarks that evaluate mathematical reasoning in visually grounded settings. Early efforts include TQA \citep{kembhavi2017you} and Geometry3K \citep{lu2021inter} introduced multimodal reasoning tasks involving diagram-based science and geometry word problems with accompanying 2D visuals. More recent benchmarks, such as GeoQA \citep{chen2021geoqa}, PGPS9K \citep{zhang2023multi}, MMMU-MATH \citep{yue2024mmmu}, GeoEval \citep{zhang2024geoeval}, MATH-Vision \citep{wang2024measuring}, OlympiadBench \citep{he2024olympiadbench}, MathVerse \citep{zhang2024mathverse}, GeomRel \citep{wang2025large}, GeoSense \citep{xu2025geosense}, and GeoLaux \citep{fu2025geolaux} have broadened coverage to thousands of multimodal math problems. However, most of these resources focus on plane geometry and 2D diagrammatic reasoning, leaving solid geometry largely underexplored. A few datasets have attempted to move toward 3D: SolidGeo \citep{wang2025solidgeo} explicitly targets solid geometry but remains static datasets vulnerable to contamination and memorization; DynaMath \citep{zou2024dynamath} introduces dynamic instance generation, but solid geometry is barely represented, with only 15 problems (3\%) in the dataset.
In contrast, DynaSolidGeo fills this gap with scalable and dynamic solid-geometry coverage.

\subsection{Vision-Language Models.}
Recent vision–language models (VLMs) such as BLIP-2 \citep{li2023blip}, Flamingo \citep{alayrac2022flamingo}, and LLaVA \citep{liu2023visual} combine pretrained large language models with visual encoders, enabling open-ended multimodal reasoning and instruction following. Building on this paradigm, the latest generation of VLMs has rapidly advanced in scale, architecture, and reasoning capability. The closed-source models include GPT-5 family \citep{openai2025gpt5systemcard}, Gemini-2.5 family \citep{comanici2025gemini}, and Claude-Sonnet-4.5 \citep{anthropic2025claudesonnet4.5}, which feature deeply integrated multimodal backbones and enhanced reasoning modules. In parallel, the open-source community has introduced competitive alternatives such as LLaVA-OneVision-1.5 family \citep{an2025llava}, GLM-4.1V-9B-Thinking \citep{hong2025glm}, Llama-4-Maverick-17B-Instruct \citep{meta2025llama4maverick17binstruct}, InternVL3.5-8B \citep{xing2025caprl}, DeepSeek-VL2 \citep{wu2024deepseek}, and the Qwen3-VL family \citep{qwen2025qwen3vl}, which push the frontier of visual capabilities. 
Yet their spatial mathematical reasoning ability remains underexplored, motivating our evaluation of DynaSolidGeo.

\section{DynaSolidGeo}
We propose DynaSolidGeo, a dynamic multimodal benchmark for spatial mathematical reasoning in solid geometry, which consists of 503 expert-annotated seed questions that can expand into unbounded question–answer instances with randomized text, images, and 360-degree rotation videos, by inputting a random seed.

\subsection{Data Collection}\label{sec:data_collection}
The seed questions of DynaSolidGeo are drawn from diverse and authoritative sources to ensure both breadth and rigor.
Specifically, we collect 503 solid geometry questions (referred to as source questions) from three major categories: 1) China’s Gaokao examinations from 2014 to 2025 (11 years), 2) international mathematics competitions such as the American Invitational Mathematics Examination (AIME), the American Mathematics Competitions (AMC), and the American High School Mathematics Examination (AHSME), and 3) high-level preparation and training materials, including competition handbooks and advanced supplementary textbooks.
These sources cover nearly the full spectrum of high-school and competition-level solid geometry categories, encompassing positional relationships, angles, distances, area and volume computation, as well as combinatorial counting, dynamic scenarios, and folding/unfolding problems (see Table \ref{tab:statistics}).

\begin{figure*}[!tbhp]
  \centering
  \includegraphics[width=1\linewidth]{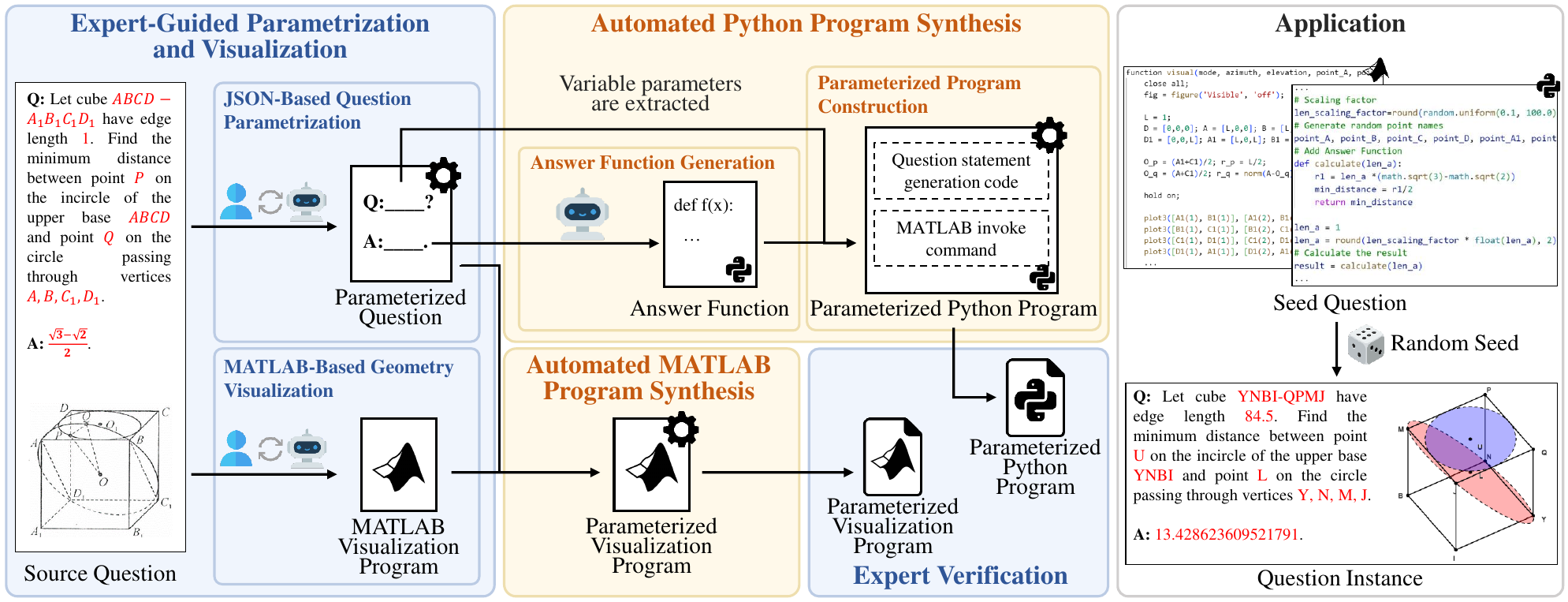}
   \caption{Overview of the data annotation pipeline and the application of seed questions. Annotation: 1) Expert-Guided Parametrization and Visualization: Each source question is first parameterized into a JSON annotation and paired with a MATLAB visualization program. 2) Automated Python Program Synthesis: The pipeline then synthesizes parameterized Python programs that generate textual descriptions and MATLAB invoke commands. 3) Automated MATLAB Program Synthesis: Correspondingly, the pipeline then synthesizes the parameterized version of MATLAB programs for figure and video rendering. 4) Expert Verification: Final human checks ensure the correctness and usability of seed questions. Application: By inputting a random seed, each seed question is instantiated into a question instance.}
   \label{fig:annotation_pipline}
\end{figure*}

\subsection{Data Annotation Pipeline}
We design a semi-automatic seed-question annotation pipeline that aims to minimize manual labeling costs while ensuring the correctness and availability of the generated programs.
Compared with a fully manual annotation process, this approach substantially reduces human effort. At the same time, in contrast to a fully automatic procedure, it preserves accuracy and reliability in handling complex solid geometry questions.
As shown in Figure \ref{fig:annotation_pipline}, our data annotation process is divided into the following components:
\begin{enumerate}
    \item \textbf{Expert-Guided Parametrization and Visualization}: Here we follow a \textit{human-in-the-loop} strategy, where human experts collaborate with large models to create a JSON annotation and a MATLAB program for each source question:
    \begin{itemize}
        \item \textit{JSON-Based Question Parametrization}: For each collected source question, mathematics experts parameterize the question statement by converting fixed values into variable parameters (e.g., endpoint labels, side lengths, areas, volumes, ratios) using f-string syntax, while ensuring correctness and availability. The corresponding answer is also expressed in terms of these variable parameters. An example of parameterized variables is highlighted in red in Fig. \ref{fig:annotation_pipline}. Additional metadata (e.g., category and difficulty level) is also included and stored in JSON format.
        \item \textit{MATLAB-Based Geometry Visualization}: MATLAB experts implement programs that render each solid geometry image and video for each source question.
    \end{itemize}
    \item \textbf{Automated Python Program Synthesis}:
    \begin{itemize}
        \item \textit{Answer Function Generation}: With the assistance of the large language model, parametrized answers in JSON are converted into Python functions that dynamically compute results.
        \item \textit{Parameterized Program Construction}: A rule-based script automatically assembles parameterized Python programs from the parametrized questions and answer functions. By inputting a random seed, the parameterized Python program randomizes both the MATLAB camera parameters (i.e., azimuth and elevation) and variable parameters in the parameterized question, and finally outputs a JSON entry of the instantiated question along with a MATLAB invoke command.
    \end{itemize}
    \item \textbf{Automated MATLAB Program Synthesis}: Each MATLAB visualization program is automatically converted into a parameterized version by a rule-based script, aligned with the annotated JSON specification. These programs can be directly invoked by the MATLAB commands generated in the previous step, enabling dynamic rendering of figures and videos consistent with the instantiated question parameters.
    \item \textbf{Expert Verification}: Final human checks ensure correctness, consistency, and usability of seed questions.
\end{enumerate}

Overall, each seed question is associated with a parameterized Python program for generating the textual description and a parameterized MATLAB program for rendering the corresponding figures and videos. By inputting a random seed, each seed question can be instantiated into a concrete question instance.

\subsection{Statistics}
Table \ref{tab:statistics} summarizes the detailed statistics of the DynaSolidGeo dataset.
In total, the benchmark contains 503 curated seed questions, all newly constructed for this work, and provides both Chinese and English versions for each question statement. The questions span a diverse set of solid geometry problem categories, including Positional relationship determination (PD, 11.7\%), Angle calculation (AN, 20.5\%), Length and distance calculation (LC, 13.1\%), Area calculation (AR, 11.3\%), Volume calculation (VC, 10.3\%), Counting problems (CP, 7.0\%), Dynamic or moving-point problems (DM, 13.1\%), and Folding and unfolding problems (FP, 12.9\%). The distribution across difficulty levels is reasonably balanced, with 27.2\% easy, 57.7\% medium, and 15.1\% hard questions.
Regarding question types, previous benchmarks (e.g., DynaMath\citep{zou2024dynamath}, SolidGeo \citep{wang2025solidgeo}, GeoSense\citep{xu2025geosense}, MathVerse\citep{zhang2024mathverse}) generally include multiple-choice questions. Such options inevitably provide strong hints to the models, thereby reducing the difficulty and making it difficult to assess their reasoning ability accurately. In contrast, we rewrite the multiple-choice and proof questions from the source data into fill-in-the-blank formats in our work. As a result, DynaSolidGeo consists of 88.3\% numerical questions and 11.7\% free-form questions, posing greater challenges to the reasoning ability of VLMs.

\begin{figure}[!htbp]
  \centering
  \includegraphics[width=1\linewidth]{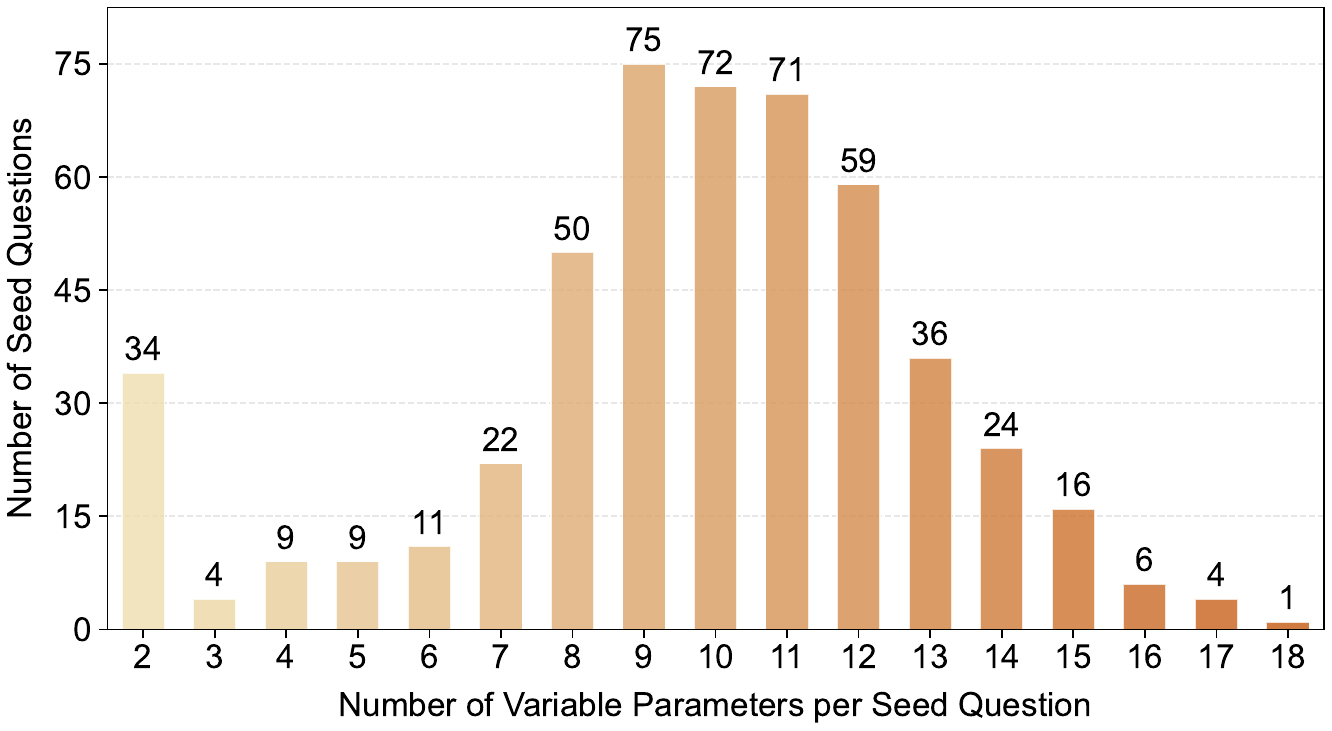}
   \caption{Distribution of variable parameters per seed question.}
   \label{fig:variable_distribution}
\end{figure}

Figure \ref{fig:variable_distribution} shows the distribution of the number of variable parameters contained in the seed questions of DynaSolidGeo. The variable parameters include camera parameters (i.e., azimuth and elevation), endpoint labels, side lengths, areas, volumes, ratios, and so on. As illustrated, the seed questions exhibit substantial variability, highlighting the richness and flexibility of our benchmark design.

\begin{table}[!htbp]
  \centering
  \resizebox{\linewidth}{!}
  {
  \begin{tabular}{ll}
  \toprule
    \textbf{Statistic} & \textbf{Number} \\
    \hline
    Total seed questions       & 503 \\
    - Newly curated questions & 503 (100.0\%) \\
    - English ver. / Chinese ver. & 503 (100.0\%) / 503 (100.0\%) \\
    \midrule
    Categories &  \\
    - Positional relationship determination (PD) & 59 (11.7\%) \\
    - Angle calculation (AN) & 103 (20.5\%) \\
    - Length and distance calculation (LC) & 66 (13.1\%) \\
    - Area calculation (AR) & 57 (11.3\%) \\
    - Volume calculation (VC) & 52 (10.3\%) \\
    - Counting problems (CP) & 35 (7.0\%) \\
    - Dynamic or moving-point problems (DM) & 66 (13.1\%) \\
    - Folding and unfolding problems (FP) & 65 (12.9\%) \\
    \midrule
    Levels &  \\
    - Easy & 137 (27.2\%) \\
    - Medium & 290 (57.7\%) \\
    - Hard & 76 (15.1\%) \\
    \midrule
    Question types &  \\
    - Numerical questions & 444 (88.3\%) \\
    - Free-form questions & 59 (11.7\%) \\
  \bottomrule
  \end{tabular}
  }
  \caption{Statistics of DynaSolidGeo}
  \label{tab:statistics}
\end{table}

\subsection{Evaluation Metrics}
Unlike most existing studies that only assess the correctness of final answers, we additionally introduce process evaluation to holistically assess the model's genuine spatial reasoning capacity.
DynaSolidGeo consists of $N=503$ seed questions, from which we instantiate $K$ batches of question instances by sampling with different random seeds. Building on these instances, we design the following evaluation metrics to systematically evaluate model performance.
\subsubsection{Answer Accuracy}
We use the Answer Accuracy (AA) to measure the correctness of the model’s answers, i.e., the proportion of final answers that are correct on average:
\begin{equation}
\mathcal{AA} = \frac{1}{K} \sum_{k=1}^K \frac{1}{N} \sum_{i=1}^N \mathbb{I}[\operatorname{Ans}_{k,i}=\operatorname{G T}_{k,i}],
\label{eq:aa}
\end{equation}
where $\operatorname{Ans}_{k,i}$ and $\operatorname{GT}_{k,i}$ denote the rule-extracted answer and the corresponding ground truth of the $i$-th question in the $k$-th batch, respectively.

\subsubsection{Process Score}
To more accurately assess the quality of the reasoning process, we also annotate parameterized reference reasoning chains.
Based on these expert-annotated reasoning chains, each reasoning process is evaluated using LLM as a judge according to the following criteria:
\begin{itemize}
    \item Logical Alignment: The reasoning presents a coherent derivation whose steps consistently lead to the stated result, with matching variables/units and without any unjustified conclusion jumps.
    \item No Extraneous Information: The reasoning does not rely on unseen quantities or unsupported facts as essential premises (standard geometric axioms and theorems are allowed).
    \item Use of Key Dependencies: The reasoning explicitly invokes the key geometric relations specified in the problem (e.g., parallelism, similarity, perpendicularity, collinearity, ratios, or angle constraints), rather than skipping these conditions and merely reporting the final result.
\end{itemize}
Based on these criteria, we introduce the process evaluation metric, termed the Process Score (PS):
\begin{equation}
\begin{aligned}
    &\mathcal{PS}=\frac{1}{K} \sum_{k=1}^K \frac{1}{N} \sum_{i=1}^N \mathcal{S}_{k,i},\\
    &\mbox{where }
    \begin{cases}
        0,& \mbox{if } \operatorname{Ans}_{k,i}\neq \operatorname{G T}_{k,i}; \\
        0.75 \le \mathcal{S}_{k,i} \le 1,& \mbox{if all above criteria are met}; \\
        0 < \mathcal{S}_{k,i} < 0.75,& \mbox{otherwise}.
    \end{cases}
\end{aligned}
\label{eq:ps}
\end{equation}
In Equation \ref{eq:ps}, $\mathcal{S}_{k,i}$ denotes the process score of the $i$-th question in the $k$-th batch scored by the judge model.
A higher PS corresponds to a reasoning process that is more accurate, coherent, and of higher quality.

\subsubsection{Process-Qualified Accuracy}
Although some questions are answered correctly, the reasoning process behind them may not be logically accurate, coherent, or supportive of the final correct answer. As a result, such ``hallucination'' cases inflate the evaluation of the model's spatial mathematical reasoning ability.
To address this, we propose a new composite metric, Process-Qualified Accuracy (PA), which combines Answer Accuracy and Process Score to more accurately measure the model’s true spatial mathematical reasoning capability:
\begin{equation}
\mathcal{PA} = \frac{1}{K} \sum_{k=1}^K \frac{1}{N} \sum_{i=1}^N \mathbb{I}[\operatorname{Ans}_{k,i}=\operatorname{G T}_{k,i}\And\mathcal{S}_{k,i} \ge 0.75].
\label{eq:pa}
\end{equation}
We believe that a model can only truly possess the ability to solve a problem when its reasoning process is logically accurate, coherent, and supports the final answer, rather than merely relying on the correctness of the final answer alone.
\begin{table*}[!thp]
\centering
\renewcommand{\arraystretch}{1.2}
\setlength\tabcolsep{4pt} 
\resizebox{\linewidth}{!}
{
\begin{tabular}{l|cccccccc|c}
\toprule
\multicolumn{1}{c|}{\smash{\raisebox{-2ex}{\textbf{Model}}}} & \textbf{PD} & \textbf{AN} & \textbf{LC} & \textbf{AR} & \textbf{VC} & \textbf{CP} & \textbf{DM} & \textbf{FP} & \textbf{ALL}\\
\cmidrule(lr){2-9} \cmidrule(lr){10-10}
& AA / PS / PA & AA / PS / PA & AA / PS / PA & AA / PS / PA & AA / PS / PA & AA / PS / PA & AA / PS / PA & AA / PS / PA & AA / PS / PA \\
    \midrule
    \multicolumn{10}{c}{\textit{\textbf{Closed-sourced MLLMs}}} \\
    \midrule
    GPT-5-Nano & 39.5 / - / - & 54.0 / - / - & 56.1 / - / - & 71.9 / - / - & 71.2 / - / - & 5.7 / - / - & 53.0 / - / - & 42.6 / - / - & 51.4 / - / - \\
    GPT-5 & \highlight{pink}{74.6} / - / - & \highlight{pink}{66.0} / - / - & \highlight{pink}{76.8} / - / - & \highlight{pink}{83.6} / - / - & \highlight{pink}{85.3} / - / - & 20.0 / - / - & \highlight{pink}{78.8} / - / - & \highlight{pink}{65.1} / - / - & \highlight{pink}{70.8} / - / - \\
    Gemini-2.5-Flash & 44.1 / 42.9 / 42.9 & 48.2 / 45.9 / 45.6 & 60.1 / 57.8 / 58.1 & 63.7 / 61.0 / 61.4 & 55.8 / 53.7 / 53.9 & 16.2 / 16.2 / 16.2 & 61.6 / 59.8 / 60.1 & 34.9 / 33.6 / 33.9 & 49.6 / 47.8 / 47.9 \\
    Gemini-2.5-Pro & 71.8 / 61.2 / 54.8 & 52.4 / 43.6 / 39.2 & 69.7 / 64.6 / 61.6 & 71.3 / 67.5 / 66.1 & 76.9 / 69.7 / 65.4 & \highlight{pink}{30.5} / 30.5 / 30.5 & 63.1 / 60.0 / 57.6 & 56.4 / 50.4 / 47.7 & 62.0 / 55.9 / 52.6 \\
    Claude-Sonnet-4.5 & 43.5 / 34.7 / 32.2 & 26.5 / 25.0 / 24.0 & 37.9 / 34.5 / 32.8 & 50.3 / 48.7 / 49.1 & 53.8 / 49.8 / 47.4 & 6.7 / 6.0 / 5.7 & 26.3 / 24.2 / 23.2 & 15.4 / 13.5 / 12.8 & 32.7 / 29.7 / 28.6 \\
    \midrule
    \multicolumn{10}{c}{\textit{\textbf{Open-sourced VLMs}}} \\
    \midrule
    LLaVA-OneVision-1.5-4B-Instruct & 11.9 / - / - & 4.5 / - / - & 7.6 / - / - & 15.8 / - / - & 10.3 / - / - & 1.9 / - / - & 4.0 / - / - & 0.0 / - / - & 6.8 / - / - \\
    LLaVA-OneVision-1.5-8B-Instruct & 17.5 / - / - & 1.9 / - / - & 7.6 / - / - & 2.9 / - / - & 2.6 / - / - & 1.0 / - / - & 5.6 / - / - & 3.1 / - / - & 5.2 / - / - \\
    GLM-4.5V & 49.7 / - / - & 31.4 / - / - & 42.9 / - / - & 57.3 / - / - & 50.6 / - / - & 7.6 / - / - & 48.0 / - / - & 12.8 / - / - & 38.1 / - / - \\
    GLM-4.1V-9B-Thinking & 29.9 / 27.3 / 26.6 & 22.7 / 21.7 / 21.7 & 33.3 / 30.7 / 31.3 & 44.4 / 42.7 / 43.3 & 41.0 / 37.8 / 39.1 & 2.9 / 2.9 / 2.9 & 26.8 / 25.5 / 25.3 & 5.6 / 4.6 / 4.1 & 26.2 / 24.6 / 24.7 \\
    Llama-3.2-90B-Vision-Instruct & 35.6 / 27.1 / 23.2 & 14.2 / 10.1 / 8.4 & 25.8 / 22.6 / 21.2 & 49.7 / 45.6 / 43.9 & 39.7 / 34.0 / 31.4 & 1.9 / 1.2 / 1.0 & 9.1 / 6.7 / 5.6 & 8.2 / 5.0 / 3.1 & 22.6 / 18.5 / 16.6 \\
    Llama-4-Maverick-17B-Instruct & 36.7 / 27.1 / 21.5 & 13.9 / 10.4 / 8.4 & 24.7 / 21.8 / 20.7 & 46.8 / 44.3 / 43.3 & 38.5 / 31.7 / 30.1 & 4.8 / 3.3 / 2.9 & 8.1 / 5.8 / 4.6 & 8.2 / 5.8 / 4.1 & 22.1 / 18.2 / 16.3 \\
    InternVL3-78B & 32.8 / 21.5 / 16.4 & 3.9 / 3.2 / 2.6 & 16.7 / 14.3 / 14.1 & 31.0 / 25.9 / 24.6 & 22.4 / 18.1 / 18.0 & 2.9 / 2.4 / 1.9 & 8.6 / 5.2 / 3.0 & 4.6 / 2.7 / 1.0 & 14.6 / 11.0 / 9.6 \\
    InternVL3.5-8B & 24.3 / 21.6 / 19.8 & 36.6 / 35.8 / 35.6 & 33.8 / 33.3 / 33.3 & 43.9 / 43.0 / 42.1 & 40.4 / 38.3 / 37.8 & 7.6 / 6.7 / 6.7 & 44.4 / 44.1 / 43.9 & 21.5 / 20.9 / 20.5 & 33.1 / 32.0 / 31.5 \\  
    DeepSeek-VL2 & 10.7 / 5.5 / 1.7 & 1.0 / 1.0 / 0.3 & 6.6 / 3.8 / 1.5 & 12.9 / 9.9 / 7.6 & 7.1 / 5.3 / 5.1 & 1.9 / 1.0 / 0.0 & 2.5 / 1.1 / 0.0 & 2.6 / 1.0 / 0.0 & 5.3 / 3.3 / 1.9 \\
    Qwen3-VL-8B-Instruct & 39.5 / 38.1 / 39.0 & 48.5 / 47.6 / 48.2 & 40.4 / 39.8 / 40.4 & 55.0 / 53.1 / 54.4 & 50.0 / 49.5 / 50.0 & 1.9 / 1.7 / 1.9 & 49.5 / 48.7 / 48.5 & 30.8 / 29.0 / 29.7 & 41.9 / 40.8 / 41.4 \\
    Qwen3-VL-8B-Thinking & 63.3 / 63.3 / 63.3 & 58.6 / 58.6 / 58.6 & 59.1 / 58.8 / 59.1 & 67.8 / 67.7 / 67.8 & 62.8 / 62.8 / 62.8 & 9.5 / 9.0 / 8.6 & 71.2 / 71.1 / 71.2 & 52.8 / 52.6 / 52.8 & 58.2 / 58.1 / 58.1 \\
    Qwen3-VL-30B-A3B-Instruct & 37.3 / 34.2 / 35.0 & 56.3 / 54.9 / 55.0 & 54.0 / 52.8 / 54.0 & 63.2 / 61.7 / 62.0 & 60.3 / 59.6 / 60.3 & 6.7 / 6.7 / 6.7 & 63.6 / 62.6 / 63.1 & 42.1 / 41.5 / 41.5 & 50.6 / 49.4 / 49.8 \\
    Qwen3-VL-30B-A3B-Thinking & 68.4 / 67.8 / 67.8 & \highlight{LightBlue}{64.4} / 64.4 / 64.4 & \highlight{LightBlue}{67.2} / 67.2 / 67.2 & 75.4 / 75.1 / 74.9 & \highlight{LightBlue}{76.3} / 75.5 / 75.6 & \highlight{LightBlue}{11.4} / 11.4 / 11.4 & \highlight{LightBlue}{78.3} / 78.2 / 78.3 & \highlight{LightBlue}{62.6} / 62.4 / 62.6 & \highlight{LightBlue}{65.6} / 65.4 / 65.4 \\
    Qwen3-VL-235B-A22B-Instruct & \highlight{LightBlue}{72.3} / 69.9 / 71.8 & 63.4 / 62.4 / 62.5 & 65.2 / 64.7 / 65.2 & \highlight{LightBlue}{76.0} / 75.1 / 75.4 & 71.8 / 70.9 / 71.2 & 6.7 / 6.7 / 6.7 & 69.7 / 69.2 / 69.2 & 57.4 / 56.9 / 57.4 & 63.1 / 62.2 / 62.6 \\

    \bottomrule
\end{tabular}
}
\caption{Comparison of model performance on the Answer Accuracy (AA), Process Score (PS), and Process-Qualified Accuracy (PA) metrics. For the GPT-5 family, LLaVA-OneVision-1.5 family, and GLM-4.5V, the PS and PA metrics are not reported, as these models either do not disclose their reasoning traces by API or inherently do not produce explicit reasoning processes.}
\label{tab:exp_main}
\end{table*}

\section{Experiment}
DynaSolidGeo supports the random generation of two visual versions: a randomized-view image and a 360-degree rotation video, for each question instance.
Since existing geometry problem-solving tasks focus exclusively on the text-image modality, here we also evaluate models under the same text-image setting.
The performance of some VLMs in the text-video modality can be found in Appendix \ref{sec:app_video}.

\subsection{Experimental Setup}\label{sec:exp_setup}
\textbf{Evaluation Models.}
We evaluate a range of the latest, popular, and state-of-the-art (SOTA) closed-source and open-source MLLMs. The closed-source models include GPT-5-Nano \citep{openai2025gpt5systemcard}, GPT-5 \citep{openai2025gpt5systemcard}, Gemini-2.5-Flash \citep{comanici2025gemini}, Gemini-2.5-Pro \citep{comanici2025gemini}, and Claude-Sonnet-4.5 \citep{anthropic2025claudesonnet4.5}. The open-source models include LLaVA-OneVision-1.5 family (4B, 8B) \citep{an2025llava}, GLM-4.5V \citep{hong2025glm}, GLM-4.1V-9B-Thinking \citep{hong2025glm}, Llama-3.2-90B-Vision-Instruct \citep{meta2024llama3_2_90B_vision_instruct}, Llama-4-Maverick-17B-Instruct \citep{meta2025llama4maverick17binstruct}, InternVL3-78B \citep{zhu2025internvl3}, InternVL3.5-8B \citep{xing2025caprl}, DeepSeek-VL2 \citep{wu2024deepseek}, and the Qwen3-VL family \citep{qwen2025qwen3vl}.

\textbf{Implementation Details.}
We sample $K = 3$ batches of question instances by setting the \textit{random seed} to 0, 1, and 2, respectively, resulting in a total of 1,509 text-image question instances.
For answer evaluation, we allow a 1\% relative error tolerance. For process evaluation, we employ Qwen3-14B \citep{yang2025qwen3} as the judge model.
For the evaluated models, we deploy small-scale models, including the LLaVA-OneVision-1.5 family, GLM-4.1V-9B-Thinking, InternVL3.5-8B, and the Qwen3-VL family (4B, 30B), on NVIDIA A800 GPUs for evaluation. DeepSeek-VL2 is evaluated via the SiliconFlow API\footnote{https://www.siliconflow.com/}, while all remaining models are accessed through the OpenRouter API\footnote{https://openrouter.ai/} for evaluation.
We set the temperature to 0.0 for all models to reduce randomness, while keeping all other hyperparameters at their default values.
More details of the experiment setup can be found in Appendix \ref{sec:app_setup}.

\subsection{Experimental Results}\label{sec:exp_res}
\textbf{Overall Results on Evaluation Metrics.}
Table \ref{tab:exp_main} presents the performance of the models in Section \ref{sec:exp_setup} on the Answer Accuracy (AA), Process Score (PS), and Process-Qualified Accuracy (PA) metrics. For the GPT-5 family, LLaVA-OneVision-1.5 family, and GLM-4.5V, the PS and PA metrics are not reported, as these models either do not disclose their reasoning traces by API or inherently do not produce explicit reasoning processes.
Among the closed-source models, GPT-5 achieves the highest overall AA score of 70.8\%, outperforming all other models. Among the open-source models, Qwen3-VL-30B-A3B-Thinking attains the highest AA score of 65.6\%, surpassing most of the closed-source models.
Among the models with available reasoning traces, Qwen3-VL-30B-A3B-Thinking achieves the highest PS and PA scores, both at 65.4\%.

\textbf{Performance Differences across Categories from the Perspective of Spatial Intelligence.}
As shown in Table \ref{tab:exp_main}, the best-performing models perform well in Area calculation (AR), Volume calculation (VC), and Dynamic or moving-point problems (DM). However, all models struggle with Counting problems (CP), and a significant performance gap can be observed between open-source and closed-source models in this category.
This discrepancy can be explained through the lens of spatial intelligence theory \citep{gardner2011frames} (for more background on spatial intelligence, see Appendix \ref{sec:app_spatial}).
Tasks such as AR, VC, and DM mainly rely on lower or mid-level spatial perception, spatial relation, and spatial orientation, where visual cues are explicit and reasoning can be simplified into formula- or rule-based deduction, rather than fine-grained 3D structural reasoning.
This aligns well with the representational strengths of current MLLMs, which operate on visual encodings and symbolic reasoning.
In contrast, Counting Problems (CP) require higher-order mental rotation and spatial visualization, requiring 3D reconstruction, occlusion reasoning, and mental manipulation of hidden or rotated objects. This explains why models perform relatively well on AR, VC, and DM tasks but fail consistently on CP tasks.

\textbf{Metric Degradation under Process Evaluation.}
As shown in Table \ref{tab:exp_main}, after introducing process evaluation, all models exhibit varying degrees of decline in both PS and PA metrics compared to AA. Among them, Gemini-2.5-Pro shows the largest drop, with PA decreasing by 9.4\% relative to AA, followed by Llama-3.2-90B-Vision-Instruct, whose PA drops by 6\%.
This suggests that these models, while capable of producing correct answers, often rely on reasoning processes that are less coherent or causally aligned with the final answers.
Furthermore, the decline in PA relative to AA is generally smaller for \textit{thinking} models than for \textit{instruct} models. For example, GLM-4.1V-9B-Thinking shows only a 1.5\% drop, and Qwen3-VL-8B-Thinking decreases by merely 0.1\%, whereas both Llama-3.2-90B-Vision-Instruct and Llama-4-Maverick-17B-Instruct experience drops exceeding 5\%. Even within the Qwen3-VL family, Qwen3-VL-8B-Thinking and Qwen3-VL-30B-A3B-Thinking exhibit smaller declines compared to Qwen3-VL-8B-Instruct, Qwen3-VL-30B-A3B-Instruct, and Qwen3-VL-235B-A22B-Instruct. These observations suggest that \textit{thinking} models generally produce reasoning processes that are more coherent, logically sound, and causally consistent with their final answers than those of \textit{instruct} models.
In addition, only a few stronger models, such as those from the Gemini and Qwen3-VL family, achieve identical AA, PS, and PA on Counting problems (CP), whereas others show clear metric gaps on different task types. This indicates that the CP task requires higher-order spatial intelligence and more rigorous symbolic reasoning, consequently making correct answers less susceptible to hallucination or logical inconsistency.

\begin{figure*}[!thbp]
  \centering
  \includegraphics[width=1\linewidth]{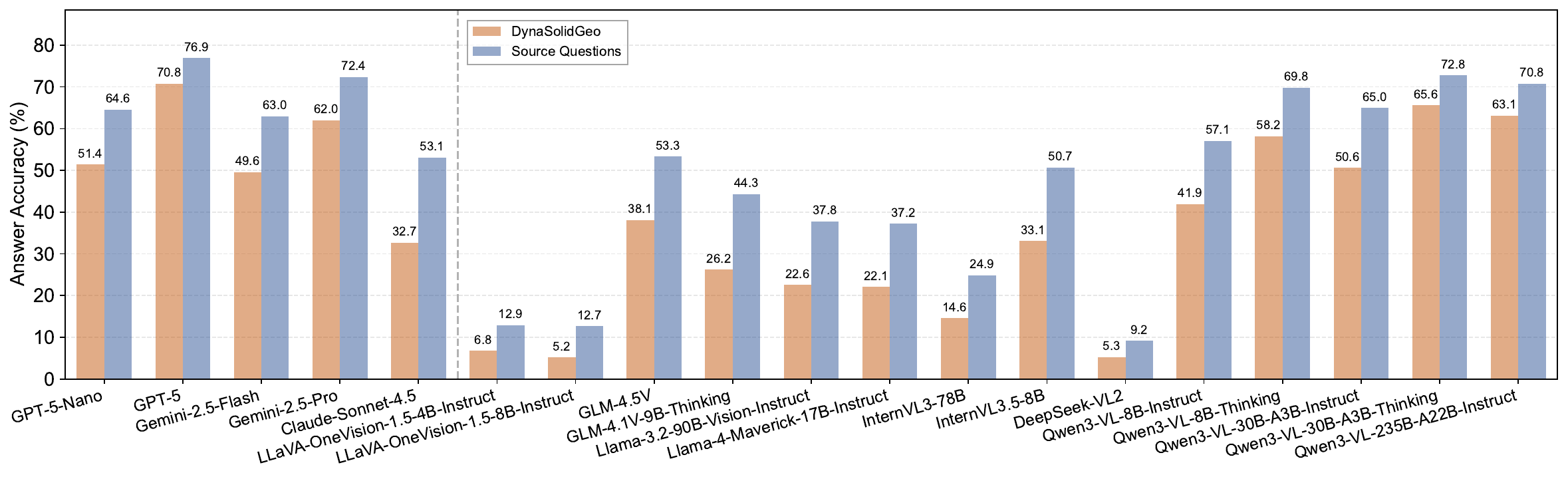}
   \caption{Comparison of model performance on Answer Accuracy (AA) between DynaSolidGeo and source questions.}
   \label{fig:comp_origin}
\end{figure*}

\textbf{Data Contamination and Memorization Phenomenon.}
To probe potential data contamination and memorization effects of VLMs on static datasets, we further evaluate their Answer Accuracy (AA) on the static source questions, as shown in Figure \ref{fig:comp_origin}.
Compared with the static source-question dataset, all models show a notable performance drop on DynaSolidGeo, with Claude-Sonnet-4.5 (-20.4\%) and InternVL3.5-8B (-17.6\%) declining the most.
This reveals that these VLMs may suffer from varying degrees of data contamination on static datasets and tend to rely on memorization-based patterns rather than genuine reasoning processes when producing answers.
In contrast, DynaSolidGeo serves as a benchmark for assessing the VLMs' genuine ability in spatial mathematical reasoning.
Furthermore, \textit{thinking} models show smaller performance declines than \textit{instruct} ones, suggesting that RL promotes reasoning-oriented behavior, while SFT encourages memorization of answers.

\begin{table}[!hbtp]
\centering
\renewcommand{\arraystretch}{1.2}
\setlength\tabcolsep{4pt} 
\resizebox{\linewidth}{!}
{
\begin{tabular}{l|c|c}
\toprule
\multicolumn{1}{c|}{\smash{\raisebox{-2ex}{\textbf{Model}}}} & \textbf{Num. of Output Tokens} & \textbf{Metrics}\\
\cmidrule(lr){2-2} \cmidrule(lr){3-3}
& Correct / Incorrect / All & AA / PS / PA \\
    \midrule
    \multicolumn{3}{c}{\textit{\textbf{Closed-sourced MLLMs}}} \\
    \midrule
    GPT-5-Nano & 8846.12 / 9739.68 / 9280.17 & 51.4 / - / - \\
    GPT-5 & 7958.86 / 9900.60 / 8526.33 & 70.8 / - / - \\
    Gemini-2.5-Flash & 4865.65 / \highlight{pink}{25448.12} / 15231.90 & 49.6 / 47.8 / 47.9 \\
    Gemini-2.5-Pro & \highlight{pink}{15054.69} / 18663.74 / \highlight{pink}{16425.13} & 62.0 / 55.9 / 52.6 \\
    Claude-Sonnet-4.5 & 987.64 / 1047.81 / 1028.15 & 32.7 / 29.7 / 28.6 \\
    \midrule
    \multicolumn{3}{c}{\textit{\textbf{Open-sourced VLMs}}} \\
    \midrule
    LLaVA-OneVision-1.5-4B-Instruct & 712.62 / 842.79 / 833.90 & 6.8 / - / - \\
    LLaVA-OneVision-1.5-8B-Instruct & 262.11 / 150.33 / 156.19 & 5.2 / - / - \\
    GLM-4.5V & 4679.18 / 5287.26 / 5055.40 & 38.1 / - / - \\
    GLM-4.1V-9B-Thinking & 6599.09 / 7562.91 / 7309.98 & 26.2 / 24.6 / 24.7 \\
    Llama-3.2-90B-Vision-Instruct & 756.23 / 821.67 / 806.88 & 22.6 / 18.5 / 16.6 \\
    Llama-4-Maverick-17B-Instruct & 761.78 / 866.27 / 843.14 & 22.1 / 18.2 / 16.3 \\
    InternVL3-78B & 562.24 / 599.96 / 594.46 & 14.6 / 11.0 / 9.6 \\
    InternVL3.5-8B & 11336.19 / 15019.64 / 13801.59 & 33.1 / 32.0 / 31.5 \\  
    DeepSeek-VL2 & 404.41 / 465.09 / 461.88 & 5.3 / 3.3 / 1.9 \\
    Qwen3-VL-8B-Instruct & 7514.76 / 18246.12 / 13751.61 & 41.9 / 40.8 / 41.4 \\
    Qwen3-VL-8B-Thinking & \highlight{LightBlue}{13251.57} / 21482.31 / \highlight{LightBlue}{16693.32} & 58.2 / 58.1 / 58.1 \\
    Qwen3-VL-30B-A3B-Instruct & 10236.35 / \highlight{LightBlue}{22068.35} / 16077.87 & 50.6 / 49.4 / 49.8 \\
    Qwen3-VL-30B-A3B-Thinking & 10954.33 / 17192.55 / 13099.88 & 65.6 / 65.4 / 65.4 \\
    Qwen3-VL-235B-A22B-Instruct & 6045.59 / 6945.21 / 6370.42 & 63.1 / 62.2 / 62.6 \\
    \bottomrule
\end{tabular}
}
\caption{Comparison of average output tokens for correct, incorrect, and overall responses with corresponding performance.}
\label{tab:exp_token}
\end{table}

\textbf{Model Inference Efficiency Analysis.}
Table \ref{tab:exp_token} summarizes, for each model, the average number of output tokens in the cases of correct and incorrect answers, as well as the overall average, together with the corresponding performance metrics.
Overall, the number of output tokens is roughly positively correlated with model performance—models with very poor accuracy consistently produce shorter outputs, which aligns with the principle of test-time scaling. In addition, for almost all models (except LLaVA-OneVision-1.5-8B-Instruct), the reasoning traces for incorrect answers are noticeably longer than those for correct ones. This is likely because, when a model encounters a problem it cannot solve or faces logical inconsistencies during reasoning, it tends to repeatedly ``rethink'' its intermediate steps, resulting in unnecessarily prolonged reasoning chains.

\begin{figure}[!thbp]
  \centering
  \includegraphics[width=1\linewidth]{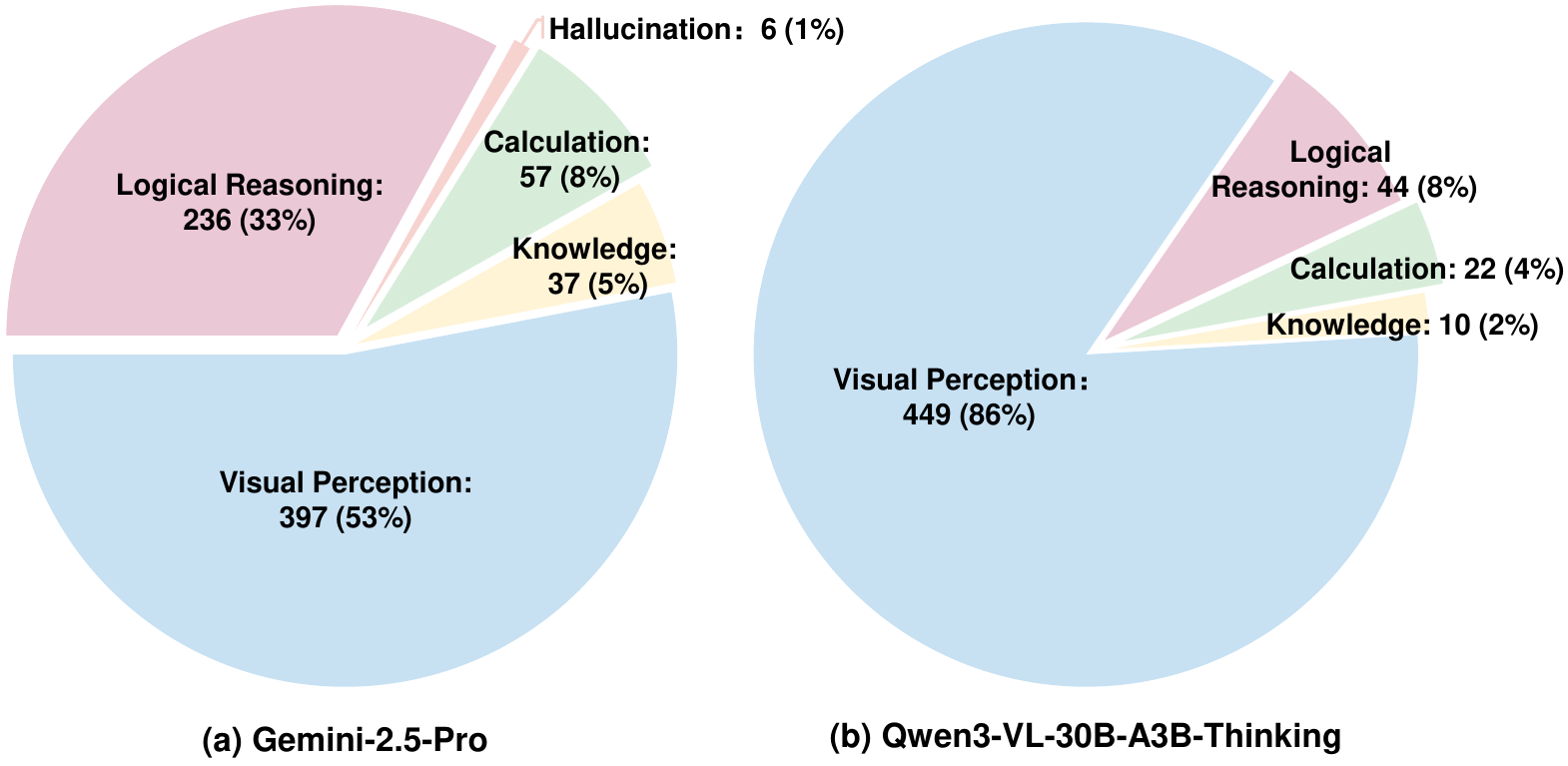}
   \caption{Error analysis.}
   \label{fig:error_analysis}
\end{figure}

\subsection{Error Analysis}
We conduct an error analysis on two representative models, Gemini-2.5-Pro (closed source) and Qwen3-VL-30B-A3B-Thinking (open source). Specifically, we categorize the errors in Process-Qualified Accuracy (PA) into five types: Visual Perception Errors, Logical Reasoning Errors, Calculation Errors, Knowledge Errors, and Hallucination Errors, as shown in Figure \ref{fig:error_analysis}.
Across the 1,509 sampled instances, Gemini-2.5-Pro makes a total of 715 errors, Qwen3-VL-30B-A3B-Thinking makes 525 errors.
Among these error types, visual perception and logical reasoning errors dominate, with visual perception errors accounting for the largest proportion. This indicates that, although these advanced models have demonstrated strong symbolic reasoning capabilities, they still lack sufficient perceptual understanding in solid geometry tasks that require spatial intelligence.
Qwen3-VL-30B-A3B-Thinking exhibits 52 more Visual Perception errors than Gemini-2.5-Pro, suggesting that it is more prone to reasoning failures triggered by inaccurate visual perception. This also explains why Qwen3-VL-30B-A3B-Thinking performs significantly worse than Gemini-2.5-Pro on Counting Problems (CP), which demand higher levels of spatial intelligence.
In addition, Qwen3-VL-30B-A3B-Thinking makes fewer errors in other categories (including Logical Reasoning Errors), demonstrating its stronger symbolic reasoning capability.
More details on the error analysis can be found in Appendix \ref{sec:app_error}.
\section{Conclusion}

In this work, we introduced DynaSolidGeo, the first dynamic benchmark for evaluating the genuine spatial mathematical reasoning capabilities of VLMs in solid geometry. Through a semi-automatic, expert-guided pipeline, DynaSolidGeo enables unbounded generation of diverse multimodal instances, effectively mitigating contamination and memorization issues found in static datasets. By integrating both answer- and process-level evaluation, we provide a more faithful assessment of logical validity and causal coherence. Comprehensive experiments uncover persistent limitations in high-level spatial intelligence and reveal substantial performance degradation under dynamic evaluation.
We expect DynaSolidGeo to provide a reliable foundation for advancing process-grounded multimodal reasoning benchmarks that mitigate data contamination and inspire future research toward robust spatial reasoning in VLMs.

\clearpage

{
    \small
    \bibliographystyle{ieeenat_fullname}
    \bibliography{main}

\begin{thebibliography}{66}
\providecommand{\natexlab}[1]{#1}
\providecommand{\url}[1]{\texttt{#1}}
\expandafter\ifx\csname urlstyle\endcsname\relax
  \providecommand{\doi}[1]{doi: #1}\else
  \providecommand{\doi}{doi: \begingroup \urlstyle{rm}\Url}\fi

\bibitem[AI(2024)]{meta2024llama3_2_90B_vision_instruct}
Meta AI.
\newblock Llama-3.2-vision-instruct.
\newblock \url{https://huggingface.co/meta-llama/Llama-3.2-90B-Vision-Instruct}, 2024.

\bibitem[AI(2025)]{meta2025llama4maverick17binstruct}
Meta AI.
\newblock Llama-4-maverick-17b-instruct.
\newblock \url{https://huggingface.co/meta-llama/Llama-4-Maverick-17B-128E-Instruct}, 2025.

\bibitem[Alayrac et~al.(2022)Alayrac, Donahue, Luc, Miech, Barr, Hasson, Lenc, Mensch, Millican, Reynolds, et~al.]{alayrac2022flamingo}
Jean-Baptiste Alayrac, Jeff Donahue, Pauline Luc, Antoine Miech, Iain Barr, Yana Hasson, Karel Lenc, Arthur Mensch, Katherine Millican, Malcolm Reynolds, et~al.
\newblock Flamingo: a visual language model for few-shot learning.
\newblock \emph{Advances in neural information processing systems}, 35:\penalty0 23716--23736, 2022.

\bibitem[An et~al.(2025)An, Xie, Yang, Zhang, Zhao, Cheng, Wang, Xu, Chen, Wu, et~al.]{an2025llava}
Xiang An, Yin Xie, Kaicheng Yang, Wenkang Zhang, Xiuwei Zhao, Zheng Cheng, Yirui Wang, Songcen Xu, Changrui Chen, Chunsheng Wu, et~al.
\newblock Llava-onevision-1.5: Fully open framework for democratized multimodal training.
\newblock \emph{arXiv preprint arXiv:2509.23661}, 2025.

\bibitem[Anthropic(2025)]{anthropic2025claudesonnet4.5}
Anthropic.
\newblock Introducing claude sonnet 4.5.
\newblock \url{https://www.anthropic.com/news/claude-sonnet-4-5}, 2025.

\bibitem[Bai et~al.(2025)Bai, Chen, Liu, Wang, Ge, Song, Dang, Wang, Wang, Tang, et~al.]{bai2025qwen2}
Shuai Bai, Keqin Chen, Xuejing Liu, Jialin Wang, Wenbin Ge, Sibo Song, Kai Dang, Peng Wang, Shijie Wang, Jun Tang, et~al.
\newblock Qwen2. 5-vl technical report.
\newblock \emph{arXiv preprint arXiv:2502.13923}, 2025.

\bibitem[Battista(2007)]{battista2007development}
MT Battista.
\newblock The development of geometric and spatial thinking.
\newblock \emph{Second handbook of research on mathematics teaching and learning/National Council of Teachers of Mathematics}, 2007.

\bibitem[Chen et~al.(2021)Chen, Tang, Qin, Liang, Liu, Xing, and Lin]{chen2021geoqa}
Jiaqi Chen, Jianheng Tang, Jinghui Qin, Xiaodan Liang, Lingbo Liu, Eric~P Xing, and Liang Lin.
\newblock Geoqa: A geometric question answering benchmark towards multimodal numerical reasoning.
\newblock \emph{arXiv preprint arXiv:2105.14517}, 2021.

\bibitem[Chen et~al.(2025)Chen, Chen, Li, Jiang, Wan, He, Ran, Gu, Li, Xie, et~al.]{chen2025recent}
Simin Chen, Yiming Chen, Zexin Li, Yifan Jiang, Zhongwei Wan, Yixin He, Dezhi Ran, Tianle Gu, Haizhou Li, Tao Xie, et~al.
\newblock Recent advances in large langauge model benchmarks against data contamination: From static to dynamic evaluation.
\newblock \emph{arXiv preprint arXiv:2502.17521}, 2025.

\bibitem[Cheng et~al.(2025{\natexlab{a}})Cheng, Zhang, Chen, Deng, Qin, and Ma]{cheng2025geouni}
Jo-Ku Cheng, Zeren Zhang, Ran Chen, Jingyang Deng, Ziran Qin, and Jinwen Ma.
\newblock Geouni: A unified model for generating geometry diagrams, problems and problem solutions.
\newblock In \emph{Proceedings of the 33rd ACM International Conference on Multimedia}, pages 3057--3066, 2025{\natexlab{a}}.

\bibitem[Cheng et~al.(2025{\natexlab{b}})Cheng, Chang, and Wu]{cheng2025survey}
Yuxing Cheng, Yi Chang, and Yuan Wu.
\newblock A survey on data contamination for large language models.
\newblock \emph{arXiv preprint arXiv:2502.14425}, 2025{\natexlab{b}}.

\bibitem[Comanici et~al.(2025)Comanici, Bieber, Schaekermann, Pasupat, Sachdeva, Dhillon, Blistein, Ram, Zhang, Rosen, et~al.]{comanici2025gemini}
Gheorghe Comanici, Eric Bieber, Mike Schaekermann, Ice Pasupat, Noveen Sachdeva, Inderjit Dhillon, Marcel Blistein, Ori Ram, Dan Zhang, Evan Rosen, et~al.
\newblock Gemini 2.5: Pushing the frontier with advanced reasoning, multimodality, long context, and next generation agentic capabilities.
\newblock \emph{arXiv preprint arXiv:2507.06261}, 2025.

\bibitem[Deng et~al.(2023)Deng, Zhao, Tang, Gerstein, and Cohan]{deng2023investigating}
Chunyuan Deng, Yilun Zhao, Xiangru Tang, Mark Gerstein, and Arman Cohan.
\newblock Investigating data contamination in modern benchmarks for large language models.
\newblock \emph{arXiv preprint arXiv:2311.09783}, 2023.

\bibitem[Fu et~al.(2025{\natexlab{a}})Fu, Chen, Xia, Liu, Feng, Zhou, Zhang, Feng, Gao, Yan, et~al.]{fu2025trustgeogen}
Daocheng Fu, Zijun Chen, Renqiu Xia, Qi Liu, Yuan Feng, Hongbin Zhou, Renrui Zhang, Shiyang Feng, Peng Gao, Junchi Yan, et~al.
\newblock Trustgeogen: Scalable and formal-verified data engine for trustworthy multi-modal geometric problem solving.
\newblock \emph{arXiv preprint arXiv:2504.15780}, 2025{\natexlab{a}}.

\bibitem[Fu et~al.(2025{\natexlab{b}})Fu, Zhu, Zhang, Zhao, Ma, Zhang, Wu, and Wu]{fu2025geolaux}
Yumeng Fu, Jiayin Zhu, Lingling Zhang, Bo Zhao, Shaoxuan Ma, Yushun Zhang, Yanrui Wu, and Wenjun Wu.
\newblock Geolaux: A benchmark for evaluating mllms' geometry performance on long-step problems requiring auxiliary lines.
\newblock \emph{arXiv preprint arXiv:2508.06226}, 2025{\natexlab{b}}.

\bibitem[Gardner(2011)]{gardner2011frames}
Howard Gardner.
\newblock \emph{Frames of mind: The theory of multiple intelligences}.
\newblock Basic books, 2011.

\bibitem[Guo et~al.(2025)Guo, Pang, Wang, Wang, Shen, and Zhang]{guo2025geovlmath}
Shasha Guo, Liang Pang, Xi Wang, Yanling Wang, Huawei Shen, and Jing Zhang.
\newblock Geovlmath: Enhancing geometry reasoning in vision-language models via cross-modal reward for auxiliary line creation.
\newblock \emph{arXiv preprint arXiv:2510.11020}, 2025.

\bibitem[He et~al.(2024)He, Luo, Bai, Hu, Thai, Shen, Hu, Han, Huang, Zhang, et~al.]{he2024olympiadbench}
Chaoqun He, Renjie Luo, Yuzhuo Bai, Shengding Hu, Zhen~Leng Thai, Junhao Shen, Jinyi Hu, Xu Han, Yujie Huang, Yuxiang Zhang, et~al.
\newblock Olympiadbench: A challenging benchmark for promoting agi with olympiad-level bilingual multimodal scientific problems.
\newblock \emph{arXiv preprint arXiv:2402.14008}, 2024.

\bibitem[Hegarty and Waller(2005)]{hegarty2005individual}
Mary Hegarty and David Waller.
\newblock Individual differences in spatial abilities.
\newblock \emph{The Cambridge handbook of visuospatial thinking}, pages 121--169, 2005.

\bibitem[Hong et~al.(2025)Hong, Yu, Gu, Wang, Gan, Tang, Cheng, Qi, Ji, Pan, et~al.]{hong2025glm}
Wenyi Hong, Wenmeng Yu, Xiaotao Gu, Guo Wang, Guobing Gan, Haomiao Tang, Jiale Cheng, Ji Qi, Junhui Ji, Lihang Pan, et~al.
\newblock Glm-4.1 v-thinking: Towards versatile multimodal reasoning with scalable reinforcement learning.
\newblock \emph{arXiv e-prints}, pages arXiv--2507, 2025.

\bibitem[Huang et~al.(2025)Huang, Wu, Lin, Zhang, Chen, and Wu]{huang2025autogeo}
Zihan Huang, Tao Wu, Wang Lin, Shengyu Zhang, Jingyuan Chen, and Fei Wu.
\newblock Autogeo: Automating geometric image dataset creation for enhanced geometry understanding.
\newblock \emph{IEEE Transactions on Multimedia}, 2025.

\bibitem[Jain et~al.(2024)Jain, Han, Gu, Li, Yan, Zhang, Wang, Solar-Lezama, Sen, and Stoica]{jain2024livecodebench}
Naman Jain, King Han, Alex Gu, Wen-Ding Li, Fanjia Yan, Tianjun Zhang, Sida Wang, Armando Solar-Lezama, Koushik Sen, and Ion Stoica.
\newblock Livecodebench: Holistic and contamination free evaluation of large language models for code.
\newblock \emph{arXiv preprint arXiv:2403.07974}, 2024.

\bibitem[Kembhavi et~al.(2017)Kembhavi, Seo, Schwenk, Choi, Farhadi, and Hajishirzi]{kembhavi2017you}
Aniruddha Kembhavi, Minjoon Seo, Dustin Schwenk, Jonghyun Choi, Ali Farhadi, and Hannaneh Hajishirzi.
\newblock Are you smarter than a sixth grader? textbook question answering for multimodal machine comprehension.
\newblock In \emph{Proceedings of the IEEE Conference on Computer Vision and Pattern recognition}, pages 4999--5007, 2017.

\bibitem[Li et~al.(2023{\natexlab{a}})Li, Li, Savarese, and Hoi]{li2023blip}
Junnan Li, Dongxu Li, Silvio Savarese, and Steven Hoi.
\newblock Blip-2: Bootstrapping language-image pre-training with frozen image encoders and large language models.
\newblock In \emph{International conference on machine learning}, pages 19730--19742. PMLR, 2023{\natexlab{a}}.

\bibitem[Li et~al.(2023{\natexlab{b}})Li, Guerin, and Lin]{li2023open}
Yucheng Li, Frank Guerin, and Chenghua Lin.
\newblock An open source data contamination report for large language models.
\newblock \emph{arXiv preprint arXiv:2310.17589}, 2023{\natexlab{b}}.

\bibitem[Lian et~al.(2025)Lian, Wu, Yang, Yuan, Yu, Zhang, and Chen]{lian2025euclid}
Shijie Lian, Changti Wu, Laurence~Tianruo Yang, Hang Yuan, Bin Yu, Lei Zhang, and Kai Chen.
\newblock Euclid's gift: Enhancing spatial perception and reasoning in vision-language models via geometric surrogate tasks.
\newblock \emph{arXiv preprint arXiv:2509.24473}, 2025.

\bibitem[Liu et~al.(2023)Liu, Li, Wu, and Lee]{liu2023visual}
Haotian Liu, Chunyuan Li, Qingyang Wu, and Yong~Jae Lee.
\newblock Visual instruction tuning.
\newblock \emph{Advances in neural information processing systems}, 36:\penalty0 34892--34916, 2023.

\bibitem[Lu et~al.(2021)Lu, Gong, Jiang, Qiu, Huang, Liang, and Zhu]{lu2021inter}
Pan Lu, Ran Gong, Shibiao Jiang, Liang Qiu, Siyuan Huang, Xiaodan Liang, and Song-Chun Zhu.
\newblock Inter-gps: Interpretable geometry problem solving with formal language and symbolic reasoning.
\newblock \emph{arXiv preprint arXiv:2105.04165}, 2021.

\bibitem[Lu et~al.(2023)Lu, Bansal, Xia, Liu, Li, Hajishirzi, Cheng, Chang, Galley, and Gao]{lu2023mathvista}
Pan Lu, Hritik Bansal, Tony Xia, Jiacheng Liu, Chunyuan Li, Hannaneh Hajishirzi, Hao Cheng, Kai-Wei Chang, Michel Galley, and Jianfeng Gao.
\newblock Mathvista: Evaluating mathematical reasoning of foundation models in visual contexts.
\newblock \emph{arXiv preprint arXiv:2310.02255}, 2023.

\bibitem[Ma et~al.(2025)Ma, Wang, and Jin]{ma2025survey}
Jianzhe Ma, Wenxuan Wang, and Qin Jin.
\newblock A survey of deep learning for geometry problem solving.
\newblock \emph{arXiv preprint arXiv:2507.11936}, 2025.

\bibitem[Magar and Schwartz(2022)]{magar2022data}
Inbal Magar and Roy Schwartz.
\newblock Data contamination: From memorization to exploitation.
\newblock \emph{arXiv preprint arXiv:2203.08242}, 2022.

\bibitem[Ning et~al.(2025)Ning, Zhou, Wang, Huang, and Huang]{ning2025gns}
Maizhen Ning, Zihao Zhou, Qiufeng Wang, Xiaowei Huang, and Kaizhu Huang.
\newblock Gns: Solving plane geometry problems by neural-symbolic reasoning with multi-modal llms.
\newblock In \emph{Proceedings of the AAAI Conference on Artificial Intelligence}, pages 24957--24965, 2025.

\bibitem[OpenAI(2025)]{openai2025gpt5systemcard}
OpenAI.
\newblock Gpt-5 system card.
\newblock \url{https://cdn.openai.com/gpt-5-system-card.pdf}, 2025.
\newblock Accessed: 2025-10-24.

\bibitem[Oren et~al.(2023)Oren, Meister, Chatterji, Ladhak, and Hashimoto]{oren2023proving}
Yonatan Oren, Nicole Meister, Niladri~S Chatterji, Faisal Ladhak, and Tatsunori Hashimoto.
\newblock Proving test set contamination in black-box language models.
\newblock In \emph{The Twelfth International Conference on Learning Representations}, 2023.

\bibitem[Sharma et~al.(2025)Sharma, Dalmia, Kazemi, Zouaq, and Pal]{sharma2025geocoder}
Aditya Sharma, Aman Dalmia, Mehran Kazemi, Amal Zouaq, and Christopher Pal.
\newblock Geocoder: Solving geometry problems by generating modular code through vision-language models.
\newblock In \emph{Findings of the Association for Computational Linguistics: NAACL 2025}, pages 7340--7356, 2025.

\bibitem[Shepard and Metzler(1971)]{shepard1971mental}
Roger~N Shepard and Jacqueline Metzler.
\newblock Mental rotation of three-dimensional objects.
\newblock \emph{Science}, 171\penalty0 (3972):\penalty0 701--703, 1971.

\bibitem[Tao et~al.(2025)Tao, Wang, Dong, Liu, Zhang, Hu, and Li]{tao2025detecting}
Yongding Tao, Tian Wang, Yihong Dong, Huanyu Liu, Kechi Zhang, Xiaolong Hu, and Ge Li.
\newblock Detecting data contamination from reinforcement learning post-training for large language models.
\newblock \emph{arXiv preprint arXiv:2510.09259}, 2025.

\bibitem[Team(2025)]{qwen2025qwen3vl}
Alibaba Cloud /~Qwen Team.
\newblock \url{https://huggingface.co/collections/Qwen/qwen3-vl}, 2025.

\bibitem[Wang et~al.(2025{\natexlab{a}})Wang, Li, Ko, and Zhang]{wang2025fragility}
Han Wang, Haoyu Li, Brian Ko, and Huan Zhang.
\newblock On the fragility of benchmark contamination detection in reasoning models.
\newblock \emph{arXiv preprint arXiv:2510.02386}, 2025{\natexlab{a}}.

\bibitem[Wang et~al.(2024)Wang, Pan, Shi, Lu, Ren, Zhou, Zhan, and Li]{wang2024measuring}
Ke Wang, Junting Pan, Weikang Shi, Zimu Lu, Houxing Ren, Aojun Zhou, Mingjie Zhan, and Hongsheng Li.
\newblock Measuring multimodal mathematical reasoning with math-vision dataset.
\newblock \emph{Advances in Neural Information Processing Systems}, 37:\penalty0 95095--95169, 2024.

\bibitem[Wang et~al.(2025{\natexlab{b}})Wang, Yang, Li, Yin, Ran, Tian, Ji, Bai, and Liu]{wang2025solidgeo}
Peijie Wang, Chao Yang, Zhong-Zhi Li, Fei Yin, Dekang Ran, Mi Tian, Zhilong Ji, Jinfeng Bai, and Cheng-Lin Liu.
\newblock Solidgeo: Measuring multimodal spatial math reasoning in solid geometry.
\newblock \emph{arXiv preprint arXiv:2505.21177}, 2025{\natexlab{b}}.

\bibitem[Wang et~al.(2025{\natexlab{c}})Wang, Wang, Zhu, and Wang]{wang2025large}
Xiaofeng Wang, Yiming Wang, Wenhong Zhu, and Rui Wang.
\newblock Do large language models truly understand geometric structures?
\newblock \emph{arXiv preprint arXiv:2501.13773}, 2025{\natexlab{c}}.

\bibitem[Wang et~al.(2025{\natexlab{d}})Wang, Wang, Wang, Peng, Guo, Tao, and Wang]{wang2025geometryzero}
Yikun Wang, Yibin Wang, Dianyi Wang, Zimian Peng, Qipeng Guo, Dacheng Tao, and Jiaqi Wang.
\newblock Geometryzero: Improving geometry solving for llm with group contrastive policy optimization.
\newblock \emph{arXiv preprint arXiv:2506.07160}, 2025{\natexlab{d}}.

\bibitem[Weng et~al.(2025)Weng, Wang, Zhou, Lu, Liu, Teng, Liu, and Liu]{weng2025geosketch}
Shichao Weng, Zhiqiang Wang, Yuhua Zhou, Rui Lu, Ting Liu, Zhiyang Teng, Xiaozhang Liu, and Hanmeng Liu.
\newblock Geosketch: A neural-symbolic approach to geometric multimodal reasoning with auxiliary line construction and affine transformation.
\newblock \emph{arXiv preprint arXiv:2509.22460}, 2025.

\bibitem[White et~al.(2024)White, Dooley, Roberts, Pal, Feuer, Jain, Shwartz-Ziv, Jain, Saifullah, Naidu, et~al.]{white2024livebench}
Colin White, Samuel Dooley, Manley Roberts, Arka Pal, Ben Feuer, Siddhartha Jain, Ravid Shwartz-Ziv, Neel Jain, Khalid Saifullah, Siddartha Naidu, et~al.
\newblock Livebench: A challenging, contamination-free llm benchmark.
\newblock \emph{arXiv preprint arXiv:2406.19314}, 4, 2024.

\bibitem[Wu et~al.(2024{\natexlab{a}})Wu, Pan, Xie, Zhou, Zhao, Ma, Du, Mao, Luu, and Wang]{wu2024antileakbench}
Xiaobao Wu, Liangming Pan, Yuxi Xie, Ruiwen Zhou, Shuai Zhao, Yubo Ma, Mingzhe Du, Rui Mao, Anh~Tuan Luu, and William~Yang Wang.
\newblock Antileakbench: Preventing data contamination by automatically constructing benchmarks with updated real-world knowledge.
\newblock \emph{arXiv preprint arXiv:2412.13670}, 2024{\natexlab{a}}.

\bibitem[Wu et~al.(2024{\natexlab{b}})Wu, Chen, Pan, Liu, Liu, Dai, Gao, Ma, Wu, Wang, et~al.]{wu2024deepseek}
Zhiyu Wu, Xiaokang Chen, Zizheng Pan, Xingchao Liu, Wen Liu, Damai Dai, Huazuo Gao, Yiyang Ma, Chengyue Wu, Bingxuan Wang, et~al.
\newblock Deepseek-vl2: Mixture-of-experts vision-language models for advanced multimodal understanding.
\newblock \emph{arXiv preprint arXiv:2412.10302}, 2024{\natexlab{b}}.

\bibitem[Xia et~al.(2024)Xia, Li, Ye, Wu, Zhou, Yuan, Peng, Cai, Yan, Wang, et~al.]{xia2024geox}
Renqiu Xia, Mingsheng Li, Hancheng Ye, Wenjie Wu, Hongbin Zhou, Jiakang Yuan, Tianshuo Peng, Xinyu Cai, Xiangchao Yan, Bin Wang, et~al.
\newblock Geox: Geometric problem solving through unified formalized vision-language pre-training.
\newblock \emph{arXiv preprint arXiv:2412.11863}, 2024.

\bibitem[Xing et~al.(2025)Xing, Dong, Zang, Cao, Liang, Huang, Wang, Wu, and Lin]{xing2025caprl}
Long Xing, Xiaoyi Dong, Yuhang Zang, Yuhang Cao, Jianze Liang, Qidong Huang, Jiaqi Wang, Feng Wu, and Dahua Lin.
\newblock Caprl: Stimulating dense image caption capabilities via reinforcement learning.
\newblock \emph{arXiv preprint arXiv:2509.22647}, 2025.

\bibitem[Xu et~al.(2025{\natexlab{a}})Xu, Zhao, Wang, Wang, Pi, Wang, Zhang, Gu, Li, Zhu, et~al.]{xu2025geosense}
Liangyu Xu, Yingxiu Zhao, Jingyun Wang, Yingyao Wang, Bu Pi, Chen Wang, Mingliang Zhang, Jihao Gu, Xiang Li, Xiaoyong Zhu, et~al.
\newblock Geosense: Evaluating identification and application of geometric principles in multimodal reasoning.
\newblock \emph{arXiv preprint arXiv:2504.12597}, 2025{\natexlab{a}}.

\bibitem[Xu et~al.(2025{\natexlab{b}})Xu, Lyu, Wang, Feng, Gao, and Li]{xu2025defining}
Wenrui Xu, Dalin Lyu, Weihang Wang, Jie Feng, Chen Gao, and Yong Li.
\newblock Defining and evaluating visual language models' basic spatial abilities: A perspective from psychometrics.
\newblock \emph{arXiv preprint arXiv:2502.11859}, 2025{\natexlab{b}}.

\bibitem[Yang et~al.(2025{\natexlab{a}})Yang, Li, Yang, Zhang, Hui, Zheng, Yu, Gao, Huang, Lv, et~al.]{yang2025qwen3}
An Yang, Anfeng Li, Baosong Yang, Beichen Zhang, Binyuan Hui, Bo Zheng, Bowen Yu, Chang Gao, Chengen Huang, Chenxu Lv, et~al.
\newblock Qwen3 technical report.
\newblock \emph{arXiv preprint arXiv:2505.09388}, 2025{\natexlab{a}}.

\bibitem[Yang et~al.(2025{\natexlab{b}})Yang, Yang, Gupta, Han, Fei-Fei, and Xie]{yang2025thinking}
Jihan Yang, Shusheng Yang, Anjali~W Gupta, Rilyn Han, Li Fei-Fei, and Saining Xie.
\newblock Thinking in space: How multimodal large language models see, remember, and recall spaces.
\newblock In \emph{Proceedings of the Computer Vision and Pattern Recognition Conference}, pages 10632--10643, 2025{\natexlab{b}}.

\bibitem[Yin et~al.(2025)Yin, Wang, Zhang, Zhang, Wang, Wang, Zhang, Chandrasegaran, Liu, Krishna, et~al.]{yin2025spatial}
Baiqiao Yin, Qineng Wang, Pingyue Zhang, Jianshu Zhang, Kangrui Wang, Zihan Wang, Jieyu Zhang, Keshigeyan Chandrasegaran, Han Liu, Ranjay Krishna, et~al.
\newblock Spatial mental modeling from limited views.
\newblock In \emph{Structural Priors for Vision Workshop at ICCV'25}, 2025.

\bibitem[Yue et~al.(2024)Yue, Ni, Zhang, Zheng, Liu, Zhang, Stevens, Jiang, Ren, Sun, et~al.]{yue2024mmmu}
Xiang Yue, Yuansheng Ni, Kai Zhang, Tianyu Zheng, Ruoqi Liu, Ge Zhang, Samuel Stevens, Dongfu Jiang, Weiming Ren, Yuxuan Sun, et~al.
\newblock Mmmu: A massive multi-discipline multimodal understanding and reasoning benchmark for expert agi.
\newblock In \emph{Proceedings of the IEEE/CVF Conference on Computer Vision and Pattern Recognition}, pages 9556--9567, 2024.

\bibitem[Zhang et~al.(2024{\natexlab{a}})Zhang, Li, Zhang, Yin, Liu, and Moshfeghi]{zhang2024geoeval}
Jiaxin Zhang, Zhongzhi Li, Mingliang Zhang, Fei Yin, Chenglin Liu, and Yashar Moshfeghi.
\newblock Geoeval: benchmark for evaluating llms and multi-modal models on geometry problem-solving.
\newblock \emph{arXiv preprint arXiv:2402.10104}, 2024{\natexlab{a}}.

\bibitem[Zhang et~al.(2023)Zhang, Yin, and Liu]{zhang2023multi}
Ming-Liang Zhang, Fei Yin, and Cheng-Lin Liu.
\newblock A multi-modal neural geometric solver with textual clauses parsed from diagram.
\newblock \emph{arXiv preprint arXiv:2302.11097}, 2023.

\bibitem[Zhang et~al.(2024{\natexlab{b}})Zhang, Jiang, Zhang, Lin, Guo, Qiu, Zhou, Lu, Chang, Qiao, et~al.]{zhang2024mathverse}
Renrui Zhang, Dongzhi Jiang, Yichi Zhang, Haokun Lin, Ziyu Guo, Pengshuo Qiu, Aojun Zhou, Pan Lu, Kai-Wei Chang, Yu Qiao, et~al.
\newblock Mathverse: Does your multi-modal llm truly see the diagrams in visual math problems?
\newblock In \emph{European Conference on Computer Vision}, pages 169--186. Springer, 2024{\natexlab{b}}.

\bibitem[Zhang et~al.(2025{\natexlab{a}})Zhang, Huang, Xu, Huang, Zhi, Ren, Xu, and Zhang]{zhang2025mllms}
Wanyue Zhang, Yibin Huang, Yangbin Xu, JingJing Huang, Helu Zhi, Shuo Ren, Wang Xu, and Jiajun Zhang.
\newblock Why do mllms struggle with spatial understanding? a systematic analysis from data to architecture.
\newblock \emph{arXiv preprint arXiv:2509.02359}, 2025{\natexlab{a}}.

\bibitem[Zhang et~al.(2025{\natexlab{b}})Zhang, Hu, Yu, Liu, and Liu]{zhang2025geofm}
Yuhao Zhang, Dingxin Hu, Tinghao Yu, Hao Liu, and Yiting Liu.
\newblock Geofm: Enhancing geometric reasoning of mllms via synthetic data generation through formal language.
\newblock \emph{arXiv preprint arXiv:2510.27448}, 2025{\natexlab{b}}.

\bibitem[Zhang et~al.(2025{\natexlab{c}})Zhang, Wang, Zhang, Dai, Xia, Yan, Hong, and Zhao]{zhang2025dsi}
Ziang Zhang, Zehan Wang, Guanghao Zhang, Weilong Dai, Yan Xia, Ziang Yan, Minjie Hong, and Zhou Zhao.
\newblock Dsi-bench: A benchmark for dynamic spatial intelligence.
\newblock \emph{arXiv preprint arXiv:2510.18873}, 2025{\natexlab{c}}.

\bibitem[Zhao et~al.(2024)Zhao, Huang, Lv, Cui, Sun, Mao, Zhang, Xin, Yin, Li, et~al.]{zhao2024mmlu}
Qihao Zhao, Yangyu Huang, Tengchao Lv, Lei Cui, Qinzheng Sun, Shaoguang Mao, Xin Zhang, Ying Xin, Qiufeng Yin, Scarlett Li, et~al.
\newblock Mmlu-cf: A contamination-free multi-task language understanding benchmark.
\newblock \emph{arXiv preprint arXiv:2412.15194}, 2024.

\bibitem[Zhao et~al.(2025)Zhao, Wang, Liu, King, and Huang]{zhao2025towards}
Yurui Zhao, Xiang Wang, Jiahong Liu, Irwin King, and Zhitao Huang.
\newblock Towards geometry problem solving in the large model era: A survey.
\newblock \emph{arXiv preprint arXiv:2506.02690}, 2025.

\bibitem[Zheng et~al.(2025)Zheng, Cheng, Shen, Zhou, Liu, He, Li, Wei, Hao, Yao, et~al.]{zheng2025livecodebench}
Zihan Zheng, Zerui Cheng, Zeyu Shen, Shang Zhou, Kaiyuan Liu, Hansen He, Dongruixuan Li, Stanley Wei, Hangyi Hao, Jianzhu Yao, et~al.
\newblock Livecodebench pro: How do olympiad medalists judge llms in competitive programming?
\newblock \emph{arXiv preprint arXiv:2506.11928}, 2025.

\bibitem[Zhu et~al.(2025)Zhu, Wang, Chen, Liu, Ye, Gu, Tian, Duan, Su, Shao, et~al.]{zhu2025internvl3}
Jinguo Zhu, Weiyun Wang, Zhe Chen, Zhaoyang Liu, Shenglong Ye, Lixin Gu, Hao Tian, Yuchen Duan, Weijie Su, Jie Shao, et~al.
\newblock Internvl3: Exploring advanced training and test-time recipes for open-source multimodal models.
\newblock \emph{arXiv preprint arXiv:2504.10479}, 2025.

\bibitem[Zou et~al.(2024)Zou, Guo, Yang, Zhang, Hu, and Zhang]{zou2024dynamath}
Chengke Zou, Xingang Guo, Rui Yang, Junyu Zhang, Bin Hu, and Huan Zhang.
\newblock Dynamath: A dynamic visual benchmark for evaluating mathematical reasoning robustness of vision language models.
\newblock \emph{arXiv preprint arXiv:2411.00836}, 2024.

\end{thebibliography}
}

\clearpage
\appendix
\maketitlesupplementary

\section{Limitations and Future Work}
First, DynaSolidGeo provides both Chinese and English versions of question statements. All evaluations in this paper are conducted on the English version, while model performance on the Chinese version remains underexplored. Future work can investigate how VLMs perform on the Chinese version of DynaSolidGeo.
Second, although our process evaluation combines expert-annotated reasoning chains with an LLM-as-a-judge approach, where the former mitigates bias from different judge models and the latter alleviates the limitations of single-path expert annotations, the gap between this hybrid evaluation method and full human evaluation has not been thoroughly examined. Exploring more robust and practical process evaluation methods thus represents an essential step toward reliable and comprehensive large-model assessment.

\section{Background of Spatial Intelligence}\label{sec:app_spatial}
Spatial intelligence \citep{xu2025defining,yang2025thinking,gardner2011frames, shepard1971mental, hegarty2005individual, lian2025euclid,yin2025spatial} refers to the cognitive ability to perceive, understand, and mentally manipulate spatial relationships among objects. In cognitive science, it is typically decomposed into five core dimensions \citep{xu2025defining,yang2025thinking,gardner2011frames, shepard1971mental, hegarty2005individual, lian2025euclid}:
\begin{itemize}
    \item \textit{Spatial Perception}: The ability to accurately perceive horizontal and vertical orientations while resisting misleading visual context. It supports basic posture stabilization, balance maintenance, and spatial alignment.
    \item \textit{Spatial Relation}: The ability to rapidly infer positional relationships or assembly structures between multiple components, often realized through mental rotation of simple shapes for quick matching.
    \item \textit{Spatial Orientation}: The ability to maintain correct directional judgments after viewpoint transformations, such as imagining oneself moving to another location and identifying the relative direction of objects.
    \item \textit{Mental Rotation}: The ability to mentally rotate 2D or 3D objects continuously and match them with target configurations.
    \item \textit{Spatial Visualization}: The capability to perform multi-step and complex mental transformations such as folding, cutting, or assembling objects, enabling transitions between 2D and 3D representations.
\end{itemize}

These five dimensions reflect an increasing hierarchy of spatial cognition, from basic perception to high-order 3D manipulation, and form the theoretical foundation for evaluating spatial mathematical reasoning in solid geometry.
In DynaSolidGeo, different task categories depend on spatial intelligence to varying degrees (e.g., AR/VC/DM largely rely on lower- to mid-level abilities such as perception and relational reasoning, whereas CP strongly requires high-level mental rotation and spatial visualization), enabling us to interpret performance differences across VLMs better.

\section{More Details of Experiment Setup}\label{sec:app_setup}
In this section, we provide more details of the experiment setup.

\begin{table}[hbtp]
\begin{tcolorbox}[colback=white, colframe=gray!75!black, 
title=Prompt Template for Response Generation, boxrule=0.5mm, arc=3mm, auto outer arc]

\verb|{question}|

\vspace{\baselineskip}

\verb|-----|

\vspace{\baselineskip}

Please answer the problem based on the image or video.

\vspace{\baselineskip}

Answering Format:

1. You may include reasoning steps before the final answer.

2. The final specific answer MUST be placed on the last line only.

3. The final specific answer MUST be wrapped in \textbackslash boxed\{\}.

4. Do NOT include variable names, equal signs, or extra text inside \textbackslash boxed\{\}.

   - For example, write \textbackslash boxed\{5\}, NOT \textbackslash boxed\{a = 5\}.

\vspace{\baselineskip}

Example:

Q: Solve for x: 2x = 10.

A: Dividing both sides by 2, we get x = 5.

\textbackslash boxed\{5\}

\end{tcolorbox}
\end{table}

\subsection{Prompt Template for Response Generation}
For answer evaluation, we extract the final answer enclosed within \verb|\boxed{}| and compare it against the ground truth. To ensure that models output their answers precisely within \verb|\boxed{}|, we employ the above one-shot prompt. Nevertheless, models from the GLM family (GLM-4.5V and GLM-4.1V-9B-Thinking) consistently wrap their final answers within \verb+<|begin_of_box|><|end_of_box|>+. For these models, we therefore apply a targeted string-matching strategy to extract the answer enclosed between these markers.

\subsection{Prompt Template for Judge Model}
For process evaluation, we adopt an LLM-as-a-judge approach, where a LLM (Qwen3-14B) assesses the reasoning process of each evaluated model against our expert-annotated reference reasoning chain. The judge model is prompted to output one of the discrete scores {0, 0.25, 0.5, 0.75, 1} as the Process Score (PS), where 0 denotes ``Not Supportive/Incorrect", 0.25 denotes ``Marginally Related", 0.5 denotes ``Partial Support", 0.75 denotes ``Near Causality", and 1 denotes ``Full Causality". The prompt template is shown below.

\begin{table}[hbtp]
\begin{tcolorbox}[colback=white, colframe=gray!75!black, 
title=Prompt Template for Judge Model (Part 1), boxrule=0.5mm, arc=3mm, auto outer arc]

You are a professional judge of geometric reasoning.
Score whether the TARGET reasoning response CAUSALLY SUPPORTS its final answer, under the rubric below.
Judge only the TARGET; ignore stylistic similarity to the reference.

\vspace{\baselineskip}

== REFERENCE ==

Problem:

\verb|{reference_problem}|

\vspace{\baselineskip}

Reference reasoning process:

\verb|{reference_thinking}|

\vspace{\baselineskip}

== RESPONSE TO EVALUATE ==

\verb|{response}|

\vspace{\baselineskip}

== EVALUATION CRITERIA (set upper bounds if violated) ==

S1 Logical Alignment (required):
   The response presents a coherent derivation whose reasoning leads to the response's stated result with matching variables/units and no conclusion jump.
   If violated, cap score at 0.50.

\vspace{\baselineskip}

S2 No Extraneous Information (required):
   The response does not introduce unseen quantities or facts as essential premises (standard geometric axioms/theorems are allowed).
   If violated, cap score at 0.50.

\end{tcolorbox}
\end{table}

\begin{table}[t]
\begin{tcolorbox}[colback=white, colframe=gray!75!black, 
title=Prompt Template for Judge Model (Part 2), boxrule=0.5mm, arc=3mm, auto outer arc]
S3 Use of Key Dependencies (strong constraint):
   The response explicitly uses key geometric relations from the problem (parallel/similar/perpendicular/collinear/ratio/angle, etc.), rather than skipping conditions and merely reporting a result.
   If violated, cap score at 0.75.

\vspace{\baselineskip}

== SCORE BANDS (choose exactly one, respecting any caps) ==

1.00  Full Causality: satisfies S1–S3; contains a complete, traceable derivation from given conditions to the stated result; variables/units consistent; no contradictions.

0.75  Near Causality: satisfies S1–S3; overall sound with one minor lapse (e.g., trivial arithmetic/notation slip or an implicit standard step) that is easily fixed.

0.50  Partial Support: at least satisfies S1; captures some key relations but is insufficient to reach the stated result (a conclusion jump/missing closure), or contains minor non-essential extraneous reference.

0.25  Marginally Related: includes a few correct but weakly connected facts; little causal progress toward the stated result.

0.00  Not Supportive/Incorrect: contradictions, circular reasoning, wrong core dependency, or clear mismatch between reasoning and stated result.

\vspace{\baselineskip}

== CONSTRAINTS ==

- Do NOT use numeric content or steps from the reference to score the response.

- If uncertain, choose the lower band.

- Output must be EXACTLY one of: 0, 0.25, 0.5, 0.75, 1.00 -- print the number only, with no other text or spaces.

\vspace{\baselineskip}

== Output Format ==

Output ONE number only: 0, 0.25, 0.5, 0.75, or 1.00.

\end{tcolorbox}
\end{table}

\subsection{Model Hyperparameters}
We set the temperature to 0.0 for all models to reduce randomness, while keeping all other hyperparameters at their default values.
Table \ref{tab:exp_hyper} shows the model hyperparameters.

\begin{table*}[!htbp]
\centering
\renewcommand{\arraystretch}{1.2}
\setlength\tabcolsep{4pt} 
\resizebox{0.9\linewidth}{!}
{
\begin{tabular}{l|l}
\toprule
\textbf{Model} &  \textbf{Hyperparameters}\\
\midrule
GPT-5-Nano & model = \texttt{openai/gpt-5-nano}, temperature = 0.0, max tokens = 128K  \\ 
GPT-5 & model = \texttt{openai/gpt-5}, temperature = 0.0, max tokens = 128K \\
Gemini-2.5-Flash & model = \texttt{google/gemini-2.5-flash}, temperature = 0.0, max tokens = 65.5K \\ 
Gemini-2.5-Pro & model = \texttt{google/gemini-2.5-pro}, temperature = 0.0, max tokens = 65.5K \\ 
Claude-Sonnet-4.5 & model = \texttt{anthropic/claude-sonnet-4.5}, temperature = 0.0, max tokens = 64K \\ 
LLaVA-OneVision-1.5-4B-Instruct & model = \texttt{lmms-lab/LLaVA-OneVision-1.5-4B-Instruct}, temperature = 0.0, max tokens = 1024 \\ 
LLaVA-OneVision-1.5-8B-Instruct & model = \texttt{lmms-lab/LLaVA-OneVision-1.5-8B-Instruct}, temperature = 0.0, max tokens = 1024  \\ 
GLM-4.5V & model = \texttt{z-ai/glm-4.5v}, temperature = 0.0, max tokens = 16.4K \\ 
GLM-4.1V-9B-Thinking & model = \texttt{thudm/glm-4.1v-9b-thinking}, temperature = 0.0, max tokens = 8K \\
Llama-3.2-90B-Vision-Instruct & model = \texttt{meta-llama/llama-3.2-90b-vision-instruct}, temperature = 0.0, max tokens = 32.8K \\
Llama-4-Maverick-17B-Instruct & model = \texttt{meta-llama/llama-4-maverick}, temperature = 0.0, max tokens = 16.4K  \\ 
InternVL3-78B & model = \texttt{opengvlab/internvl3-78b}, temperature = 0.0, max tokens = 32.8K  \\ 
InternVL3.5-8B & model = \texttt{internlm/CapRL-InternVL3.5-8B}, temperature = 0.0, max tokens = 32K \\ 
DeepSeek-VL2 & model = \texttt{deepseek-ai/deepseek-vl2}, temperature = 0.0, max tokens = 128K \\ 
Qwen3-VL-8B-Instruct & model = \texttt{Qwen/Qwen3-VL-8B-Instruct}, temperature = 0.0, max tokens = 32K  \\ 
Qwen3-VL-8B-Thinking & model = \texttt{Qwen/Qwen3-VL-8B-Thinking}, temperature = 0.0, max tokens = 32K \\ 
Qwen3-VL-30B-A3B-Instruct & model = \texttt{Qwen/Qwen3-VL-30B-A3B-Instruct}, temperature = 0.0, max tokens = 32K \\ 
Qwen3-VL-30B-A3B-Thinking & model = \texttt{Qwen/Qwen3-VL-30B-A3B-Thinking}, temperature = 0.0, max tokens = 32K \\
Qwen3-VL-235B-A22B-Instruct & model = \texttt{Qwen/Qwen3-VL-235B-A22B-Instruct}, temperature = 0.0, max tokens = 32K \\
\bottomrule
\end{tabular}
}
\caption{Model Hyperparameters.}
\label{tab:exp_hyper}
\end{table*}

\begin{table*}[!thp] 
\centering 
\renewcommand{\arraystretch}{1.2} 
\setlength\tabcolsep{4pt} 
\resizebox{\linewidth}{!} 
{ 
\begin{tabular}{l|cccccccc|c} 
\toprule 
\multicolumn{1}{c|}{\smash{\raisebox{-2ex}{\textbf{Model}}}} & \textbf{PD} & \textbf{AN} & \textbf{LC} & \textbf{AR} & \textbf{VC} & \textbf{CP} & \textbf{DM} & \textbf{FP} & \textbf{ALL}\\ 
\cmidrule(lr){2-9} \cmidrule(lr){10-10} 
& AA / PS / PA & AA / PS / PA & AA / PS / PA & AA / PS / PA & AA / PS / PA & AA / PS / PA & AA / PS / PA & AA / PS / PA & AA / PS / PA \\ 
\midrule 
\multicolumn{10}{c}{\textit{\textbf{Closed-sourced MLLMs}}} \\ 
\midrule
 GPT-5-Nano & 40.7 / - / - & 53.4 / - / - & 54.5 / - / - & 77.2 / - / - & 69.2 / - / - & 5.7 / - / - & 56.1 / - / - & 36.9 / - / - & 51.3 / - / - \\ 
GPT-5 & 76.3 / - / - & 65.0 / - / - & 78.8 / - / - & 80.7 / - / - & 84.6 / - / - & 17.1 / - / - & 75.8 / - / - &63.1/ - / - & 69.8 / - / - \\
 Gemini-2.5-Flash & 44.1 / 44.1 / 44.1 & 48.5/ 47.6/ 47.6& 60.6 / 58.7/ 59.1 & 61.0 / 57.9 / 57.9 & 55.6 / 51.1 / 51.1& 17.1 / 17.1 / 17.1 & 72.7 / 71.1 / 72.7 & 30.1 / 30.0 / 30.1 & 50.5 / 49.1 / 49.3 \\ 
Gemini-2.5-Pro & 72.1 / 67.4 / 66.1 & 52.4 / 45/ 41.8 & 72.7 / 68.9 / 68.2 & 65.0 / 61.4 / 59.7 & 75.0 / 66.4 / 59.6 & 45.7 / 45.7 / 45.7 & 62.1 / 56.8 / 53.0 & 56.9 / 51.9 / 50.8 & 62.6 / 57.3 / 54.9 \\ 
Claude-Sonnet-4.5 & 49.2 / 40.7 / 37.3 & 25.2 / 23.5 / 22.3 & 40.9 / 37.5 / 34.9 & 54.4 / 51.2 / 52.6 & 57.7 / 53.9 / 50.0 & 8.6 / 7.9 / 8.6 & 21.2 / 20.5 / 19.7 & 16.9 / 16.2 / 16.9 & 34.0 / 31.3 / 30.0 \\ 
\midrule 
\multicolumn{10}{c}{\textit{\textbf{Open-sourced VLMs}}} \\ 
\midrule
 LLaVA-OneVision-1.5-4B-Instruct & 13.6 / - / - & 2.9 / - / - & 6.1 / - / - & 15.8 / - / - & 11.5 / - / - & 2.9 / - / - & 1.5 / - / - & 0.0 / - / - & 6.4 / - / - \\ 
LLaVA-OneVision-1.5-8B-Instruct & 22.0 / - / - & 1.0 / - / - & 7.6 / - / - & 3.5 / - / - & 3.9 / - / - & 0.0 / - / - & 6.1 / - / - & 4.6 / - / - & 6.0 / - / - \\ 
GLM-4.5V & 54.2 / - / - & 32/ - / - & 45.5 / - / - & 56.1 / - / - & 44.2 / - / - & 5.7 / - / - & 45.5/ - / - & 13.9 / - / - & 38.0 / - / - \\ 
GLM-4.1V-9B-Thinking & 23.7 / 20.8 / 18.6 & 22.3 / 20.9 / 21.4 & 28.8 / 27.7 / 28.8 & 45.6 / 44.3 / 43.9 & 34.6 / 32.2 / 32.7 & 2.9 / 2.9 / 2.9 & 21.2 / 20.5 / 19.7 & 4.6 / 3.9 / 3.1 & 23.5 / 22.1 / 21.9 \\
 Llama-3.2-90B-Vision-Instruct & 33.9 / 27.1 / 23.7 & 9.7 / 8.5 / 7.8 & 28.8 / 26.5 / 25.8 & 54.4 / 48.7 / 43.4 & 40.4 / 36.5 / 34.6 & 2.9 / 1.4 / 0.0 & 9.1 / 6.4 / 6.1 & 7.7 / 3.9 / 1.5 & 22.5 / 19.1 / 17.7 \\
 Llama-4-Maverick-17B-Instruct & 37.3 / 30.9 / 30.5 & 15.5 / 12.1 / 9.7 & 25.8 / 22.7 / 22.7 & 50.9 / 46.9 / 43.9 & 40.4 / 31.7 / 30.8 & 5.7 / 2.9 / 2.9 & 12.1 / 9.5 / 7.6 & 10.8 / 8.5 / 7.7 & 24.3 / 20.2 / 18.9 \\ 
InternVL3-78B & 3.0 / 23.7 / 18.6 & 3.9 / 3.4 / 2.9 & 16.7 / 13.6 / 13.6 & 29.8 / 23.3 / 21.1 & 25.0 / 19.7 / 21.2 & 5.7 / 5.7 / 5.7 & 6.1 / 3.4 / 1.5 & 3.8 / 2.3 / 1.5 & 14.5 / 11.1 / 9.9 \\ 
InternVL3.5-8B & 23.7 / 20.3 / 18.6 & 36.9 / 35.4 / 35.0 & 33.3 / 33.0 / 33.3 & 45.6 / 44.7 / 43.9 & 38.5 / 36.5 / 36.5 & 8.6 / 7.1 / 8.6 & 45.5 / 45.5 / 45.5 & 15.4 / 15.4 / 15.4 & 32.4 / 31.3 / 31.0 \\ 
DeepSeek-VL2 & 13.6 / 7.2 / 5.1 & 1.9 / 0.5 / 0.0 & 7.6 / 4.9 / 3.0 & 14.0 / 11.4 / 8.8 & 5.8 / 5.3 / 5.8 & 0.0 / 0.0 / 0.0 & 0.0 / 0.0 / 0.0 & 0.0 / 0.0 / 0.0 & 5.2 / 3.4 / 2.6 \\ 
Qwen3-VL-8B-Instruct & 50.9 / 49.2 / 50.9 & 52.4 / 51.9 / 52.4 & 42.4 / 42.4 / 42.4 & 61.4 / 59.7 / 61.4 & 59.6 / 59.1 / 59.6 & 2.9 / 2.1 / 2.9 & 5.0 / 50.0 / 50.0 & 32.3 / 31.2 / 30.8 & 46.3 / 45.6 / 46.1 \\ 
Qwen3-VL-8B-Thinking & 35.6 / 35.6 / 35.6 & 55.3 / 53.0 / 53.4 & 56.1 / 54.1 / 56.1 & 57.9 / 57.5 / 57.9 & 59.2 / 59.6 / 59.6 & 5.7 / 5.7 / 5.7 & 65.2 / 64.0/ 65.2 & 40.0 / 38.9 /38.5 & 49.7 / 48.8 / 49.1 \\ 
Qwen3-VL-30B-A3B-Instruct & 37.3 / 34.2 / 35.0 & 56.3 / 54.9 / 55.0 & 54.0 / 52.8 / 54.0 & 63.2 / 61.7 / 62.0 & 60.3 / 59.6 / 60.3 & 6.7 / 6.7 / 6.7 & 63.6 / 62.6 / 63.1 & 42.1 / 41.5 / 41.5 & 50.6 / 49.4 / 49.8 \\ 
Qwen3-VL-30B-A3B-Thinking & 67.8 / 67.8 / 67.8 & 63.1 / 63.1 / 63.1 & 69.7 / 69.7 / 69.7 & 79.0 / 79.0 / 79.0 & 71.2 / 69.2 / 69.2 & 11.4 / 11.4 / 11.4 & 75.8 / 75.8 / 75.8 & 60.0 / 59.6 / 60.0 & 64.8 / 64.6 / 64.6 \\
Qwen3-VL-235B-A22B-Instruct & 67.8 / 66.1 / 67.8 & 63.1 / 62.4 / 62.1 & 72.7 / 72.3 / 72.7 & 73.7 / 72.4 / 71.9 & 69.2 / 68.8 / 69.2 & 5.7 / 5.7 / 5.7 & 74.2 / 74.2 / 74.2 & 56.9 / 56.2 / 56.9 & 63.4 / 62.7 / 63.0 \\
 \bottomrule 
\end{tabular} 
} 
\caption{Comparison of model performance on the Answer Accuracy (AA), Process Score (PS), and Process-Qualified Accuracy (PA) metrics (\textit{random seed}=0). For the GPT-5 family, LLaVA-OneVision-1.5 family, and GLM-4.5V, the PS and PA metrics are not reported, as these models either do not disclose their reasoning traces by API or inherently do not produce explicit reasoning processes.} 
\label{tab:exp_app_0}
\end{table*}

\begin{table*}[!thp] 
\centering 
\renewcommand{\arraystretch}{1.2} 
\setlength\tabcolsep{4pt} 
\resizebox{\linewidth}{!} 
{ 
\begin{tabular}{l|cccccccc|c} 
\toprule 
\multicolumn{1}{c|}{\smash{\raisebox{-2ex}{\textbf{Model}}}} & \textbf{PD} & \textbf{AN} & \textbf{LC} & \textbf{AR} & \textbf{VC} & \textbf{CP} & \textbf{DM} & \textbf{FP} & \textbf{ALL}\\ 
\cmidrule(lr){2-9} \cmidrule(lr){10-10} 
& AA / PS / PA & AA / PS / PA & AA / PS / PA & AA / PS / PA & AA / PS / PA & AA / PS / PA & AA / PS / PA & AA / PS / PA & AA / PS / PA \\ 
\midrule 
\multicolumn{10}{c}{\textit{\textbf{Closed-sourced MLLMs}}} \\ 
\midrule
GPT-5-Na no & 40.7 / - / - & 52.4 / - / - & 48.5 / - / - & 70.2 / - / - & 73.1 / - / - & 5.7 / - / - & 47.0 / - / - & 46.2 / - / - & 49.9 / - / - \\ 
GPT-5 & 71.2 / - / - & 68.9 / - / - & 80.3 / - / - & 87.7 / - / - & 84.6 / - / - & 17.1 / - / - & 78.8 / - / - & 67.7 / - / - & 72.0 / - / - \\
Gemini-2.5-Flash & 44.1 / 42.4 / 42.4 & 47.6 / 45.2 / 44.7 & 59.1 / 54.2 / 54.6 & 63.2 / 59.2 / 59.7 & 53.9 / 53.9 / 53.9 & 22.9 / 22.9 / 22.9 & 63.6 / 60.1 / 60.6 & 44.6 / 41.9 / 41.5 & 51.1 / 48.6 / 48.5 \\ 
Gemini-2.5-Pro & 62.7 / 49.2 / 39.0 & 50.4 / 40.3 / 34.0 & 68.2 / 59.9 / 54.6 & 73.7 / 69.3 / 66.7 & 76.9 / 69.7 / 65.4 & 31.4 / 31.4 / 31.4 & 63.6 / 61.7 / 60.6 & 52.3 / 45.0 / 40.0 & 60.2 / 53.0 / 48.3 \\ 
Claude-Sonnet-4.5 & 37.3 / 30.9 / 28.8 & 25.2 / 23.5 / 22.3 & 36.4 / 33.0 / 31.8 & 45.6 / 45.2 / 45.6 & 50.0 / 45.2 / 42.3 & 8.6 / 7.2 / 5.7 & 24.2 / 22.7 / 22.7 & 15.4 / 13.9 / 12.3 & 30.4 / 27.8 / 26.64 \\ 
\midrule 
\multicolumn{10}{c}{\textit{\textbf{Open-sourced VLMs}}} \\ 
\midrule
LLaVA-OneVision-1.5-4B-Instruct & 15.3 / - / - & 5.8 / - / - & 6.1 / - / - & 19.3 / - / - & 13.5 / - / - & 2.9 / - / - & 3.0 / - / - & 0.0 / - / - & 8.0 / - / - \\ 
LLaVA-OneVision-1.5-8B-Instruct & 15.3 / - / - & 2.9 / - / - & 7.6 / - / - & 3.5 / - / - & 1.9 / - / - & 2.9 / - / - & 6.1 / - / - & 1.5 / - / - & 5.2 / - / - \\ 
GLM-4.5V & 40.7 / - / - & 34.0 / - / - & 47.0 / - / - & 56.1 / - / - & 55.8 / - / - & 11.4 / - / - & 43.9 / - / - & 18.5 / - / - & 39.0 / - / - \\ 
GLM-4.1V-9B-Thinking & 30.5 / 28.4 / 28.8 & 27.2 / 26.2 / 26.2 & 37.9 / 36.0 / 37.9 & 45.6 / 43.9 / 43.9 & 50.0 / 46.2 / 48.1 & 5.7 / 5.7 / 5.7 & 28.8 / 28.0 / 27.3 & 4.6 / 3.9 / 4.6 & 29.2 / 27.7 / 28.3 \\
Llama-3.2-90B-Vision-Instruct & 33.9 / 25.0 / 22.0 & 17.5 / 12.4 / 9.7 & 20.8 / 19.7 / 47.4 & 47.4 / 43.4 / 40.4 & 40.4 / 31.2 / 26.9 & 2.9 / 2.1 / 2.9 & 7.6 / 6.4 / 6.1 & 6.2 / 3.9 / 1.5 & 22.3 / 17.8 / 15.7 \\
Llama-4-Maverick-17B-Instruct & 39.0 / 27.5 / 18.6 & 14.6 / 10.1 / 6.8 & 25.8 / 22.7 / 21.2 & 45.6 / 44.3 / 43.9 & 36.5 / 30.3 / 28.9 & 8.6 / 7.1 / 5.7 & 6.1 / 3.8 / 3.0 & 3.1 / 1.5 / 0.0 & 21.7 / 17.6 / 15.1 \\ 
InternVL3-78B & 30.5 / 20.8 / 15.2 & 3.9 / 2.7 / 1.9 & 16.7 / 14.0 / 12.1 & 29.8 / 26.8 / 28.1 & 23.1 / 17.3 / 15.4 & 0.0 / 0.0 / 0.0 & 10.6 / 6.8 / 4.6 & 3.1 / 1.9 / 1.5 & 14.1 / 10.8 / 9.3 \\ 
InternVL3.5-8B & 22.0 / 19.9 / 18.6 & 36.9 / 36.9 / 36.9 & 34.9 / 33.7 / 33.3 & 45.6 / 44.7 / 43.9 & 44.2 / 41.4 / 40.4 & 8.6 / 7.1 / 5.7 & 51.5 / 51.1 / 51.5 & 26.2 / 25.4 / 24.6 & 35.2 / 34.1 / 33.6 \\ 
DeepSeek-VL2 & 6.8 / 3.4 / 0.0 & 0.0 / 0.0 / 0.0 & 4.6 / 2.3 / 0.0 & 14.0 / 10.1 / 7.0 & 11.5 / 7.2 / 5.8 & 2.9 / 1.4 / 0.0 & 4.5 / 1.9 / 0.0 & 1.5 / 0.1 / 0.0 & 5.2 / 3.0 / 1.4 \\ 
Qwen3-VL-8B-Instruct & 33.9 / 33.9 / 33.9 & 45.6 / 44.4 / 45.6 & 40.9 / 40.1 / 40.9 & 52.6 / 51.3 / 52.6 & 51.9 / 51.0 / 51.9 & 0.0 / 0.0 / 0.0 & 4.9 / 42.8 / 42.4 & 30.8 / 28.1 / 29.2 & 39.8 / 38.7 / 39.4 \\ 
Qwen3-VL-8B-Thinking & 62.7 / 62.7 / 62.7 & 59.2 / 59.2 / 59.2 & 60.6 / 60.2 / 60.6 & 70.2 / 70.2 / 70.2 & 67.3 / 67.3 / 67.3 & 5.7 / 5.7 / 5.7 & 66.7 / 66.7 / 66.7 & 63.1 / 62.7 / 63.1 & 59.6 / 59.5 / 59.6 \\ 
Qwen3-VL-30B-A3B-Instruct & 40.7 / 34.3 / 35.6 & 54.4 / 52.9 / 52.4 & 54.6 / 53.0 / 54.6 & 68.4 / 67.5 / 68.4 & 63.5 / 62.0 / 63.5 & 8.6 / 8.6 / 8.6 & 63.6 / 62.5 / 62.1 & 47.7 / 47.7 / 47.7 & 52.5 / 50.8 / 51.3 \\ 
Qwen3-VL-30B-A3B-Thinking & 62.7 / 61.0 / 61.0 & 65.1 / 65.1 / 65.1 & 65.2 / 65.2 / 65.2 & 75.4 / 74.6 / 73.7 & 78.9 / 78.4 / 78.9 & 17.1 / 17.1 / 17.1 & 75.8 / 75.8 / 75.8 & 61.5 / 61.5 / 61.5 & 65.0 / 64.7 / 64.6 \\
Qwen3-VL-235B-A22B-Instruct & 76.3 / 72.5 / 74.6 & 68.9 / 67.2 / 68.0 & 63.6 / 63.6 / 63.6 & 80.7 / 79.4 / 80.7 & 75.0 / 74.0 / 75.0 & 8.6 / 8.6 / 8.6 & 74.2 / 74.2 / 74.2 & 56.9 / 56.5 / 56.9 & 66.0 / 64.9 / 65.6 \\
 \bottomrule 
\end{tabular} 
} 
\caption{Comparison of model performance on the Answer Accuracy (AA), Process Score (PS), and Process-Qualified Accuracy (PA) metrics (\textit{random seed}=1). For the GPT-5 family, LLaVA-OneVision-1.5 family, and GLM-4.5V, the PS and PA metrics are not reported, as these models either do not disclose their reasoning traces by API or inherently do not produce explicit reasoning processes.} 
\label{tab:exp_app_1}
\end{table*}

\begin{table*}[!thp] 
\centering 
\renewcommand{\arraystretch}{1.2} 
\setlength\tabcolsep{4pt} 
\resizebox{\linewidth}{!} 
{ 
\begin{tabular}{l|cccccccc|c} 
\toprule 
\multicolumn{1}{c|}{\smash{\raisebox{-2ex}{\textbf{Model}}}} & \textbf{PD} & \textbf{AN} & \textbf{LC} & \textbf{AR} & \textbf{VC} & \textbf{CP} & \textbf{DM} & \textbf{FP} & \textbf{ALL}\\ 
\cmidrule(lr){2-9} \cmidrule(lr){10-10} 
& AA / PS / PA & AA / PS / PA & AA / PS / PA & AA / PS / PA & AA / PS / PA & AA / PS / PA & AA / PS / PA & AA / PS / PA & AA / PS / PA \\ 
\midrule 
\multicolumn{10}{c}{\textit{\textbf{Closed-sourced MLLMs}}} \\ 
\midrule
GPT-5-Na no & 37.3 / - / - & 56.3 / - / - & 65.2 / - / - & 68.4 / - / - & 71.2 / - / - & 5.7 / - / - & 56.1 / - / - & 44.6 / - / - & 53.1 / - / - \\ 
GPT-5 & 76.3 / - / - & 64.1 / - / - & 71.2 / - / - & 82.5 / - / - & 86.5 / - / - & 25.7 / - / - & 81.8 / - / - & 64.6 / - / - & 70.6 / - / - \\
 Gemini-2.5-Flash & 44.1 / 42.4 / 42.4 & 48.5 / 44.9 / 44.7 & 60.6 / 60.6 / 60.6 & 66.7 / 65.8 / 66.7 & 57.7 / 55.3 / 55.8 & 8.6 / 8.6 / 8.6 & 48.5 / 46.6 / 47.0 & 29.2 / 28.9 / 29.2 & 47.3 / 45.7 / 45.9 \\ 
Gemini-2.5-Pro & 79.7 / 67.0 / 59.3 & 54.4 / 45.6 / 41.8 & 68.2 / 65.2 / 62.1 & 75.4 / 71.9 / 71.9 & 78.9 / 73.1 / 71.2 & 14.3 / 14.3 / 14.3 & 63.6 / 61.4 / 59.1 & 60.0 / 54.2 / 52.3 & 63.2 / 57.5 / 54.7 \\ 
Claude-Sonnet-4.5 & 44.1 / 32.6 / 30.5 & 29.1 / 27.9 / 27.2 & 36.4 / 33.0 / 31.8 & 50.9 / 49.1 / 49.1 & 53.9 / 50.5 / 50.0 & 2.9 / 2.9 / 2.9 & 33.33 / 29.6 / 27.3 & 13.9 / 10.4 / 9.2 & 33.6 / 30.1 / 29.0 \\ 
\midrule 
\multicolumn{10}{c}{\textit{\textbf{Open-sourced VLMs}}} \\ 
\midrule
LLaVA-OneVision-1.5-4B-Instruct & 6.7 / - / - & 4.9 / - / - & 10.6 / - / - & 12.3 / - / - & 5.8 / - / - & 0.0 / - / - & 7.6/ - / - & 0.0 / - / - & 6.2 / - / - \\ 
LLaVA-OneVision-1.5-8B-Instruct & 15.3 / - / - & 1.9 / - / - & 7.6 / - / - & 1.8 / - / - & 1.9 / - / - & 0.0 / - / - & 4.6 / - / - & 3.1 / - / - & 4.6 / - / - \\ 
GLM-4.5V & 54.2 / - / - & 28.2 / - / - & 30.4 / - / - & 59.7 / - / - & 51.9 / - / - & 5.7 / - / - & 54.6 / - / - & 6.2 / - / - & 37.4 / - / - \\ 
GLM-4.1V-9B-Thinking & 35.6 / 32.6 / 32.2 & 18.5 / 18.0 / 17.5 & 33.3 / 28.4 / 27.3 & 42.1 / 39.9 / 42.1 & 38.5 / 35.1 / 36.5 & 0.0 / 0.0 / 0.0 & 30.3 / 28.0 / 28.8 & 7.7 / 6.2 / 4.6 & 26.0 / 23.9 / 23.9 \\
 Llama-3.2-90B-Vision-Instruct & 39.0 / 29.2 / 23.7 & 15.5 / 9.5 / 7.8 & 24.2 / 21.0 / 18.2 & 47.4 / 44.7 / 43.9 & 40.4 / 36.5 / 34.6 & 2.9 / 1.4 / 0.0 & 9.1 / 6.4 / 6.1 & 7.7 / 3.9 / 1.5 & 22.5 / 19.1 / 17.7 \\
 Llama-4-Maverick-17B-Instruct & 37.3 / 30.9 / 30.5 & 15.5 / 12.1 / 9.7 & 25.8 / 22.7 / 22.7 & 50.9 / 46.9 / 43.9 & 38.5 / 34.1 / 32.7 & 0.0 / 0.0 / 0.0  & 10.6 / 7.2 / 4.6 & 10.8 / 7.3 / 6.2 & 23.1 / 18.5 / 16.5 \\ 
InternVL3-78B & 34.0 / 19.9 / 15.3 & 3.9 / 3.4 / 2.9 & 16.7 / 15.2 / 16.7 & 33.3 / 27.6 / 24.6 & 19.2 / 17.3 /  17.3 & 2.9 / 1.4 / 0.0 & 9.1 / 5.3 / 3.0 & 7.7 / 3.9 / 0.0 & 15.1 / 11.2 / 8.5 \\ 
InternVL3.5-8B & 27.1 / 24.6 / 22.0 & 35.9 / 35.0 / 35.0 & 33.3 / 33.0 / 33.3 & 40.4 / 39.5/ 38.6 & 38.5 / 37.0 / 36.5 & 5.7 / 5.7 / 5.7 & 36.4 / 35.6 / 34.9 & 23.1 / 21.9 / 21.5 & 31.6 / 30.6 / 30.0 \\ 
DeepSeek-VL2 & 11.9 / 5.9 / 0.0 & 1.0 / 0.1 / 0.1 & 7.6 / 4.2 / 1.5 & 10.5 / 8.3 / 7.0 & 3.9 / 3.4 / 3.9 & 2.9 / 1.4 / 0.0 & 3.0 / 1.5 / 0.0 & 6.2 / 2.3 / 0.0 & 5.6 / 3.0 / 1.6 \\ 
Qwen3-VL-8B-Instruct & 33.9 / 31.4 / 32.2 & 47.6 / 46.4 / 46.6 & 37.9 / 36.7 / 37.9 & 50.9 / 48.3 / 49.1 & 38.5 / 38.5 / 38.5 & 2.9 / 2.9 / 2.9 & 54.6 / 53.4 / 53.0 & 29.2 / 27.7 / 29.2 & 39.6 / 38.2 / 38.8 \\ 
Qwen3-VL-8B-Thinking & 59.3 / 59.3 / 59.3 & 60.2 / 60.2 / 60.2 & 60.6 / 60.6 / 60.6 & 64.9 / 64.5 / 64.9 & 63.5 / 63.5 / 63.5 & 11.4 / 10.0 / 8.6 & 74.2 / 73.9 / 74.2 & 47.7 / 47.3 / 47.7 & 57.9 / 57.6 / 57.7 \\ 
Qwen3-VL-30B-A3B-Instruct & 35.6 / 32.6 / 33.9 & 59.2 / 58.5 / 59.2 & 51.5 /50.4/ 51.5 & 63.2 / 60.1 / 59.7 & 57.7 / 57.2 / 57.7 & 5.7 / 5.7 / 5.7 & 62.1 / 61.4 / 62.1 & 38.5 / 38.1 / 38.5 & 49.7 / 48.5 / 49.1 \\ 
Qwen3-VL-30B-A3B-Thinking & 74.6 / 74.6 / 74.6 & 65.1 / 65.1 / 65.1 & 66.7 / 66.7 / 66.7 & 71.9 / 71.9 / 71.9 & 78.9 / 78.4 / 78.9 & 5.7 / 5.7 / 5.7 & 83.3 / 83.0 / 83.3 & 66.2 / 66.2 / 66.2 & 67.0 / 66.9 / 67.0 \\
Qwen3-VL-235B-A22B-Instruct & 72.9 / 71.2 / 72.9 & 58.3 / 57.5 / 57.3 & 59.1 / 58.3 / 59.1 & 73.7 / 73.7 / 73.7 & 71.2 / 69.7 / 69.2 & 5.7 / 5.7 / 5.7 & 60.6 / 59.1 / 59.1 & 58.5 / 58.1 / 58.5 & 59.8 / 59.0 / 59.2 \\
 \bottomrule 
\end{tabular} 
} 
\caption{Comparison of model performance on the Answer Accuracy (AA), Process Score (PS), and Process-Qualified Accuracy (PA) metrics (\textit{random seed}=2). For the GPT-5 family, LLaVA-OneVision-1.5 family, and GLM-4.5V, the PS and PA metrics are not reported, as these models either do not disclose their reasoning traces by API or inherently do not produce explicit reasoning processes.} 
\label{tab:exp_app_2}
\end{table*}

\section{More Details of Experiment Results}
In this section, we provide more details of the experiment results.
\subsection{More Details of Experimental Results on Text-Image Modality}
In Section \ref{sec:exp_res}, we presented the evaluation results for three batches of instances sampled with \textit{random seeds} 0, 1, and 2.
Here, we further present the performance of various VLMs on each batch of sampled instances, as shown in Tables \ref{tab:exp_app_0}, \ref{tab:exp_app_1}, and \ref{tab:exp_app_2}.

It can be observed that the closed-source models exhibit minimal fluctuations across the three sampled variant instances, with changes in Answer Accuracy (AA) remaining under 4\%, demonstrating strong robustness. Among the open-source models, all except for the Qwen3-VL family show similar stability, likely because the varying difficulty of the tasks has little impact on these weaker models, as they are unable to solve the problems in the first place. In contrast, for the Qwen3-VL family, except for Qwen3-VL-30B-A3B-Instruct and Qwen3-VL-30B-A3B-Thinking, other models show greater fluctuations in performance. Specifically, Qwen3-VL-8B-Instruct shows a 6.7\% change in AA, Qwen3-VL-8B-Thinking shows a 9.3\% change, and Qwen3-VL-235B-A22B-Instruct shows a 6.2\% change. This suggests that while the Qwen3-VL family models generally perform on par with some closed-source models, their robustness still requires improvement.

\begin{table*}[thbp]
\centering
\renewcommand{\arraystretch}{1.2}
\setlength\tabcolsep{4pt} 
\resizebox{\linewidth}{!}
{
\begin{tabular}{l|cccccccc|c}
\toprule
\multicolumn{1}{c|}{\smash{\raisebox{-2ex}{\textbf{Model}}}} & \textbf{PD} & \textbf{AN} & \textbf{LC} & \textbf{AR} & \textbf{VC} & \textbf{CP} & \textbf{DM} & \textbf{FP} & \textbf{ALL}\\
\cmidrule(lr){2-9} \cmidrule(lr){10-10}
& AA / PS / PA & AA / PS / PA & AA / PS / PA & AA / PS / PA & AA / PS / PA & AA / PS / PA & AA / PS / PA & AA / PS / PA & AA / PS / PA \\
    \midrule
    Qwen3-VL-8B-Instruct & 16.9 / 16.9 / 16.9 & 35.9 / 35.2 / 35.9 & 28.8 / 27.7 / 28.8 & 24.6 / 24.1 / 24.6 & 34.6 / 34.1 / 34.6 & 0.0 / 0.0 / 0.0 & 43.9 / 43.2 / 43.9 & 16.9 / 16.9 / 16.9 & 27.4 / 26.9 / 27.4 \\
    Qwen3-VL-8B-Thinking & 35.6 / 35.6 / 35.6 & 44.7 / 44.7 / 44.7 & 36.4 / 36.4 / 36.4 & 47.4 / 47.4 / 47.4 & 28.9 / 28.9 / 28.9 & 2.9 / 2.9 / 2.9 & 45.5 / 45.5 / 45.5 & 18.5 / 18.5 / 18.5 & 35.0 / 35.0 / 35.0 \\
    Qwen3-VL-30B-A3B-Instruct & 32.2 / 29.2 / 30.5 & 40.8 / 40.5 / 40.8 & 40.9 / 40.5 / 40.9 & \highlight{LightBlue}{61.4} / 61.4 / 61.4 & 40.4 / 39.9 / 40.4 & \highlight{LightBlue}{8.6} / 8.6 / 8.6 & 59.1 / 59.1 / 59.1 & 24.6 / 24.2 / 24.6 & 40.2 / 39.6 / 40.0 \\
    Qwen3-VL-30B-A3B-Thinking & \highlight{LightBlue}{54.2} / 54.2 / 54.2 & \highlight{LightBlue}{53.4} / 53.4 / 53.4 & \highlight{LightBlue}{47.0} / 47.0 / 47.0 & 47.4 / 47.4 / 47.4 & \highlight{LightBlue}{48.1} / 48.1 / 48.1 & \highlight{LightBlue}{8.6} / 8.6 / 8.6 & \highlight{LightBlue}{62.1} / 62.1 / 62.1 & \highlight{LightBlue}{36.9} / 36.9 / 36.9 & \highlight{LightBlue}{47.3} / 47.3 / 47.3 \\

    \bottomrule
\end{tabular}
}
\caption{Comparison of model performance on the Answer Accuracy (AA), Process Score (PS), and Process-Qualified Accuracy (PA) metrics in the text-video modality (\textit{random seed}=0).}
\label{tab:exp_video}
\end{table*}

\subsection{Experimental Results on Text-Video Modality}\label{sec:app_video}
Here, we further evaluate VLMs on the text-video modality of DynaSolidGeo. Since the OpenRouter API does not yet support video inputs for most MLLMs--and even for the Gemini family, the support remains unstable (frequent network errors occur during API calls)--we therefore locally deploy and evaluate several representative open-source models: Qwen3-VL-8B-Instruct, Qwen3-VL-8B-Thinking, Qwen3-VL-30B-A3B-Instruct, and Qwen3-VL-30B-A3B-Thinking.
In particular, we conduct the evaluation on a batch of instances sampled with \textit{random seed} 0, while keeping all model hyperparameters consistent with those in Section \ref{sec:exp_setup}.
The results are summarized in Table \ref{tab:exp_video}.

Overall, these VLMs perform worse on the text-video modality compared to the text-image modality. Across different task categories, the best performance on Area calculation (AR) is achieved by Qwen3-VL-30B-A3B-Instruct, with AA, PS, and PA all reaching 61.4\%. In all other categories, Qwen3-VL-30B-A3B-Thinking performs best. On Counting Problems (CP), all models perform extremely poorly, with all metrics below 9\%, and Qwen3-VL-8B-Instruct even drops to 0.0\%, revealing a major limitation of current VLMs on this task. Furthermore, the \textit{thinking} models show smaller drops in PA relative to AA compared to \textit{instruct} models, indicating stronger reasoning capability, which is consistent with the trend observed in the text-image modality.

\subsection{More Details of Error Analysis}\label{sec:app_error}
\begin{figure*}[!t]
  \centering
  \includegraphics[width=1\linewidth]{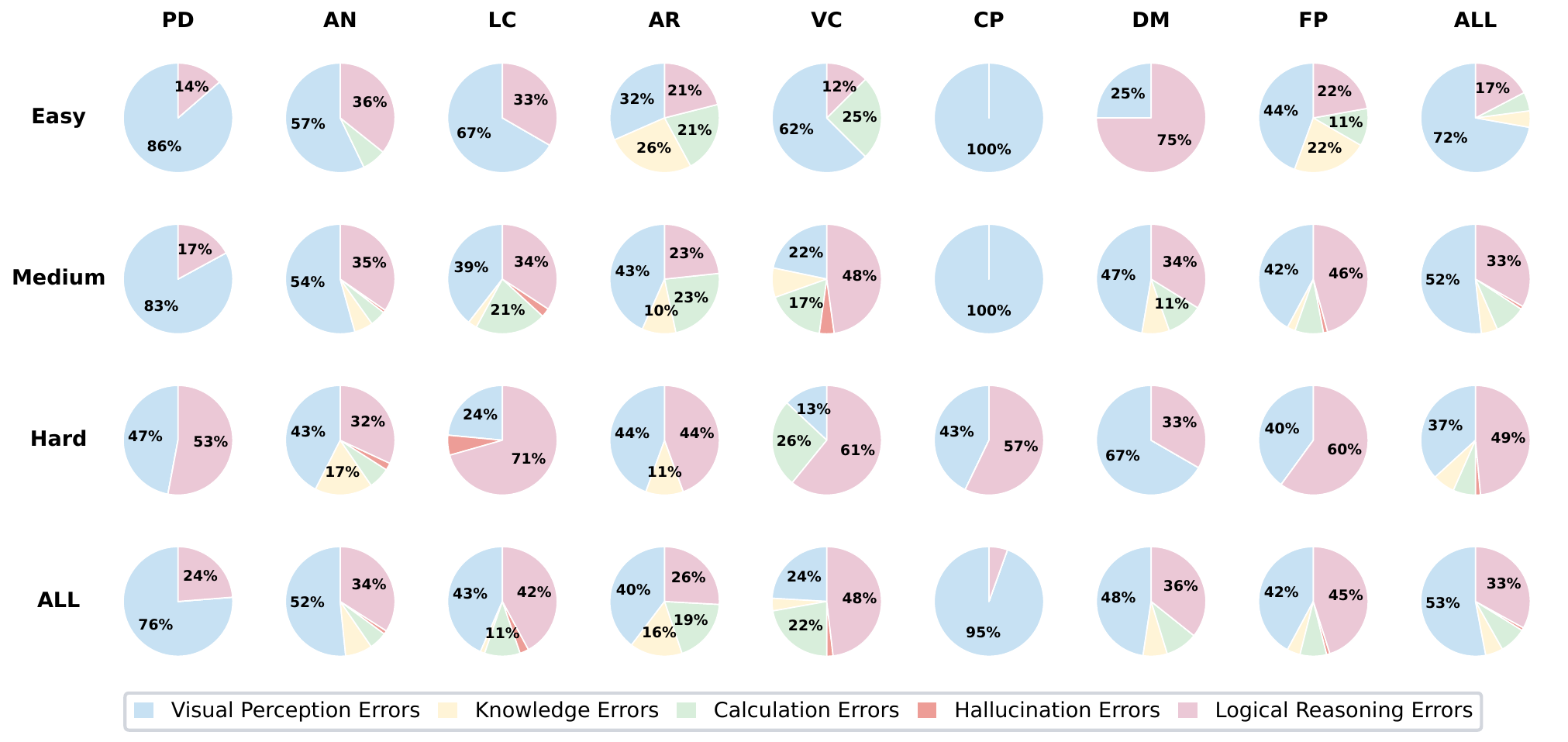}
   \caption{Detailed error distribution on Gemini-2.5-Pro.}
   \label{fig:gemini_matrix_pies}
\end{figure*}

\begin{figure*}[!thbp]
  \centering
  \includegraphics[width=1\linewidth]{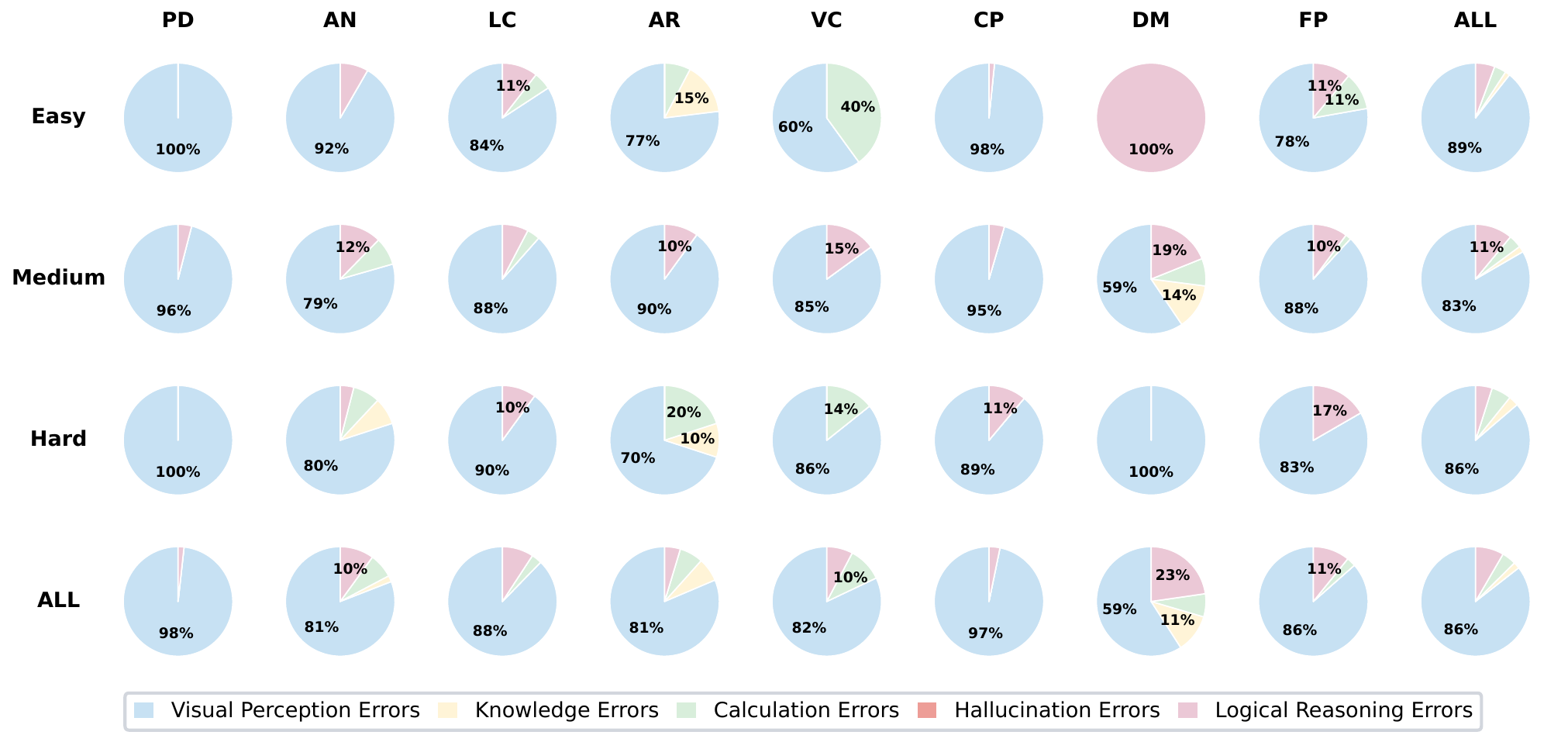}
   \caption{Detailed error distribution on Qwen3-VL-30B-A3B-Thinking.}
   \label{fig:qwen_matrix_pies}
\end{figure*}

Here, we provide additional details of the error analysis conducted on two representative models: Gemini-2.5-Pro and Qwen3-VL-30B-A3B-Thinking. Specifically, we thoroughly examine the distribution of error types--Visual Perception Errors, Logical Reasoning Errors, Calculation Errors, Knowledge Errors, and Hallucination Errors--across different task categories and difficulty levels. The results are presented in Figures \ref{fig:gemini_matrix_pies} and \ref{fig:qwen_matrix_pies}.

For Gemini-2.5-Pro, Visual Perception Errors and Logical Reasoning Errors dominate across most categories.
In Positional Relationship Determination (PD), Angle Calculation (AN), Length and Distance Calculation (LC), Volume Calculation (VC), Counting Problems (CP), and Folding and Unfolding Problems (FP), the proportion of Visual Perception Errors for Gemini-2.5-Pro gradually decreases as the difficulty level increases, while the share of Logical Reasoning Errors rises accordingly.
This suggests that when reasoning cues become more complex, Gemini-2.5-Pro maintains relatively stable perception but struggles to reason through multi-step or abstract geometric relations.
Moreover, Hallucination Errors start to emerge in medium- and high-difficulty problems, implying that the model tends to compensate for uncertainty in perception or reasoning by fabricating unsupported intermediate steps, revealing its limited robustness in handling complex 3D spatial reasoning.

By contrast, Qwen3-VL-30B-A3B-Thinking exhibits a much higher proportion of Visual Perception Errors but markedly fewer reasoning and hallucination errors, indicating that it maintains more consistent logical structures once visual understanding is correct. This finds that Qwen3-VL-30B-A3B-Thinking is stronger in symbolic reasoning but more sensitive to visual ambiguity or occlusion.

Overall, the two models display complementary weaknesses: Gemini-2.5-Pro tends to misreason over correct visual cues, while Qwen3-VL-30B-A3B-Thinking tends to misperceive geometric relations even when reasoning remains sound. These trends further confirm that current VLMs still struggle to integrate visual perception and logical reasoning coherently in 3D spatial problem solving.


\section{DynaSolidGeo as a Training Dataset}
To assess the effectiveness and practical utility of DynaSolidGeo, we additionally examine its impact on VLM performance when used as a training dataset.

\subsection{Experimental Setup}
We sample $K = 10$ batches of instances using \textit{random seeds} from 0 to 9, resulting in a total of 5,030 samples. These samples are divided into a training set (3,627 samples), a validation set (403 samples), and a test set (1,000 samples). The distribution across task categories is kept consistent among the three subsets, and all subsets are ensured to be non-overlapping with respect to the seed questions.
Using the training set, we fine-tuned Qwen3-VL-4B-Instruct and Qwen3-VL-8B-Instruct with the Group Relative Policy Optimization (GRPO) algorithm, obtaining Qwen3-VL-4B-DynaSolidGeo and Qwen3-VL-8B-DynaSolidGeo, respectively. We then compared the changes in their Answer Accuracy (AA) before and after training.
We conduct GRPO training using the Verl framework on 8*NVIDIA H100 GPUs. Specifically, we adopt an answer-based reward scheme, assigning a reward of 1 for correct answers and 0 otherwise. Each question is rolled out 5 responses, with a maximum response length of 4096 tokens. For other hyperparameters, we set the learning rate to 1e-6, the batch size to 512, and train for 3 epochs, totaling 21 steps.
\subsection{Experimental Results}
The experimental results are shown in Table \ref{tab:exp_train}.
After being fine-tuned on the DynaSolid training set, the performances of Qwen3-VL-4B-Instruct and Qwen3-VL-8B-Instruct improved by 8.6\% and 9.0\%, respectively.
The remarkable performance gain highlights the effectiveness and high quality of DynaSolidGeo dataset.
As a large-scale dynamic benchmark, it establishes a new foundation for evaluating and improving spatial mathematical reasoning in multimodal models, offering the community a robust and extensible resource for advancing reasoning-centric VLM research.

\begin{table}[!htbp]
\centering
\renewcommand{\arraystretch}{1.2}
\setlength\tabcolsep{4pt} 
\resizebox{0.8\linewidth}{!}
{
\begin{tabular}{l|c}
\toprule
\textbf{Model} &  \textbf{Answer Accuracy (AA)}\\
    \midrule
    Qwen3-VL-4B-Instruct & 16.2  \\
    Qwen3-VL-4B-DynaSolidGeo & 24.8 \textcolor{green}{(+8.6)} \\
    \midrule
    Qwen3-VL-8B-Instruct & 17.0  \\
    Qwen3-VL-8B-DynaSolidGeo & 26.0 \textcolor{green}{(+9.0)} \\
    \bottomrule
\end{tabular}
}
\caption{Performance gain after training with DynaSolidGeo.}
\label{tab:exp_train}
\end{table}

\section{More Details of Data Annotation Pipeline}
In this section, we provide more details of data annotation pipeline.
\subsection{Parameterized Question Example}
In the \textit{JSON-Based Question Parametrization} stage, the variable parameters in the question statements (e.g., endpoints, side lengths, areas, volumes, ratios) are represented as variables enclosed in curly braces \{\} using f-string syntax. For example, the endpoint ``A" is replaced with \{point\_A\}, and the side length is replaced with \{len\_a\}. In this way, during the subsequent \textit{Parameterized Program Construction} stage, these placeholders can be directly assigned to Python variables when assembling the parameterized program.
An example of the parameterized question annotation (in JSON format) is shown below.

\begin{table}[hbtp]
\begin{tcolorbox}[colback=white, colframe=gray!75!black, 
title=An Example of Parameterized Question (Part 1), boxrule=0.5mm, arc=3mm, auto outer arc]

\{

        ``id": ``hsmcel\_s\_12\_1",
        
        ``type": 7,
        
        ``level": 3,
        
        ``cn\_problem": ``......",
        
        ``en\_problem": ``Let cube \{point\_A\}\{point\_B\}\{point\_C\}\{point\_D\}-\{point\_A1\}\{point\_B1\}\{point\_C1\}\{point\_D1\} have edge length \{len\_a\}. Find the minimum distance between point \{point\_P\} on the incircle of the upper base \{point\_A\}\{point\_B\}\{point\_C\}\{point\_D\} and point \{point\_Q\} on the circle passing through vertices \{point\_A\}, \{point\_B\}, \{point\_C1\}, \{point\_D1\}.",
        
        ``cn\_think": ``......",

\end{tcolorbox}
\end{table}

\begin{table}[hbtp]
\begin{tcolorbox}[colback=white, colframe=gray!75!black, 
title=An Example of Parameterized Question (Part 2), boxrule=0.5mm, arc=3mm, auto outer arc]
``en\_think": ``[Basic properties of geometric objects]\textbackslash n1. Upper base incircle:\textbackslash n   The upper base \{point\_A\}\{point\_B\}\{point\_C\}\{point\_D\} is a square with edge length \{len\_a\}, its incircle has center at the upper base center \{point\_O1\}, radius $r_1$ = \{len\_a\}/2. This circle is the intersection of the edge-tangent sphere with the upper base, the sphere center is the cube center \{point\_O\} (midpoint of body diagonal), radius is half the face diagonal, i.e., r = \{len\_a\}$\sqrt{2}$/2. Therefore, any point \{point\_P\} on the incircle has constant distance to sphere center \{point\_O\}: \{point\_O\}\{point\_P\} = r = \{len\_a\}$\sqrt{2}$/2.\textbackslash n\textbackslash n2. Circle passing through \{point\_A\}, \{point\_B\}, \{point\_C1\}, \{point\_D1\}:\textbackslash n   The four points \{point\_A\}, \{point\_B\}, \{point\_C1\}, \{point\_D1\} form a rectangle (\{point\_A\}\{point\_B\} is edge length \{len\_a\}, \{point\_A\}\{point\_D1\}, \{point\_B\}\{point\_C1\} are face diagonals \{len\_a\}$\sqrt{2}$), its circumcircle is a great circle of the cube's circumscribed sphere. The circumscribed sphere radius is half the body diagonal, i.e., R = \{len\_a\}$\sqrt{3}$/2, so any point \{point\_Q\} on this circle has constant distance to sphere center \{point\_O\}: \{point\_O\}\{point\_Q\} = R = \{len\_a\}$\sqrt{3}$/2.\textbackslash n\textbackslash n[Derivation of minimum distance (triangle inequality)]\textbackslash nFor any point \{point\_P\} (on the incircle) and \{point\_Q\} (on the circle through \{point\_A\}, \{point\_B\}, \{point\_C1\}, \{point\_D1\}), by triangle inequality:\textbackslash n\{point\_P\}\{point\_Q\} $\ge$ $\mid$\{point\_O\}\{point\_Q\} - \{point\_O\}\{point\_P\}$\mid$\textbackslash nEquality holds if and only if \{point\_P\}, \{point\_Q\}, \{point\_O\} are collinear and \{point\_P\} is between \{point\_O\} and \{point\_Q\}.\textbackslash n\textbackslash n[Calculation and verification]\textbackslash n- Substituting \{point\_O\}\{point\_P\} = \{len\_a\}$\sqrt{2}$/2 and \{point\_O\}\{point\_Q\} = \{len\_a\}$\sqrt{3}$/2, we get:\textbackslash n $\mid$ \{point\_O\}\{point\_Q\} - \{point\_O\}\{point\_P\} $\mid$ = \{len\_a\}$\sqrt{3}$/2 - \{len\_a\}$\sqrt{2}$/2 = \{len\_a\}($\sqrt{3}$ - $\sqrt{2}$)/2\textbackslash n\textbackslash n- Collinearity verification:\textbackslash n  The midpoint \{point\_P0\}(\{len\_a\}, \{len\_a\}/2, \{len\_a\}) of upper base edge \{point\_A\}\{point\_B\} satisfies both the upper base incircle equation and lies on the plane of the circle through \{point\_A\}, \{point\_B\}, \{point\_C1\}, \{point\_D1\}. The ray \{point\_O\}\{point\_P0\} intersects the great circle at \{point\_Q\}, at which time \{point\_P0\}, \{point\_Q\}, \{point\_O\} are collinear, so equality holds.",
        
        ``solution": ``\textbackslash displaystyle d\_\{\textbackslash text\{min\}\} = \textbackslash frac\{\{len\_a\}(\textbackslash sqrt\{3\} - \textbackslash sqrt\{2\})\}\{2\}"
        
\}
\end{tcolorbox}
\end{table}

\subsection{Answer Function Example}
To ensure the dynamic generation of ground-truth answers according to the variable parameters defined in each question statement, a dedicated Python function (Answer Function) is created for every seed question. Each function computes the ground truth using the annotated variables from the parameterized question. An example of the annotated answer function is provided below.

\begin{table}[hbtp]
\begin{tcolorbox}[colback=white, colframe=gray!75!black, 
title=An Example of Answer Function (Python), boxrule=0.5mm, arc=3mm, auto outer arc]

{\small
\begin{verbatim}
def calculate(len_a):
  r1=len_a*(math.sqrt(3)-math.sqrt(2))
  min_distance=r1/2
  return min_distance
\end{verbatim}
}
\end{tcolorbox}
\end{table}

\subsection{Parameterized Python Program Example}
In the \textit{Parameterized Program Construction} stage, we assemble the Parameterized Python Program by combining the parameterized question with its corresponding answer function. An example is shown below.
The program randomizes question variables and image camera viewpoints.

\begin{table}[htbp]
\begin{tcolorbox}[colback=white, colframe=gray!75!black, 
title=An Example of Parameterized Python Program, boxrule=0.5mm, arc=3mm, auto outer arc]
{\small
\begin{verbatim}
......

# Scaling factor
len_scaling_factor = round(
random.uniform(0.1, 100.0), 1)

# Generate random point names
point_A, point_B, point_C, point_D,
point_A1, point_B1, point_C1,
point_D1, point_P, point_Q = 
random.sample(string.ascii_uppercase,10)

# Add answer function
def calculate(len_a):
  r1 = len_a*(math.sqrt(3)-math.sqrt(2))
  min_distance = r1/2
  return min_distance

len_a = 1
# Generate random lengths
len_a = round(len_scaling_factor
* float(len_a), 2)
# Calculate the result
result = calculate(len_a)

# --- 1. save JSON ---------------
json_data = {
    "id": "hsmcel_s_12_1",
    "type": 7,
    "level": 3,
    "cn_problem": f"......",
    "en_problem": f"......",
    "solution": f"{result}",
    "image": f"....png"
}

# video mode
if args.mode == 1:
    json_data["image"] = f"....mp4"
    
# --- 2. save MATLAB command JSONL --
azimuth = (-150+random.randint(0, 360))
elevation = (25+random.randint(0, 360))

......
\end{verbatim}
}
\end{tcolorbox}
\end{table}

\subsection{Parameterized Visualization Program Example}
In the \textit{MATLAB-Based Geometry Visualization} stage, the parameterized question and the static MATLAB visualization program are integrated into a Parameterized Visualization Program. An example is shown below.
It takes the variable parameters (including camera view angles) passed from the Parameterized Python Program and dynamically renders the image or video.

\begin{table*}[htbp]
\begin{tcolorbox}[colback=white, colframe=gray!75!black, 
title=An Example of Parameterized Visualization Program (MATLAB), boxrule=0.5mm, arc=3mm, auto outer arc]

{\small
\begin{verbatim}
function visual(mode, azimuth, elevation, point_A, point_B, point_C, point_D, point_A1,
point_B1, point_C1, point_D1, point_P, point_Q)
  close all;
  fig = figure('Visible', 'off');
  L = 1;
  D = [0,0,0]; A = [L,0,0];
  B = [L,L,0]; C = [0,L,0];
  D1 = [0,0,L]; A1 = [L,0,L];
  B1 = [L,L,L]; C1 = [0,L,L];
  O_p = (A1+C1)/2; r_p = L/2;
  O_q = (A+C1)/2; r_q = norm(A-O_q);
  hold on;  
  plot3([A1(1),B1(1)], [A1(2),B1(2)], [A1(3),B1(3)], 'k-', 'LineWidth', 2);
  plot3([B1(1),C1(1)], [B1(2),C1(2)], [B1(3),C1(3)], 'k-', 'LineWidth', 2);
  plot3([C1(1),D1(1)], [C1(2),D1(2)], [C1(3),D1(3)], 'k-', 'LineWidth', 2);
  plot3([D1(1),A1(1)], [D1(2),A1(2)], [D1(3),D1(3)], 'k-', 'LineWidth', 2);
  ......

  text(A(1)+0.05,A(2)-0.1,A(3),point_A, 'FontSize',14,'FontWeight','bold');
  text(B(1)+0.05,B(2)+0.05,B(3),point_B, 'FontSize',14,'FontWeight','bold');
  text(C(1)-0.1,C(2)+0.05,C(3),point_C, 'FontSize',14,'FontWeight','bold');
  text(D(1)-0.1,D(2)-0.1,D(3),point_D, 'FontSize',14,'FontWeight','bold');
  ......
  
  if mode == 0
    img_dir = fullfile('..', '..', 'data', 'images');
    if ~exist(img_dir, 'dir')
        mkdir(img_dir);
    end
    img_path = fullfile(img_dir, [mfilename, '.png']);
    frame = getframe(gcf);

    imwrite(frame.cdata, img_path);
    fprintf('Image saved as: %s \n', img_path);
  elseif mode == 1
    vid_dir = fullfile('..', '..', 'data', 'videos');
    if ~exist(vid_dir, 'dir')
        mkdir(vid_dir);
    end
    vid_path = fullfile(vid_dir, [mfilename, '.mp4']);
    video = VideoWriter(vid_path, 'MPEG-4');
    video.FrameRate = 24;
    open(video);
    ......

    for angle = (azimuth+3):3:(azimuth+360)
        view(angle, elevation);
        frame = getframe(gcf);
        writeVideo(video, frame);
    end
    ......
  end
  hold off;
  close(fig);
end
\end{verbatim}
}
\end{tcolorbox}
\end{table*}

\subsection{Prompts for Data Annotation Pipeline}
In the \textit{Expert-Guided Parametrization and Visualization} stage of the data annotation pipeline, we follow a \textit{human-in-the-loop} strategy, where human experts collaborate with large models to create a JSON annotation and a MATLAB program for each source question. 
The prompts for JSON generation and MATLAB code generation are shown below.

\begin{table*}[htbp]
\begin{tcolorbox}[colback=white, colframe=gray!75!black, 
title=Prompt for JSON generation (Part 1), boxrule=0.5mm, arc=3mm, auto outer arc]

\# Annotation Task Instructions (Markdown Version)

\vspace{\baselineskip}

\#\# Task Description

Perform structured annotation on input math problems and output in JSON format.

Only process "problems with definite answers requiring calculation", excluding proof or discussion problems.

If a problem contains multiple sub-questions, extract only the calculation sub-questions and split them into separate individual problems.

\vspace{\baselineskip}

\#\# Output Format Requirements (All Fields Required)

\`{}\`{}\`{}json

\{

    ``id": ``filename-topic\_number-lecture-question\_number-i",
    
    ``type": 1-8,
    
    ``level": 1-3,
    
    ``origin\_problem": ``Original problem text (Chinese + LaTeX, no variable substitution)",
    
    ``cn\_problem": ``Chinese problem text (necessary variables using f-string + LaTeX)",
    
    ``en\_problem": ``English problem text (necessary variables using f-string + LaTeX)",
    
    ``cn\_think": ``Chinese solution process (with variables and LaTeX)",
    
    ``en\_think": ``English solution process (with variables and LaTeX)",
    
    ``solution": ``Final answer expressed with variables (LaTeX + f-string)"
    
\}

\`{}\`{}\`{}

Output as a JSON array.

Return [] if no extractable content.

\vspace{\baselineskip}

\#\# Problem Type Classification (type)

$\mid$ type $\mid$ Description $\mid$

$\mid$------$\mid$-------------$\mid$

$\mid$ 1 $\mid$ Position relationship determination between lines and planes $\mid$

$\mid$ 2 $\mid$ Angle or trigonometric value calculation $\mid$

$\mid$ 3 $\mid$ Length and distance calculation $\mid$

$\mid$ 4 $\mid$ Area calculation $\mid$

$\mid$ 5 $\mid$ Volume calculation $\mid$

$\mid$ 6 $\mid$ Solid geometry counting problems $\mid$

$\mid$ 7 $\mid$ Dynamic/moving point problems $\mid$

$\mid$ 8 $\mid$ Folding and unfolding problems $\mid$

\vspace{\baselineskip}

\#\# Difficulty Classification (level)

$\mid$ level $\mid$ Description $\mid$

$\mid$-------$\mid$-------------$\mid$

$\mid$ 1 $\mid$ Easy $\mid$

$\mid$ 2 $\mid$ Medium $\mid$

$\mid$ 3 $\mid$ Hard (Final/Challenging) $\mid$

\vspace{\baselineskip}

\#\# Variable Substitution Rules

$\mid$ Original Content $\mid$ Substitution Method $\mid$

$\mid$-----------------$\mid$---------------------$\mid$

$\mid$ Point letters P, A, B… $\mid$ \{point\_P\}, \{point\_A\}, \{point\_B\} $\mid$

$\mid$ P1, A', etc. $\mid$ \{point\_P1\}, \{point\_A\_prime\} $\mid$

$\mid$ Line segment lengths and numerical values $\mid$ \{len\_AB\}, \{len\_PC\} $\mid$

\end{tcolorbox}
\end{table*}

\begin{table*}[thbp]
\begin{tcolorbox}[colback=white, colframe=gray!75!black, 
title=Prompt for JSON generation (Part 2), boxrule=0.5mm, arc=3mm, auto outer arc]

$\mid$ Auxiliary letters $\mid$ Variablize $\mid$

$\mid$ Mathematical expressions $\mid$ Use LaTeX $\mid$

$\mid$ If problem has diagram $\mid$ Add $<$image$>$ (omit if no diagram) $\mid$

\vspace{\baselineskip}

Example:

Original: AB=2, find the cosine of the angle between PA and AC.

After substitution:

\`{}\`{}\`{}

\{point\_A\}\{point\_B\} = \{len\_AB\}, find the cosine of the angle between skew lines \{point\_P\}\{point\_A\} and \{point\_A\}\{point\_C\}.

\`{}\`{}\`{}

\vspace{\baselineskip}

\#\# Important Notes

- Each sub-question should be an independent JSON entry

- All fields must be filled (no empty values allowed)

- LaTeX does not escape JSON quotes

- Output only JSON, no additional explanations or text

\vspace{\baselineskip}

\#\# Execution Method

When I send problem content, please output structured JSON according to the above rules.

\end{tcolorbox}
\end{table*}

\begin{table*}[htbp]
\begin{tcolorbox}[colback=white, colframe=gray!75!black, 
title=Prompt for MATLAB code generation (Part 1), boxrule=0.5mm, arc=3mm, auto outer arc]
\# Role: Matlab Code Generation Expert for 3D Geometry Visualization

\vspace{\baselineskip}

\#\# Task

Given an image and description of a solid geometry problem, your task is to generate Matlab code to visualize the solid geometric structure of the problem. 

\vspace{\baselineskip}

Please refer to variable settings that correspond to the problem's Description. Make sure to include these variables at the `2. establish coordinate system according to the problem conditions (need change)' of the Matlab template below, allowing for easy adjustment of the geometry.

\vspace{\baselineskip}

Notice:

- All lines and points need to be black.

- The generated image should show only the geometry without legends, axes, or any extraneous annotations.

- Please output only Matlab code.

\vspace{\baselineskip}

\#\# Skills

- Able to generate accurate Matlab code based on input.

- Experienced in handling geometric figures such as circles, spheres, and cylinders, even when they are not directly defined in the input.

\vspace{\baselineskip}

\#\# Workflow

1. **Problem Parsing**:

   - Extract geometric type (prism/pyramid/cylinder/cone/cuboid/etc.) and key parameters (edge lengths l, heights h, angles $\theta$)
   
   - Identify labeled points (A/B/C/D...) and their relationships (e.g., ``A,B,C are base vertices", ``D is apex")

\vspace{\baselineskip}

2. **Coordinate System Initialization**:

   - Position the base face's centroid at the origin (0,0,0) on the XY-plane, using the Z-axis for vertical alignment in symmetric cases.
   
   - Reference points in problems guide system setup for asymmetric shapes (e.g., vertex coordinates).
   
   - Define Z-axis as vertical (height direction); set apex/height z-coordinate = h (or k*h with scaling factor k)

\vspace{\baselineskip}

3. **Point Coordinate Derivation**:

   - For regular bases: Calculate vertices using n-gon formulas (see Rule 2)
   
   - For irregular bases: Solve coordinate system using given distances/angles (e.g., right triangles, parallelograms)
   
   - For height-related points: Involve apex coordinates aligned vertically from base vertices, Apex D = (Ax, Ay, h) for apex above base point A

\vspace{\baselineskip}

4. **Edge Type Classification**:

   - Enumerate all edges (AB, BC, CD...)
   
   - Classify each edge:
   
     - Solid: If it is a structural edge (part of the shape's skeleton) or lies on a face boundary
     
     - Dashed: If it is a non-structural edge (diagonal, space diagonal, or non-face boundary connection)

\vspace{\baselineskip}

5. **Validation \& Adjustment**:

   - Check edge lengths against problem statement (allow ±0.01 tolerance)
   
   - Verify face membership for solid edges (ensure they belong to $\ge$ 1 face)
   
   - Adjust coordinates if length discrepancies exceed tolerance (e.g., scale radius by 0.99)

\vspace{\baselineskip}

6. **Special Geometric Shapes Handling**:

\end{tcolorbox}
\end{table*}

\begin{table*}[htbp]
\begin{tcolorbox}[colback=white, colframe=gray!75!black, 
title=Prompt for MATLAB code generation (Part 2), boxrule=0.5mm, arc=3mm, auto outer arc]
   - For cylinders, cones, and frustums: Include additional attributes such as type (cylinder/cone/frustum), base center, radius, and height in the JSON.
   
    - For spheres: Include additional attributes such as type (sphere), center, and radius in the JSON.

\vspace{\baselineskip}

\#\# Rules

You must strictly follow the Matlab template below. Only edit the parts marked with `\# need change', all code marked with '\# fixed' must remain unmodified.

\vspace{\baselineskip}

\# --- 1. function header (fixed) ----------------------------

function visual(mode, azimuth, elevation)

  \ \ \% mode parameter: 0=save current view image, 1=save rotation animation video
    
  \ \ \% Close all existing figure windows and create a new invisible window
    
  \ \ close all;
    
  \ \ fig = figure(`Visible', `off');

\vspace{\baselineskip}

\# --- 2. establish coordinate system according to the problem conditions (need change) -------------

  \ \ \% Variable settings (adjustable)
    
  \ \ len\_a = 2;                    \% Side length variable
    
  \ \ ang\_theta = pi/3;             \% Angle variable (60 degrees)

    \vspace{\baselineskip}
    
  \ \ \% Establish coordinate system according to the problem conditions
    
  \ \ \% Take plane ABC as the xy-plane
    
  \ \ C = [0, 0, 0];
    
  \ \ B = [len\_a, 0, 0];
    
  \ \ A = [2*len\_a*cos(ang\_theta)\^{}2, 2*len\_a*cos(ang\_theta)*sin(ang\_theta), 0];

    \vspace{\baselineskip}
    
  \ \ \% Calculate point E (midpoint of AC)
    
  \ \ E = (A + C) / 2;
    
    \vspace{\baselineskip}
    
  \ \ \% Construct point D
    
  \ \ \% D is on the perpendicular bisector of AC, satisfying AD$\perp$CD and AD=CD
    
  \ \ D = [len\_a*cos(ang\_theta)\^{}2, len\_a*cos(ang\_theta)*sin(ang\_theta), len\_a*cos(ang\_theta)];

    \vspace{\baselineskip}
    
  \ \ \% Calculate optimal point F (when the area of triangle AFC is minimized)
    
  \ \ lambda\_min = sin(ang\_theta)\^{}2;
    
  \ \ F = B + lambda\_min * (D - B);

\vspace{\baselineskip}

\# --- 3. draw (need change) ---------------------------------

   \ \ hold on;

    \vspace{\baselineskip}

   \ \ \% Draw the base triangle ABC
    
   \ \ plot3([A(1), B(1)], [A(2), B(2)], [A(3), B(3)], `k-', `LineWidth', 2);
    
    \ \ plot3([B(1), C(1)], [B(2), C(2)], [B(3), C(3)], `k-', `LineWidth', 2);
    
    \ \ plot3([C(1), A(1)], [C(2), A(2)], [C(3), A(3)], `k-', `LineWidth', 2);

    \vspace{\baselineskip}
    
    \ \ \% Draw the edges from D to each point on the base
    
    \ \ plot3([D(1), A(1)], [D(2), A(2)], [D(3), A(3)], `k-', `LineWidth', 2);
    
    \ \ plot3([D(1), B(1)], [D(2), B(2)], [D(3), B(3)], `k-', `LineWidth', 2);
    
    \ \ plot3([D(1), C(1)], [D(2), C(2)], [D(3), C(3)], `k-', `LineWidth', 2);

    \vspace{\baselineskip}
    
    \ \ \% Draw triangle AFC

\end{tcolorbox}
\end{table*}

\begin{table*}[htbp]
\begin{tcolorbox}[colback=white, colframe=gray!75!black, 
title=Prompt for MATLAB code generation (Part 3), boxrule=0.5mm, arc=3mm, auto outer arc]
    \ \ plot3([A(1), F(1)], [A(2), F(2)], [A(3), F(3)], `k-', `LineWidth', 2);
    
    \ \ plot3([F(1), C(1)], [F(2), C(2)], [F(3), C(3)], `k-', `LineWidth', 2);

    \vspace{\baselineskip}
    
    \ \ \% Draw dashed lines from point E to other points
    
    \ \ plot3([E(1), A(1)], [E(2), A(2)], [E(3), A(3)], `k- -', `LineWidth', 1);
    
    \ \ plot3([E(1), C(1)], [E(2), C(2)], [E(3), C(3)], `k- -', `LineWidth', 1);
    
    \ \ plot3([E(1), B(1)], [E(2), B(2)], [E(3), B(3)], `k- -', `LineWidth', 1);

    \vspace{\baselineskip}
    
    \ \ \% Mark each point
    
    \ \ scatter3(A(1), A(2), A(3), 100, `ko', `filled');
    
    \ \ scatter3(B(1), B(2), B(3), 100, `ko', `filled');
    
    \ \ scatter3(C(1), C(2), C(3), 100, `ko', `filled');
    
    \ \ scatter3(D(1), D(2), D(3), 100, `ko', `filled');
    
    \ \ scatter3(E(1), E(2), E(3), 100, `ko', `filled');
    
    \ \ scatter3(F(1), F(2), F(3), 100, `ko', `filled');

    \vspace{\baselineskip}
    
    \ \ \% Add labels for each point
    
    \ \ text(A(1)+0.1, A(2)+0.1, A(3)+0.1, `A', `FontSize', 14, `FontWeight', `bold');
    
    \ \ text(B(1)+0.1, B(2)+0.1, B(3)+0.1, `B', `FontSize', 14, `FontWeight', `bold');
    
    \ \ text(C(1)+0.1, C(2)+0.1, C(3)+0.1, `C', `FontSize', 14, `FontWeight', `bold');
    
    \ \ text(D(1)+0.1, D(2)+0.1, D(3)+0.1, `D', `FontSize', 14, `FontWeight', `bold');
    
    \ \ text(E(1)+0.1, E(2)+0.1, E(3)+0.1, `E', `FontSize', 14, `FontWeight', `bold');
    
    \ \ text(F(1)+0.1, F(2)+0.1, F(3)+0.1, `F', `FontSize', 14, `FontWeight', `bold');

\vspace{\baselineskip}

\# --- 4. save (fixed) ------------------------------------------

    \ \ \% Set figure properties
    
    \ \ axis equal;
    
    \ \ axis off;           \% Hide axes
    
    \ \ view(azimuth, elevation); \% 45, 30

    \vspace{\baselineskip}
    
    \ \ \% Set background and appearance
    
    \ \ set(gca, `Color', `white');
    
    \ \ set(gcf, `Color', `white');
    
    \ \ set(gcf, `ToolBar', `none');    \% Hide toolbar
    
    \ \ set(gcf, `MenuBar', `none');    \% Hide menu bar

    \vspace{\baselineskip}
    
    \ \ \% Perform corresponding operation according to mode parameter
    
    \ \ if mode == 0
    
        \ \ \ \ \% Save current view image
        
        \ \ \ \ print(mfilename, `-dpng', `-r300');
        
        \ \ \ \ fprintf(`Image saved as: \%s.png\textbackslash n', mfilename);
        
    \ \ elseif mode == 1
    
        \ \ \ \ \% Save rotation animation video
        
        \ \ \ \ video = VideoWriter([mfilename, `.mp4'], `MPEG-4');
        
        \ \ \ \ video.FrameRate = 24;
        
        \ \ \ \ open(video);

        \vspace{\baselineskip}
    
        \ \ \ \ \% Lock camera parameters to prevent distance change during rotation
        
        \ \ \ \ set(gca, `CameraViewAngleMode', `manual');
        
        \ \ \ \ set(gca, `CameraPositionMode', `manual');
        
        \ \ \ \ set(gca, `CameraTargetMode', `manual');

\end{tcolorbox}
\end{table*}

\begin{table*}[thbp]
\begin{tcolorbox}[colback=white, colframe=gray!75!black, 
title=Prompt for MATLAB code generation (Part 4), boxrule=0.5mm, arc=3mm, auto outer arc]
        \ \ \ \ \% The following loop is used to generate the rotation animation video
        
        \ \ \ \ \% angle from 0 to 357, increase by 3 degrees each time, total 120 frames
        
\ \ \ \ for angle = 30:3:390
        
            \ \ \ \ \ \ view(angle, elevation);      \% Set current view azimuth to angle, elevation to 30 degrees
            
            \ \ \ \ \ \ frame = getframe(gcf);\% Capture current figure window frame
            
            \ \ \ \ \ \ writeVideo(video, frame); \% Write frame to video file
            
        \ \ \ \ end

        \vspace{\baselineskip}
        
        \ \ \ \ close(video);
        
        \ \ \ \ fprintf(`Video saved as: \%s.mp4\textbackslash n', mfilename);
        
    \ \ end
    
    \ \ \% Close figure at the end
    
    \ \ hold off;
    
    \ \ close(fig);
    
end
\end{tcolorbox}
\end{table*}

\section{Question Instance Examples}
In this section, we provide several examples of question instances, as shown in the text boxes at the end of this appendix. For each task category, we present three variations sampled with \textit{random seeds} 0, 1, and 2, with each variable parameter highlighted in red.

\begin{table*}[thbp]
\begin{tcolorbox}[colback=white, colframe=gray!75!black, 
title=Question Instance Examples: Positional relationship determination (PD), boxrule=0.5mm, arc=3mm, auto outer arc]
\begin{center}
  \includegraphics[width=1\linewidth]{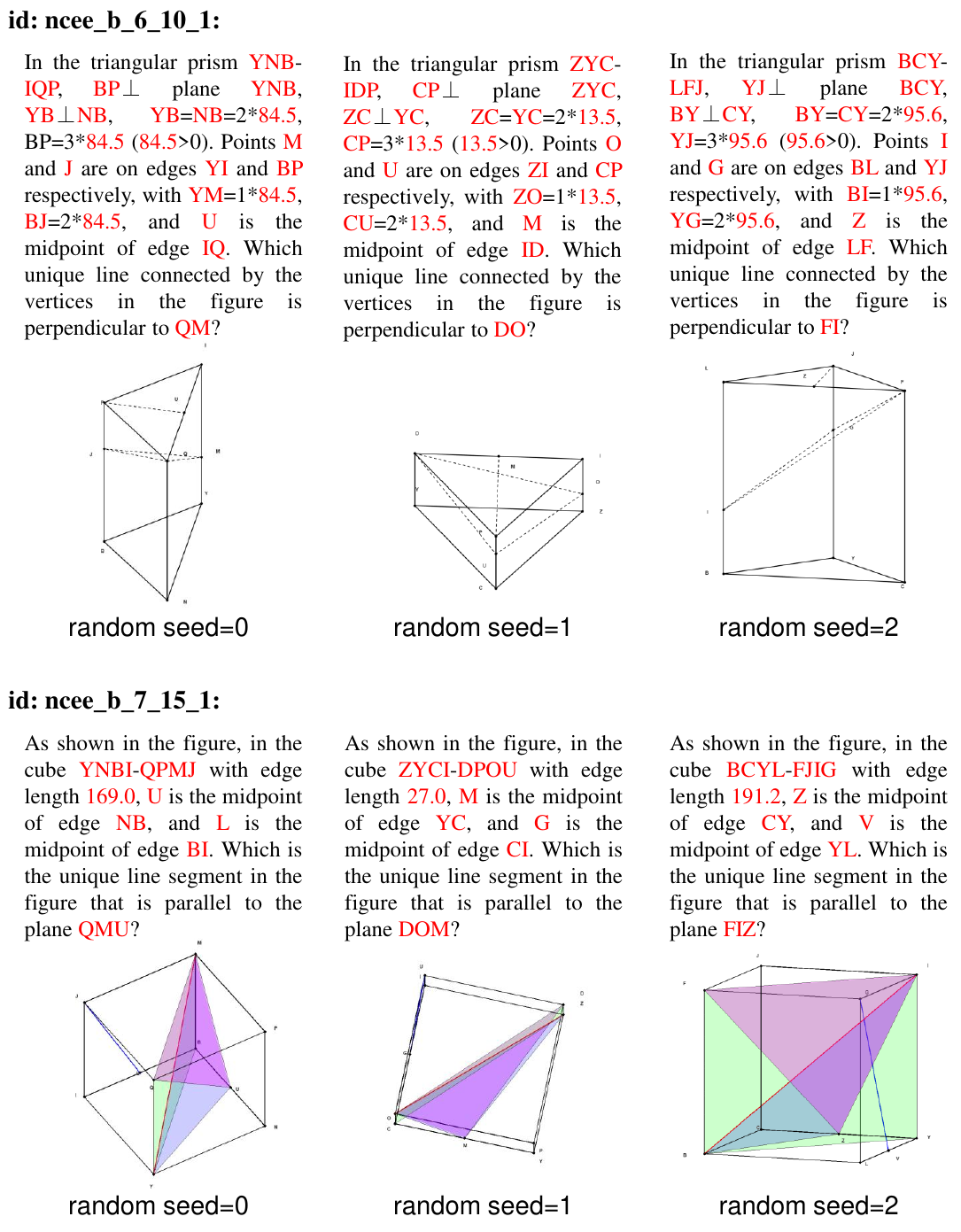}
\end{center}
\end{tcolorbox}
\end{table*}

\begin{table*}[thbp]
\begin{tcolorbox}[colback=white, colframe=gray!75!black, 
title=Question Instance Examples: Angle calculation (AN), boxrule=0.5mm, arc=3mm, auto outer arc]
\begin{center}
  \includegraphics[width=1\linewidth]{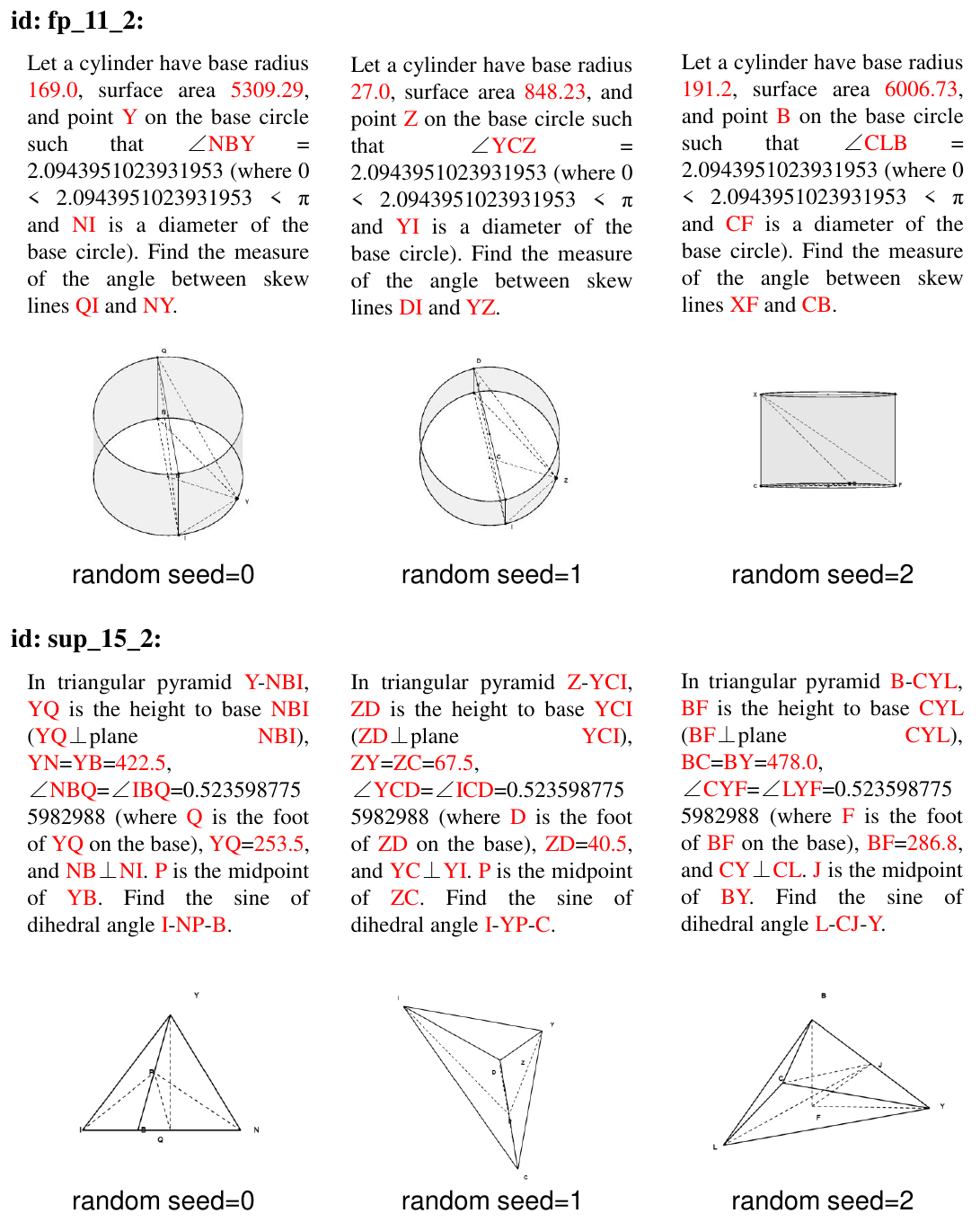}
\end{center}
\end{tcolorbox}
\end{table*}

\begin{table*}[thbp]
\begin{tcolorbox}[colback=white, colframe=gray!75!black, 
title=Question Instance Examples: Length and distance calculation (LC), boxrule=0.5mm, arc=3mm, auto outer arc]
\begin{center}
  \includegraphics[width=1\linewidth]{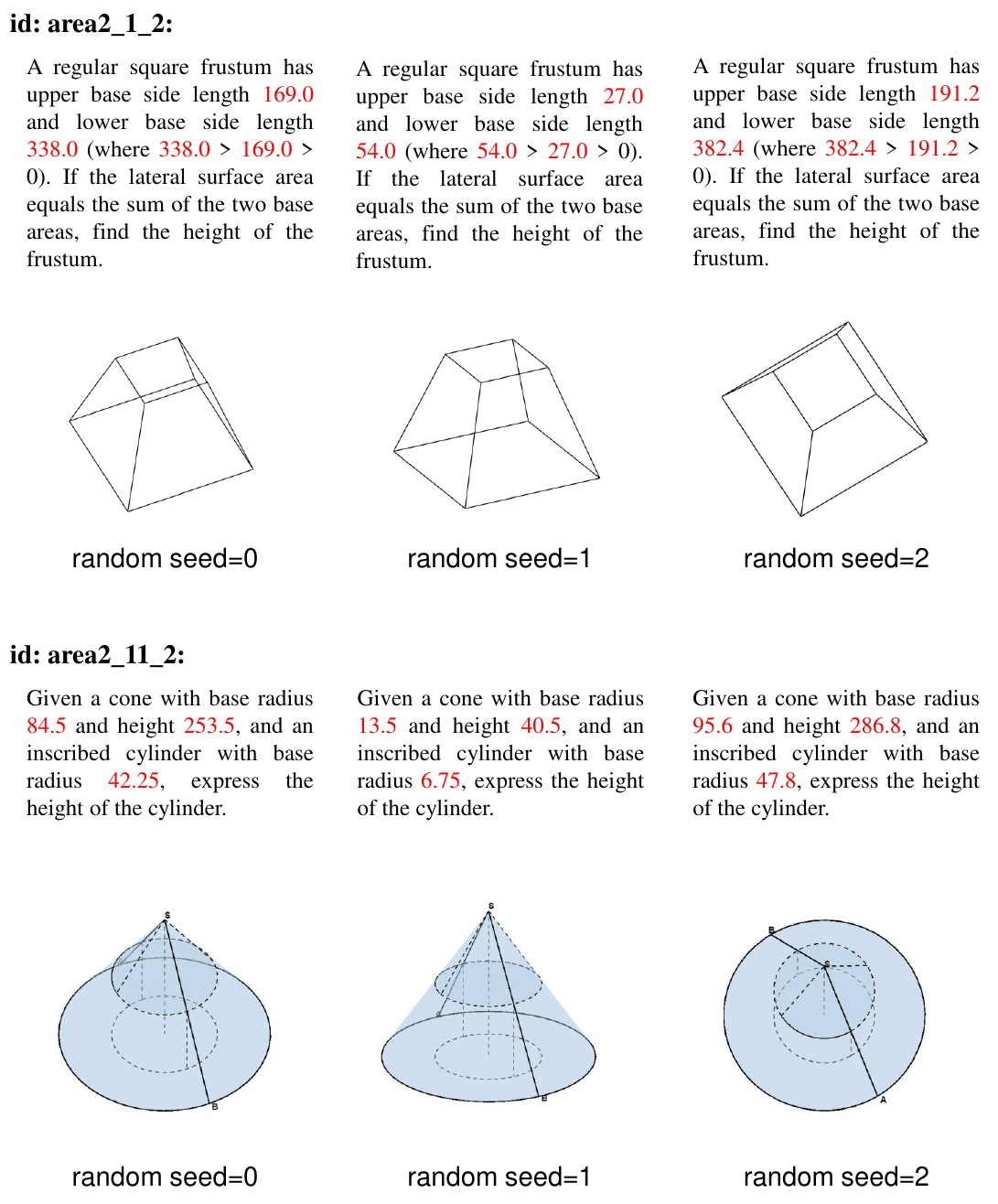}
\end{center}
\end{tcolorbox}
\end{table*}

\begin{table*}[thbp]
\begin{tcolorbox}[colback=white, colframe=gray!75!black, 
title=Question Instance Examples: Area calculation (AR), boxrule=0.5mm, arc=3mm, auto outer arc]
\begin{center}
  \includegraphics[width=1\linewidth]{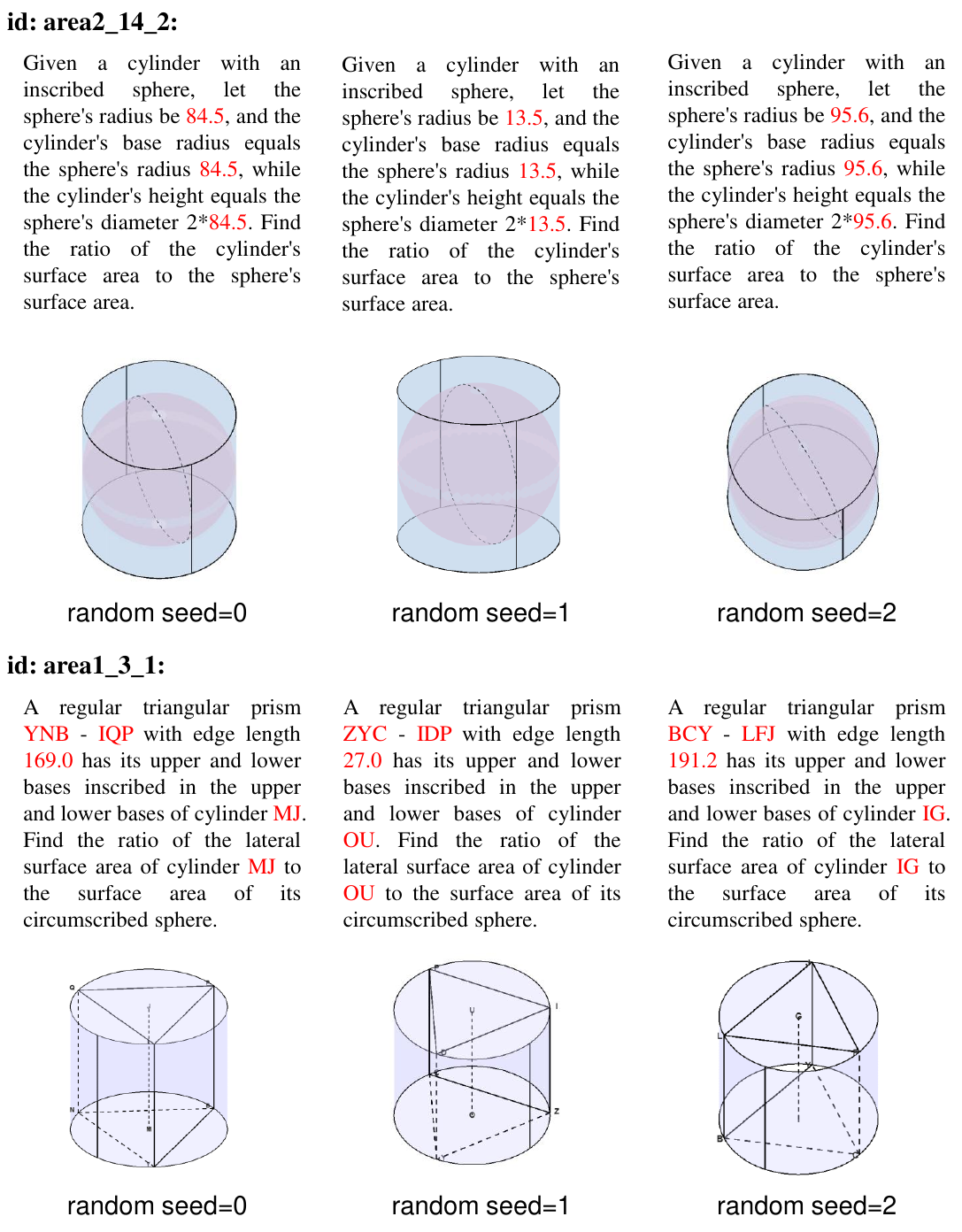}
\end{center}
\end{tcolorbox}
\end{table*}

\begin{table*}[thbp]
\begin{tcolorbox}[colback=white, colframe=gray!75!black, 
title=Question Instance Examples: Volume calculation (VC), boxrule=0.5mm, arc=3mm, auto outer arc]
\begin{center}
  \includegraphics[width=1\linewidth]{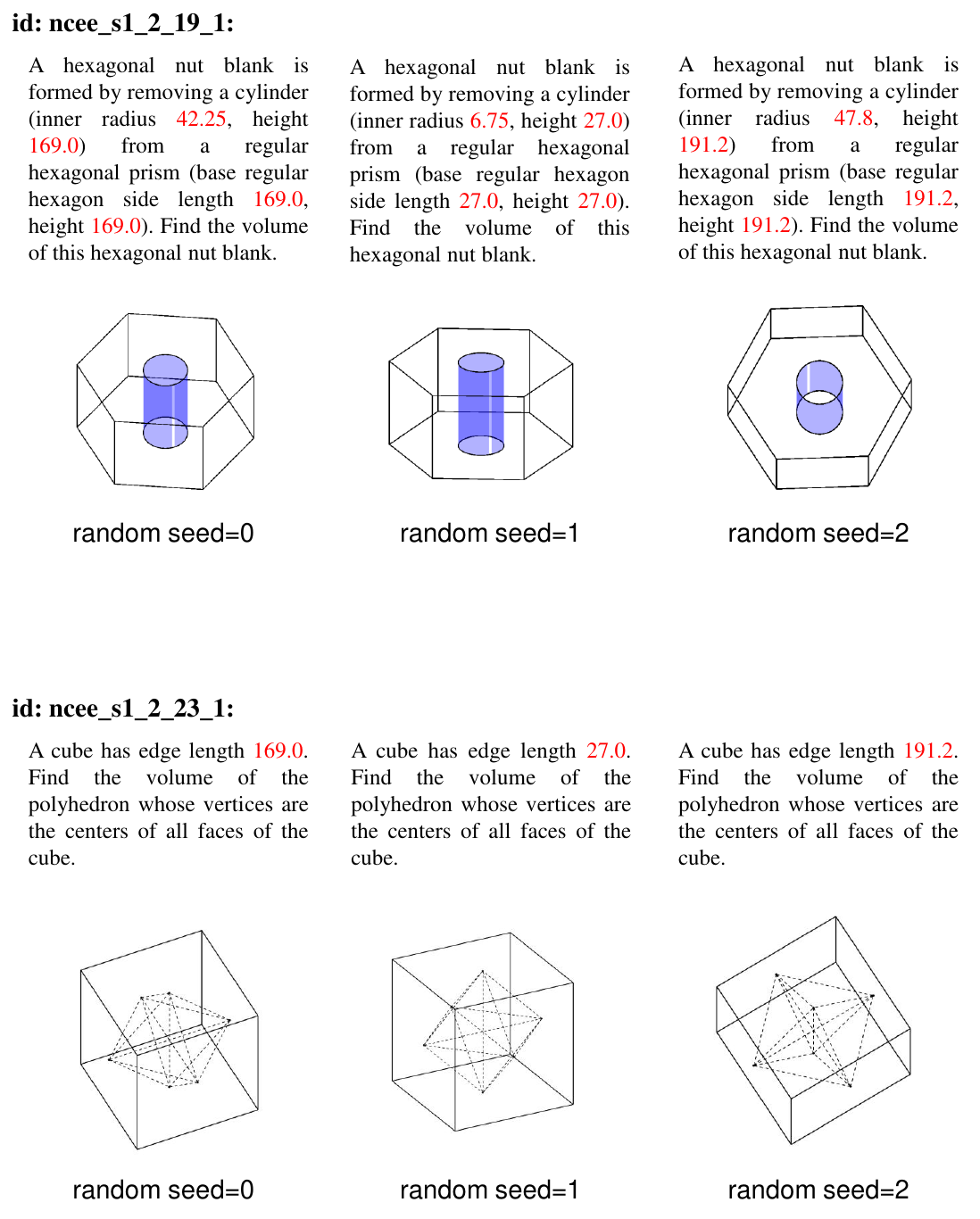}
\end{center}
\end{tcolorbox}
\end{table*}

\begin{table*}[thbp]
\begin{tcolorbox}[colback=white, colframe=gray!75!black, 
title=Question Instance Examples: Counting problems (CP), boxrule=0.5mm, arc=3mm, auto outer arc]
\begin{center}
  \includegraphics[width=1\linewidth]{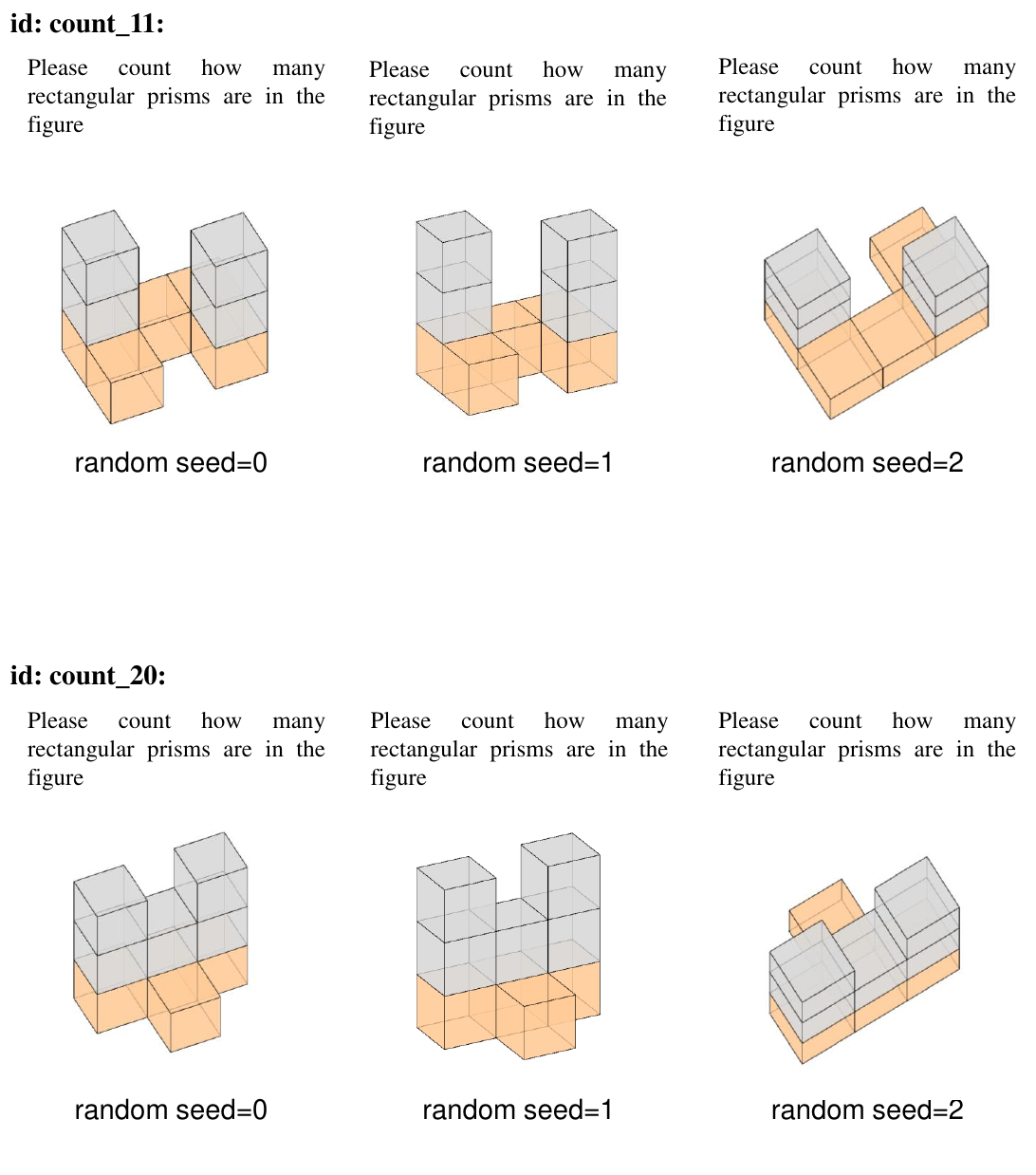}
\end{center}
\end{tcolorbox}
\end{table*}

\begin{table*}[thbp]
\begin{tcolorbox}[colback=white, colframe=gray!75!black, 
title=Question Instance Examples: Dynamic or moving-point problems (DM), boxrule=0.5mm, arc=3mm, auto outer arc]
\begin{center}
  \includegraphics[width=1\linewidth]{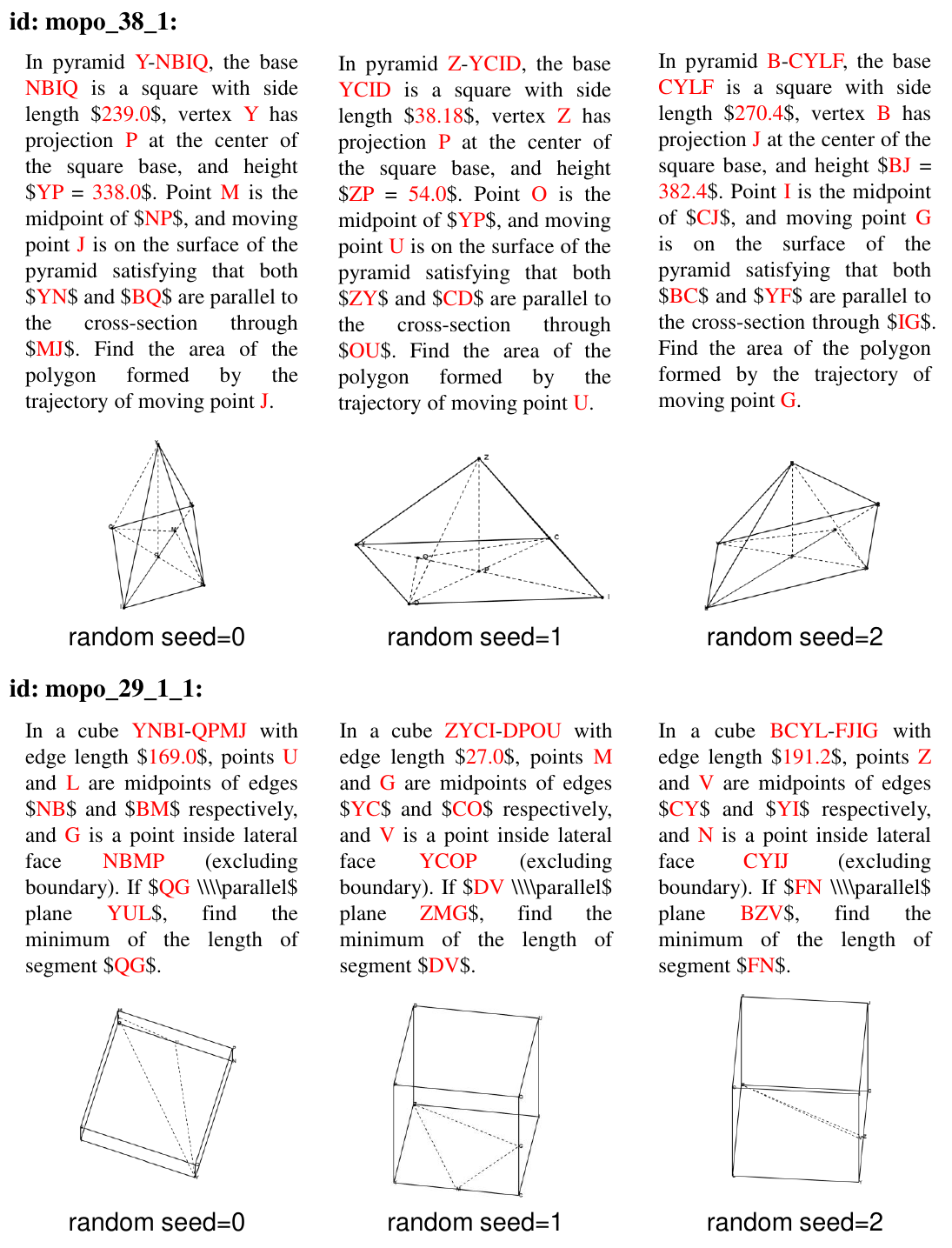}
\end{center}
\end{tcolorbox}
\end{table*}

\begin{table*}[thbp]
\begin{tcolorbox}[colback=white, colframe=gray!75!black, 
title=Question Instance Examples: Folding and unfolding problems (FP), boxrule=0.5mm, arc=3mm, auto outer arc]
\begin{center}
  \includegraphics[width=1\linewidth]{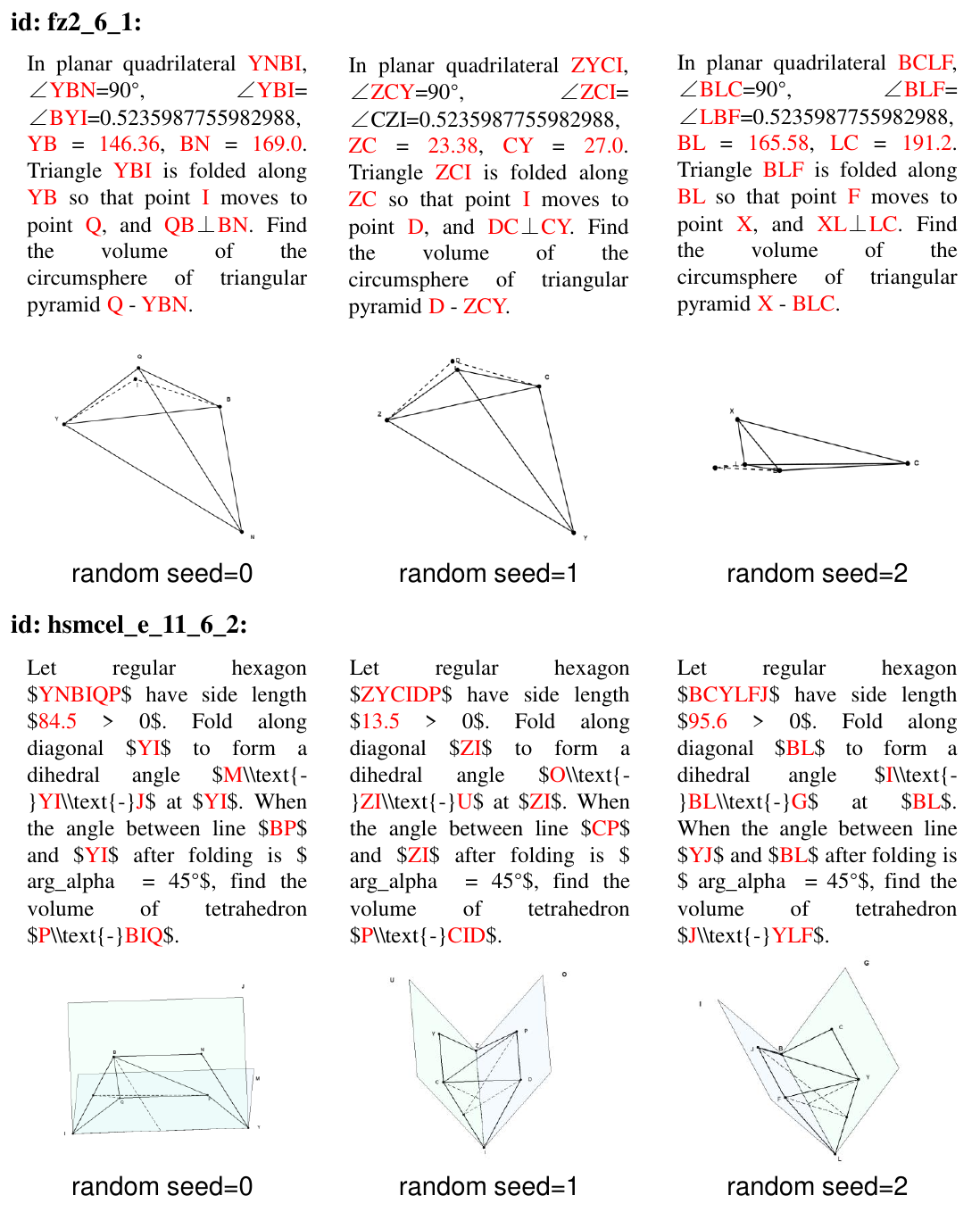}
\end{center}
\end{tcolorbox}
\end{table*}

\end{document}